\documentclass[lettersize,journal]{IEEEtran}
\usepackage{amsmath,amsfonts}
\usepackage{algorithmic}
\usepackage{algorithm}
\usepackage{array}
\usepackage[caption=false,font=normalsize,labelfont=sf,textfont=sf]{subfig}
\usepackage{textcomp}
\usepackage[normalem]{ulem}
\usepackage{stfloats}
\usepackage{url}
\usepackage{verbatim}
\usepackage{graphicx}
\usepackage{cite}
\usepackage{orcidlink}

\usepackage{enumitem}
\newlist{steps}{enumerate}{1}
\setlist[steps, 1]{label = {Step \arabic*:}, align=left}
\begin{document}

\title{Physics-informed neural networks for solving forward and inverse problems in complex beam systems}

\author{Taniya Kapoor \orcidlink{0000-0002-6361-446X}, Hongrui Wang \orcidlink{0000-0001-7194-4728}, ~\IEEEmembership{Member,~IEEE}, Alfredo Núñez \orcidlink{0000-0001-5610-6689}, ~\IEEEmembership{Senior Member,~IEEE},   Rolf Dollevoet \orcidlink{0000-0002-8279-8284}
\thanks{Manuscript submitted April 1, 2022, Revised February 24, 2023, May 19, 2023, July 22, 2023, Accepted August 23, 2023 (Corresponding author: Hongrui Wang)}
\thanks{T. Kapoor, H. Wang, A. Núñez, R. Dollevoet are with the Section of Railway Engineering, Department of Engineering Structures, Delft University of Technology, The Netherlands. (e-mail: t.kapoor@tudelft.nl; h.wang-8@tudelft.nl; a.a.nunezvicencio@tudelft.nl; r.p.b.j.dollevoet@tudelft.nl).}
}

\markboth{Accepted for Publication in IEEE Transactions on Neural Networks and Learning Systems}
{Shell \MakeLowercase{\textit{et al.}}: A Sample Article Using IEEEtran.cls for IEEE Journals}

\maketitle

\begin{abstract}

This paper proposes a new framework using physics-informed neural networks (PINNs) to simulate complex structural systems that consist of single and double beams based on Euler-Bernoulli and Timoshenko theory, where the double beams are connected with a Winkler foundation. \color{black}In particular, forward and inverse problems for the Euler-Bernoulli and Timoshenko partial differential equations (PDEs) are solved using nondimensional equations with the physics-informed loss function. Higher-order complex beam PDEs are efficiently solved for forward problems to compute the transverse displacements and cross-sectional rotations with less than $1e-3$ percent error. Furthermore, inverse problems are robustly solved to determine the unknown dimensionless model parameters and applied force in the entire space-time domain, even in the case of noisy data. The results suggest that PINNs are a promising strategy for solving problems in engineering structures and machines involving beam systems.
\end{abstract}
\color{black}

\section{Introduction}

Complex engineering issues in real-life scenarios are often characterized by the connection between various subsystems and uncertainty in behavior caused by internal and external variables and their interactions. Furthermore, the design and maintenance of complex systems, such as engineering structures and machines, is made challenging by the unpredictable collective behaviors and properties of these concurrently operating and interacting components. These issues are typically difficult to analyze through conventional methods \cite{kobbacy2008complex}. Most of these complex engineering systems are continuous, and partial differential equation (PDE) models are used to characterize and understand their behavior. These PDE models are used to simulate a wide range of engineering phenomena, ranging from multiple beam systems in suspension bridge cables (Timoshenko beam equations)\cite{chatzis2013modeling} to catenary-pantograph interactions in railways (damped beam equations) \cite{liu2018advances} to simulating air turbulence that disrupts flight (Navier-Stokes equations) \cite{foias2001navier, lucor2022simple}, among many others \cite{stojanovic2015vibrations, mercieca2016estimation, li2010modeling, kapoor2022predicting, chandra2022physics, yuan2022pinn, fallah2023physics, kapoor2023physics}. Solutions to governing PDEs enable real challenges such as structural health monitoring \cite{kapteyn2021probabilistic, wang2019bayesian, yuan2020machine} and optimal structural design \cite{borggaard1997pde, lai2021structural} to be addressed.

The development of algorithms for diagnostics and prognosis is an issue in maintaining complex engineering systems\cite{kobbacy2008complex}. Insights could be obtained by solving the forward and inverse problems for the governing PDEs of interest to forecast the system's behavior and minimize unexpected downtimes of complex systems. These equations range in complexity from being extremely nonlinear (Navier-Stokes equation \cite{raissi2018hidden}) to incorporating intricate higher-order boundary conditions (fourth-order beam equations \cite{abhyankar1993chaotic}). In practice, these equations are too complicated to be solved analytically and must be solved numerically. Numerical methods such as the finite-difference and finite-element methods have been used to approximate the solutions of these PDEs. Despite their success in practice, these methods encounter some difficulties, such as mesh creation, which is more difficult for complex geometries in higher dimensions\cite{karniadakis2021physics, fang2021high}.

In recent years, scientific machine learning, which combines scientific computing with machine learning methodologies to estimate PDEs solutions, has made remarkable developments and has emerged as a viable alternative to the aforementioned numerical methods. The review papers \cite{karniadakis2021physics, cuomo2022scientific, blechschmidt2021three} extensively discuss state-of-the-art breakthroughs in scientific machine learning, including works on real-world engineering problems. However, data-driven methods require a large amount of data, which is possibly computationally expensive and susceptible to noise in some engineering systems \cite{raissi2019physics}. One possible way to mitigate the effects of these problems is to utilize the known physical knowledge of the underlying system in the learning procedure \cite{meng2021physics, xu2022physics, 7959606}. Prior physical knowledge could be incorporated into the learning procedure by collocating the PDE residual at training points, similar to leveraging the physical equation in the training process. The underlying neural networks proposed in \cite{raissi2019physics} are called physics-informed neural networks (PINNs).

PINNs utilize neural networks' universal function approximation property \cite{hassoun1995fundamentals} and embed the well-posed physical equations modeled by PDEs in the loss function. Prior knowledge of physical principles works as a regularization agent in neural network training, restricting the space of admissible solutions and improving function approximation accuracy. As a result, given some knowledge of the physical features of the problem and some training data, PINN can be utilized to identify a high-fidelity solution. PINNs have already proven to be a very effective paradigm for approximating solutions of PDEs for real-world problems \cite{oszkinat2022uncertainty, hua2023physics}, as discussed in the review papers \cite{cuomo2022scientific, karniadakis2021physics}. 

However, several challenges for PINNs have also been found \cite{wang2022and}. One such challenge for PINNs is to learn relevant physical phenomena for more complex problems with large coefficients in the physical equation \cite{krishnapriyan2021characterizing}. A sequence-to-sequence learning task was proposed in \cite{krishnapriyan2021characterizing} as a remedy to this problem. However, this can be computationally expensive when the scale is large. In \cite{kissas2020machine}, the importance of using nondimensional equations in the PINN framework was highlighted for cardiovascular blood flow. We build on these works and address the challenge of multiscale complex beam systems. Accordingly, this paper uses nondimensional PDEs instead of dimensional PDEs in the loss function. This provides a way to simulate realistic physical equations with computational tractability.

 Accurate prediction of the dynamics of structures \cite{9597485} and structural elements, such as plates \cite{joghataie2013simulating}, and beams \cite{9741838, 7879337}, is crucial in the field of structural engineering. However, measuring quantities of interest in beam systems through lab experiments can prove to be difficult, as it necessitates specialized prototypes, training, and safety during the testing process, increasing the overall cost of the experiment. PINNs offer a simulation-based solution as a mesh-free method that does not require discretizing the domain into a finite number of elements, making it computationally inexpensive compared to numerical methods. PINNs can effectively integrate incomplete or noisy information with prior physical knowledge. The proposed framework converts dimensionalized PDEs to a nondimensionalized form, increasing the suitability for neural networks and enabling the prediction of deflections and rotations for any material, resulting in a more generalizable method.

This paper provides a framework to simulate complex structural systems consisting of two or more basic structural systems connected by an elastic layer. In particular, the forced vibration of two elastically connected beams is studied, which is commonly encountered in the mechanical, construction, and aeronautical industries \cite{stojanovic2015vibrations}. These double-beam systems in engineering structures have received significant attention in the scientific community and are considered complex systems. Studies have been conducted to predict the dynamics of these systems under various loading and force conditions, such as those found in papers \cite{onis1, stojanovic2012forced, onis2, abu2006dynamic, zhao2020forced, li2007spectral, palmeri2011galerkin, ying2017response, ying2018vibration}, among others. These studies include the use of analytical and closed-form solutions \cite{zhao2020forced, murmu2010nonlocal, chen2021closed, liu2019closed, zhao2021free}; however, analytical methods have limitations in applicability, as they may be useful only for specific types of problems and can become complex for systems with many variables or nonlinear equations. Other approaches, such as the state-space method presented in \cite{li2021state, palmeri2011galerkin}, may also be computationally expensive for systems with a large number of states. Additionally, modal analysis methods as presented in \cite{ong2022coupled, stojanovic2015vibrations} have been used to study the natural frequencies and modes of vibration, but they do not provide information on the full response of the system and cannot be used to predict the time-domain response at any instant.

The considered governing equations are modeled using  Euler-Bernoulli and Timoshenko theory. In addition to solving the forward problem and computing the physical quantities of interest, we also solve the inverse problem. For the inverse problem, one may not necessarily have complete information about the inputs to the PDEs, such as initial or boundary data, coefficients \cite{9298920, 7302572, 9716789} or applied forces. This lack of knowledge makes the forward problem ill-posed, and subsequently, the forward problem cannot be solved uniquely. In this paper, access to data for quantities of interest is leveraged to determine the PDEs' unknown inputs, for instance, the model parameters and applied forces. 

The main contributions of the current paper are as follows,

\begin{itemize}
  	\item To the best of the authors' knowledge, this is the first work to use physics-informed machine learning to solve the forward and inverse problems of Euler-Bernoulli and Timoshenko complex beam models.

	\item We address a challenge for PINNs in solving multiscale complex beam PDEs and propose a framework for using nondimensional equations in the loss function.
	\item The proposed nondimensional PINN framework is employed to address ill-posed inverse problems for complex systems and to identify the unknown model parameters and the applied force on the beam components. This is achieved by utilizing data from indirect measurements such as the displacement and cross-sectional rotations of the beams.
    \item The presented methodology is robust to noise and can accomodate potential uncertainty in the measurement data, making it well suited for real-world applications where data are incomplete or uncertain.
\end{itemize}

The rest of the article is organized as follows. In Section II, the PINN method is presented to simulate the dimensional Euler-Bernoulli beam equation. Due to the limitations of PINNs in simulating the dimensional Euler-Bernoulli beam equation, an alternative approach of using nondimensional equations in the PINN's loss function is proposed and successfully used to solve the dimensionless Euler-Bernoulli equation in Section III. Section IV first applies the proposed framework to simulate the Timoshenko beam model for solving forward and inverse problems. The forward problem of the Euler-Bernoulli double-beam equation is then solved. Additionally, Section IV covers forward and inverse Timoshenko double-beam system problems. Section V concludes this paper.

\section{PINNs for Dimensional PDEs}
In this section, the method of PINNs to simulate PDEs is presented in brief using an abstract dimensional PDE. The method is then used to simulate the dimensional Euler-Bernoulli equation. The following abstract dimensional PDE is considered with implicit initial and boundary conditions:

\begin{equation}
\mathcal{\bar{K}}(\bar{x}, \bar{t}) := \mathcal{D}[\bar{u}](\bar{x}, \bar{t}; \bar{\lambda}) - \bar{f}(\bar{x}, \bar{t}) \quad \forall (\bar{x}, \bar{t})\in  \bar{\Omega} \times \bar{T} \subset  \mathbb{R}^\mathrm{d} \times \mathbb{R}
\label{eq1}
\end{equation}
where $\mathcal{D}[^.]$ denotes the differential operator, $\bar{u}$ is the quantity of interest, $\bar{x} \in$ $\bar{\Omega}$ $\subset$ $\mathbb{R}^\mathrm{d}$, $\bar{t} \in \bar{T}$ $\subset$ $\mathbb{R}$ for $d \geq 1$, $\bar{\Omega}$ denotes the spatial boundary contained in the d-dimensional Cartesian spatial  space and $\bar{T}$ denotes the temporal domain, $\bar{\lambda} \in \mathbb{R}$ is the model parameter, $\bar{f}(\bar{x}, \bar{t})$ is the external force, and $\bar{K}$ is the notation for the abstract physical equation.

Deep neural networks are the core for PINNs in which inputs $(\bar{x}, \bar{t})$ map to output ($\bar{u}$) through an iterative composition of hidden layers. The composition consists of weights ($w$), biases ($b$), and linear or nonlinear activation function(s) ($\sigma$). The inputs undergo a linear composition within a neuron, where they are multiplied by respective weights and summed along with a bias term. Subsequently, this combined input is passed through a nonlinear activation function $(\sigma)$ as presented in Fig.~\ref{fig4}. This allows the neural network to introduce nonlinearity, enabling the network to capture intricate relationships between inputs and outputs.

To train the neural network, one needs training set ($\Delta$), consisting of spatial boundary points ($\Delta_\mathrm{b}$), temporal boundary points ($\Delta_\mathrm{i}$) and interior points ($\Delta_\mathrm{int}$). As a result, the training set can be written as $\Delta=\Delta_\mathrm{i} \cup \Delta_\mathrm{b} \cup \Delta_\mathrm{int}$. In this work, $\Delta_\mathrm{i}$, $\Delta_\mathrm{b}$, and $\Delta_\mathrm{int}$ are considered to have $N_\mathrm{i}$, $N_\mathrm{b}$ and $N_\mathrm{int}$ training points respectively. The total number of training points is denoted by $N_\mathrm{train}$. To approximate the quantity of interest $\bar{u}$, one needs to minimize the loss function containing the physical model in the form of a PDE with initial and boundary conditions of ~\eqref{eq1}. No additional data are required in the loss function for forward problems. The loss function $\bar{\mathcal{L}}$ is defined as follows:

\begin{equation}
    \bar{\mathcal{L}}(\theta) = \mathrm{\underset{\mathrm{\theta} }{Min}}(\frac{1}{N_\mathrm{train}}\sum_{n=1}^{ N_\mathrm{train}}||\mathcal{\bar{K}}(\bar{x}_\mathrm{n}, \bar{t}_\mathrm{n})||^2)
\label{eq2}
\end{equation}
where $(\bar{x}_\mathrm{n}, \bar{t}_\mathrm{n})$ represents the training tuple for each n. Minimizing this loss function using a suitable optimization algorithm provides optimal parameters $\theta =\{w, b\}$. 

Now, we employ the PINN algorithm for the dimensional Euler-Bernoulli beam equation and evaluate the corresponding performance. The dynamic Euler-Bernoulli beam equation is given by

\begin{equation}
    \rho A \bar{u}_\mathrm{\bar{t}\bar{t}} + EI \bar{u}_\mathrm{\bar{x}\bar{x}\bar{x}\bar{x}} = \bar{f}(\bar{x}, \bar{t}) \quad \bar{x} \in [0, \bar{l}], \bar{t} \in [0, t_\mathrm{end}]
\label{eq3}
\end{equation}

\begin{figure}[htbp]
\centerline{\includegraphics[width=0.8\columnwidth]{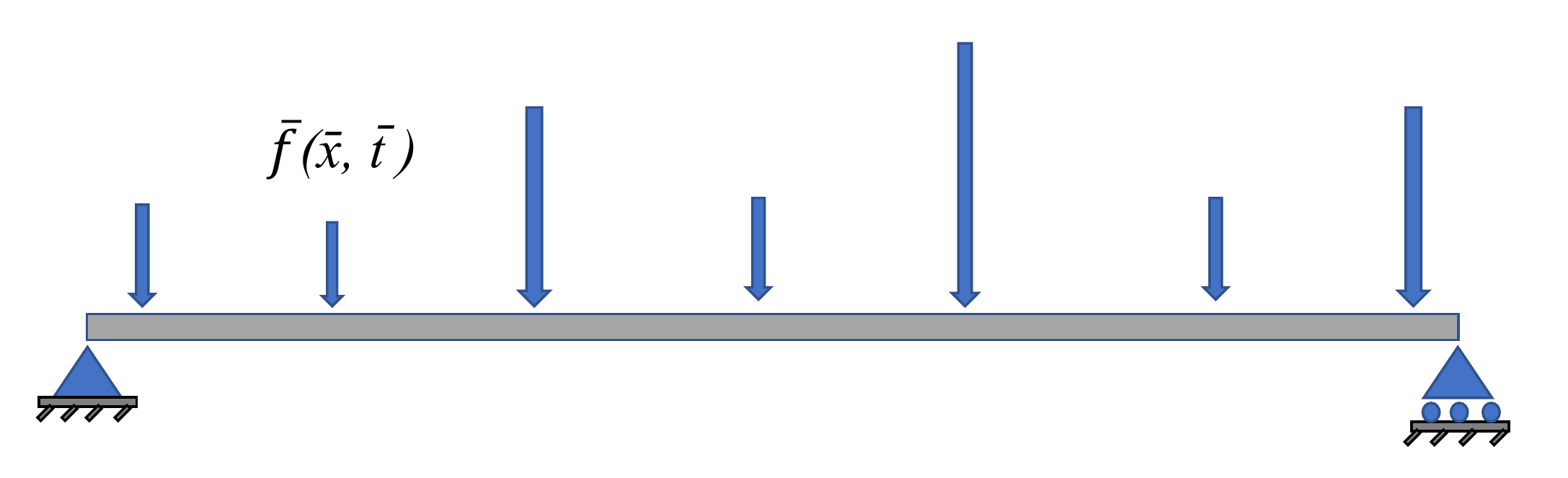}}
\caption{Simply supported beam with varying transverse force.}
\label{fig1}
\end{figure}

Here, $\bar{l}$ and $t_{\mathrm{end}}$ refer to the length of the beam and final time, respectively. This equation models the transverse displacement of beam $\bar{u}$ in the space-time domain subject to the external transverse force $\bar{f}$ as shown in Fig.~\ref{fig1}. This work considers a uniform cross-sectioned beam with constant material properties throughout the beam. The parameters $\rho$ and $A$ denote the density and cross-sectional area of the beam, respectively. The parameters $E$ and $I$ are Young's modulus and the moment of inertia of the beam, respectively. The external force $\bar{f}$ acts nonuniformly on the body, and $\bar{u}$ is the transverse displacement of the beam, which is the only unknown in the governing PDE. In addition, $u_\mathrm{tt}$ represents the second order partial derivative of u with respect to t, and $u_\mathrm{xxxx}$ represents the fourth order partial derivative of $\textit{u}$ with respect to x.
The goal of the forward problem is to compute the transverse displacement of the beam supplemented with the initial and boundary conditions. For this study, simply supported beams are considered, which rest on two supports and are free to move horizontally. Real-world applications of simply supported beams include railway tracks, and bridges, to name a few. Mathematically, the simply supported boundary condition for~\eqref{eq3} is given by 

\begin{equation*}
    \bar{u}(0, \bar{t}) = \bar{u}(\bar{l}, \bar{t}) = \bar{u}_\mathrm{\bar{x}\bar{x}}(0, \bar{t}) = \bar{u}_\mathrm{\bar{x}\bar{x}}(\bar{l}, \bar{t}) = 0
\end{equation*}

For the numerical experiment, the parameter values of aluminium-like material are considered in the physical equation, which are widely used for making beams. The parameter values taken for the problem are $\rho = 2 \times 10^{3}$kg/$\mathrm{m}^{3}$, $A = 5 \times 10^{-2}$ $\mathrm{m}^{2}$, $E = 10^{10}$N/$\mathrm{m^{2}}$, and $I = 4 \times 10^{-4}$$\mathrm{m^{4}}$. Additionally, the beam is taken to be $\pi^{2}$ meters long, and the external force $\bar{f}$ is taken to be $EI(1 - 16\pi^2)\sin{(\bar{x}/\pi)}\cos(4c\bar{t}/\pi)/\bar{l}^{3}$N, where $c = \sqrt{\frac{EI}{\rho A}}$. Taking the final time to be $\pi^{2}/200$, the PDE to be solved takes the form

\begin{multline}
    10^{2} \bar{u}_\mathrm{\bar{t}\bar{t}} + 4\times 10^{6} \bar{u}_\mathrm{\bar{x}\bar{x}\bar{x}\bar{x}} = \\
    4\times 10^{6}(1 - 16\pi^2)\sin{(\bar{x}/\pi)}\cos(800\bar{t}/\pi)/\pi^{3}
\label{eq4}
\end{multline}
in the domain $\bar{x} \in [0, \pi^{2}]$ and $\bar{t} \in [0, \pi^{2}/200]$. For~\eqref{eq4} to be well-posed the initial condition of the beam is taken to be $\sin(\bar{x}/l)$ with zero initial velocity, where $l=\sqrt{\bar{l}}$.

\begin{figure}[htbp]
\centerline{\includegraphics[width=0.6\columnwidth]{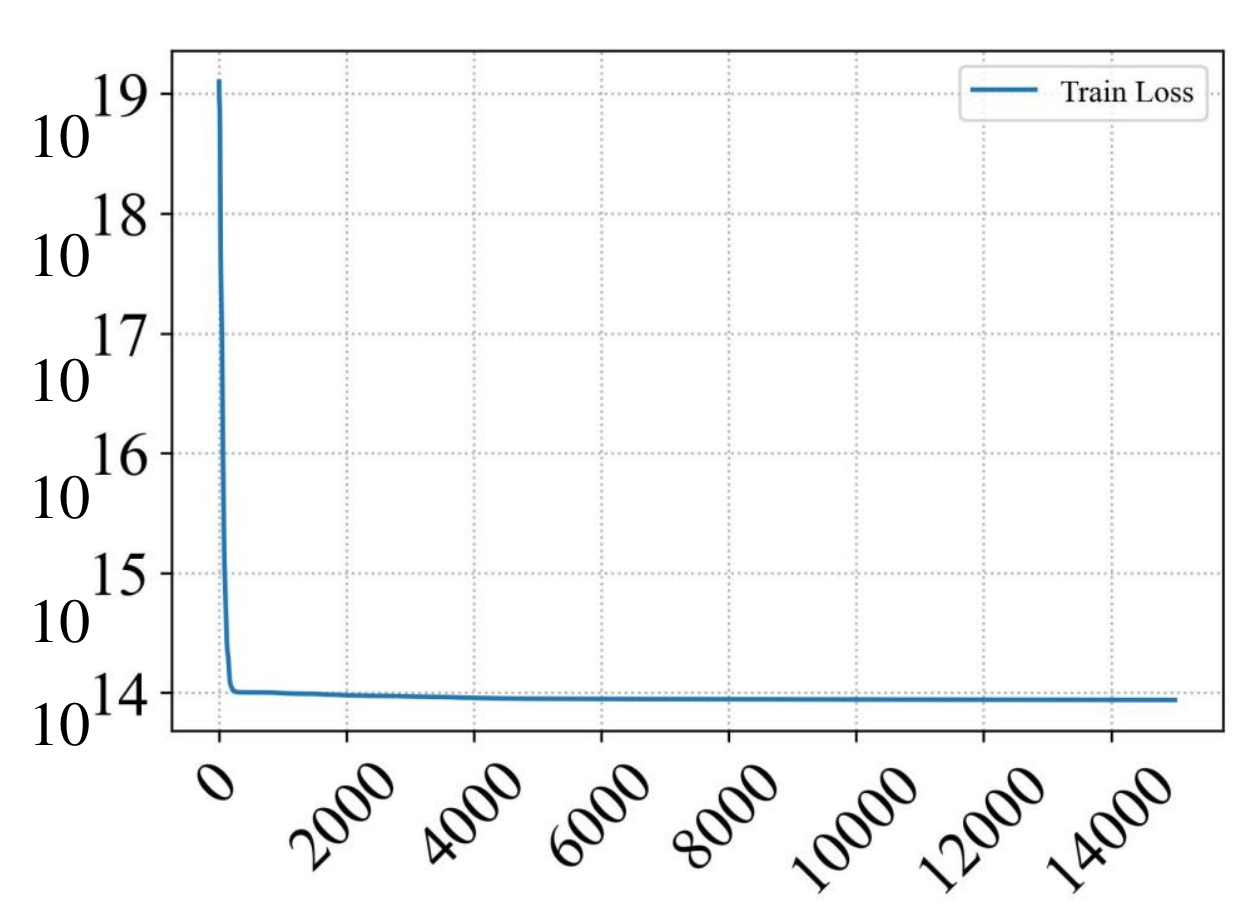}}
\caption{L-BFGS training loss vs. the number of epochs for the dimensional Euler-Bernoulli beam equation.}
\label{fig2}
\end{figure}

For training the neural network, $16000$ random training points are generated with the distribution $N_{\mathrm{i}} = 2000$, $N_{\mathrm{b}} = 4000$, and $N_{\mathrm{int}}=10000$. The neural network consists of $4$ hidden layers with $20$ neurons in each hidden layer. The $\tanh$ activation function, which is one of the most commonly used activation functions in the PINN literature, as described in the review paper \cite{cuomo2022scientific}, is chosen. The loss function~\eqref{eq2} consists of the initial condition, boundary condition and PDE. The PDE is regularized in the loss function with the residual parameter $0.1$\cite{mishra2022estimates}. The L-BFGS optimizer, which is again one of the most commonly used optimizers in the PINN literature\cite{cuomo2022scientific} is used to minimize the loss function. As shown in Fig.~\ref{fig2} $15000$ epochs are performed. However, the figure clearly illustrates that the optimizer does not converge to the solution, and a vast training loss of $10^{14}$ is obtained. Additionally, the graph shows that the optimizer is stuck in the local minima and hence will not converge even if the number of epochs is increased for the same neural network configuration. 

In \cite{yuan2020machine, bazmara2023physics}, the problem of free vibrations in the Euler-Bernoulli single-beam equation was successfully solved by PINNs, where the coefficients of the PDE were taken to be unity. This shows that PINNs can simulate the beam equations, and the challenge lies in the multiscale coefficient values that arise when dealing with a real-life physical equation. The nonconvergence in our case is due to the high value of coefficients, which is due to the dimensional equation. Consequently, a pressing need arises to transform the dimensional form of the equation into a nondimensional form. It may be possible that for some configurations containing hundreds of hidden layers and neurons, this problem may be solved without the need to non-dimensionalizing the PDE. However, nondimensionalization aims to provide computational tractability. 

\section{PINNs for NonDimensional PDEs}
This section presents the proposed framework of using nondimensional equations in the PINN loss function. The method for nondimensionalizing the governing PDE is described first. Then, the algorithms for forward and inverse problems using dimensionless equations in PINNs are presented. To nondimensionalize the abstract PDE given by~\eqref{eq1}, the following transformations are performed
\begin{equation}
\bar{x} = \xi_1(x); \quad \bar{t} = \xi_2(t); \quad \bar{u} = \xi_3(u); \quad \bar{f} = \xi_4(f)
\label{eq5}
\end{equation}
where, $\xi_1$, $\xi_2$, $\xi_3$, and $\xi_4$ are suitable functions that map the dimensional quantities $\bar{x}$, $\bar{t}$, $\bar{u}$, and $\bar{f}$ to the corresponding nondimensional quantities. After substituting the above transformations in ~\eqref{eq1} and introducing the dimensionless parameter $\lambda$, one obtains
\begin{equation}
\mathcal{K}(x, t) := \mathcal{D}[u](x, t; \lambda) - f(x, t) \quad \forall (x, t)\in  \Omega \times T \subset  \mathbb{R}^d \times \mathbb{R}
\label{eq6}
\end{equation}

The proposed framework uses dimensionless equations to simplify and stabilize the problem computationally. By nondimensionalizing the variables and parameters, they are kept within a specific range, resulting in improved performance and generalization of the neural network. Furthermore, dimensionless equations generate more interpretable solutions by eliminating the units of measure, making it easier to understand the underlying physical phenomena and to compare results across different physical systems in the form of ratios and parameters. Hence, using dimensionless equations in PINNs can enhance the neural network's computational stability, generalization, and interpretability.

\subsection{PINN Framework for Forward Problems}

$\mathcal{K}$, the nondimensional PDE corresponding to the dimensional PDE $\mathcal{\bar{K}}$, is now used in the loss function $\mathcal{L}$ defined as follows:

\begin{equation}
    \mathcal{L}(\theta) = \mathrm{\underset{\mathrm{\theta} }{Min}}(\frac{1}{N_\mathrm{train}}\sum_{n=1}^{ N_\mathrm{train}}||\mathcal{K}(x_\mathrm{n}, t_\mathrm{n})||^2)
\label{eq7}
\end{equation}
\begin{figure*} 
\centering
\includegraphics[width=1\columnwidth]{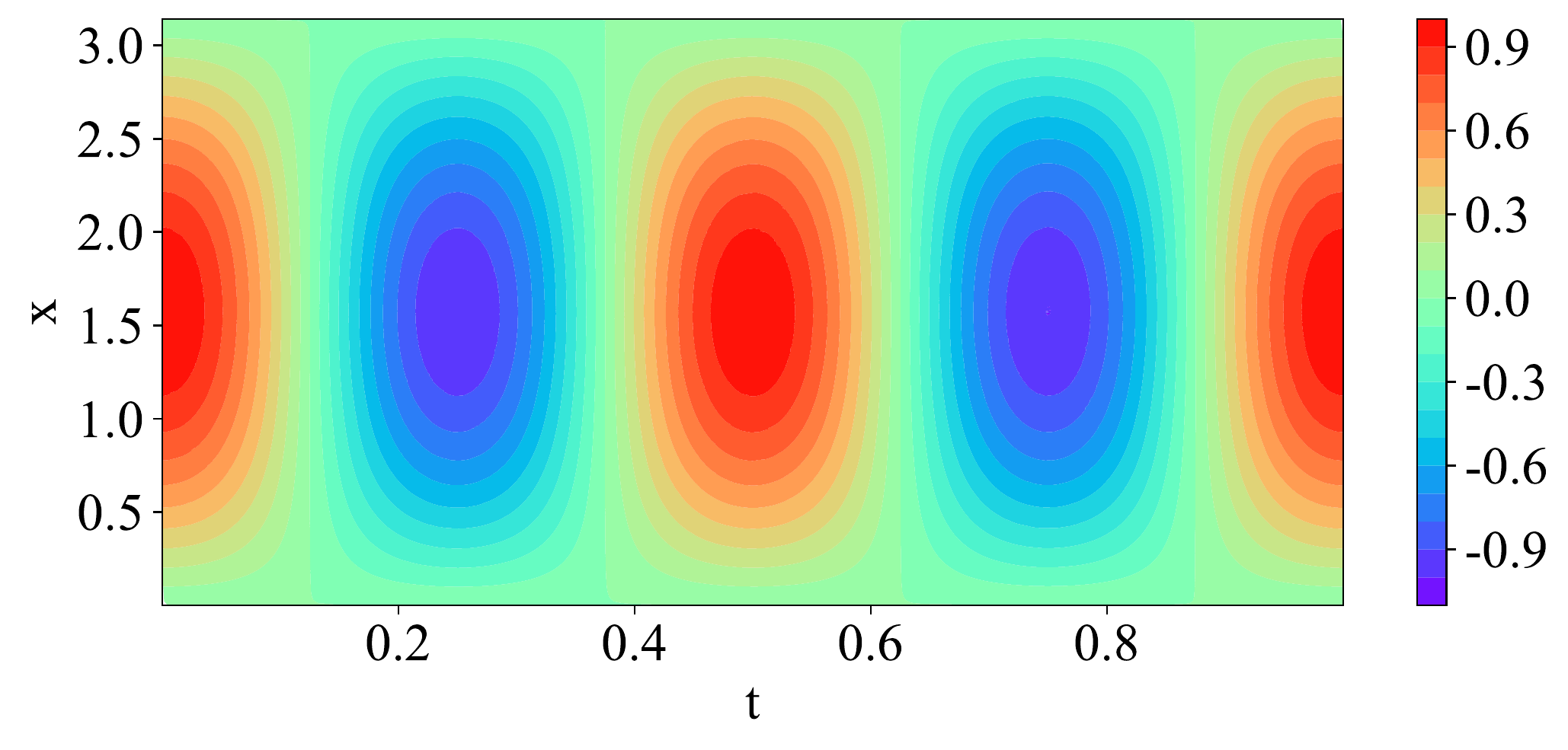}
\includegraphics[width=1\columnwidth]{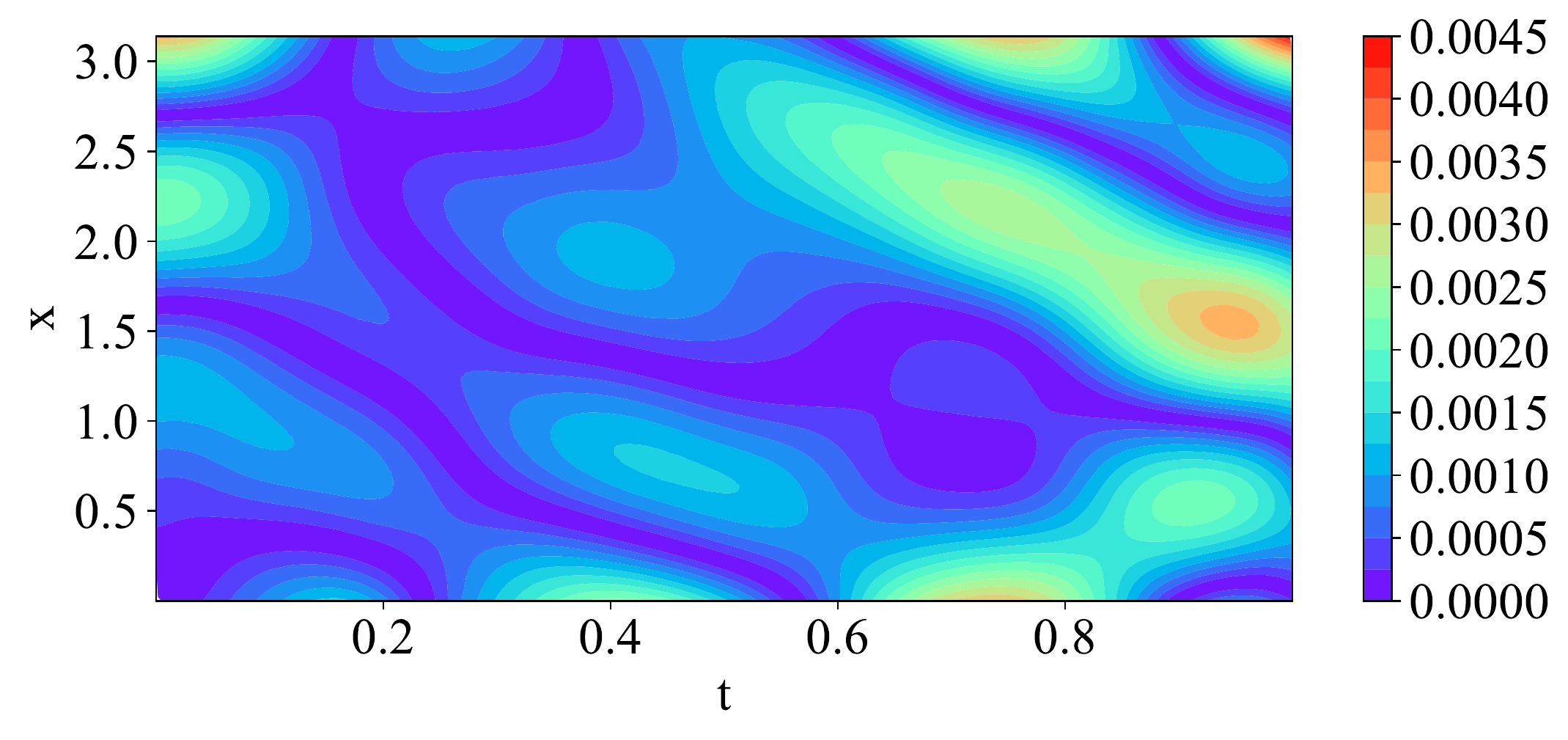}
\caption{Nondimensional Euler-Bernoulli beam equation Color bar represents \textbf{Left:} Predicted solution ($u^{*}$); \textbf{Right:} Absolute error in prediction ($|u - u^*|$)  }
\label{fig3}
\end{figure*}

\begin{figure}[htbp]
\centerline{\includegraphics[width=1\columnwidth]{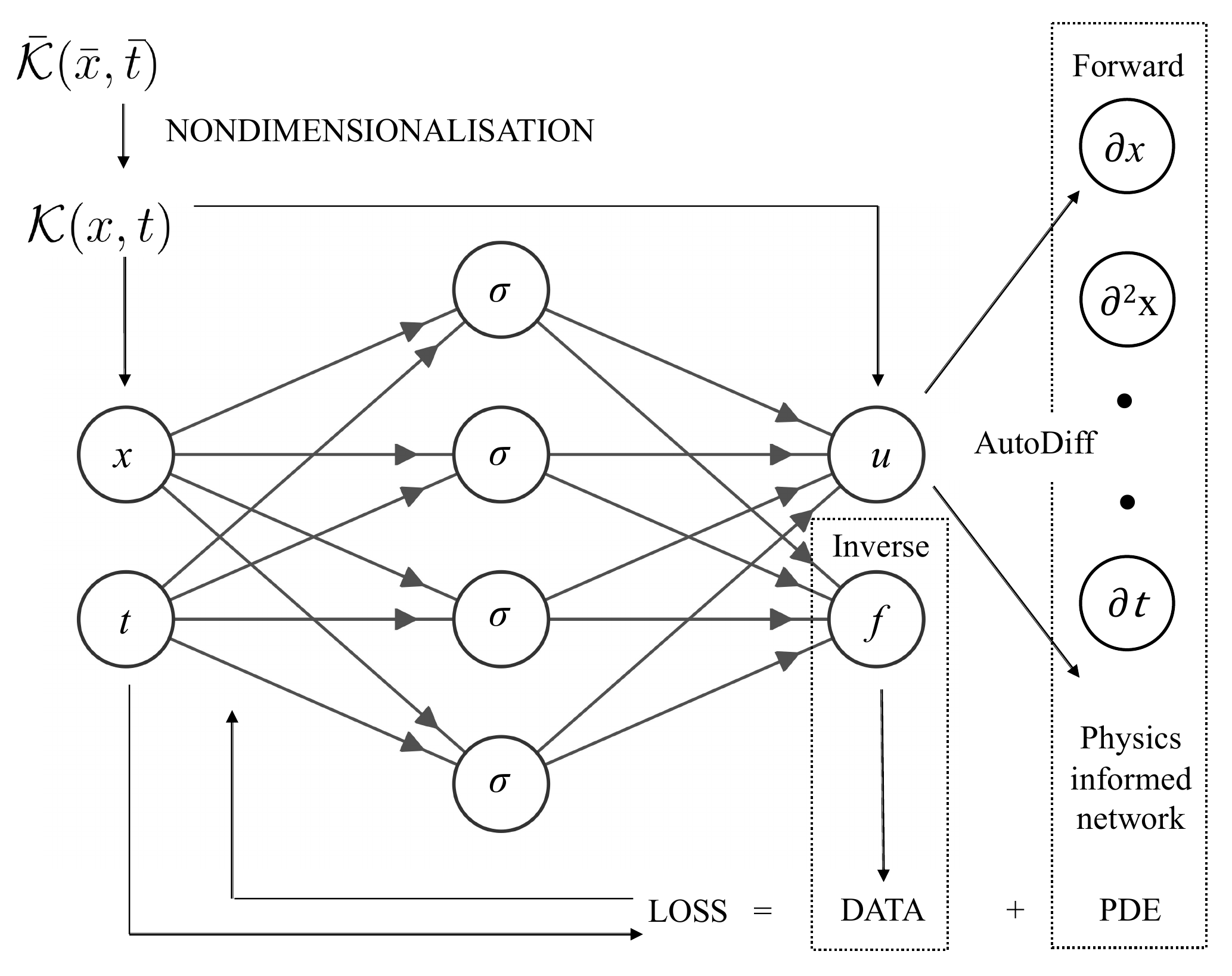}}
\caption{PINN framework for beam systems: For forward problems, the loss function comprises the nondimensional PDEs and the boundary and initial conditions. For inverse problems, the nondimensional PDEs are supplemented with extra data and potential initial/boundary conditions.}
\label{fig4}
\end{figure}

A schematic representation of the proposed PINN-based framework is illustrated in Fig~\ref{fig4}. 

\subsection{Nondimensional Euler-Bernoulli Beam Equation}
We now test the nondimensional equation in the PINN framework and evaluate the corresponding performance. To nondimensionalize~\eqref{eq3}, following transformations are used:

\begin{equation}
u = \bar{u}/l;\quad x = \bar{x}/l;\quad t = c\bar{t}/l^{2};\quad f = \bar{f}l^{3}/(EI) 
\label{eq8}
\end{equation}

Upon substituting these values in~\eqref{eq3}, one obtains

\begin{equation}
    u_\mathrm{tt} + u_\mathrm{xxxx} = f(x, t) \quad x \in [0, \pi], t \in [0, 1]
    \label{eq9}
\end{equation}
where $f(x, t) = (1 - 16\pi^2)\sin{(x)}\cos(4\pi t)$, with initial and boundary conditions

\begin{equation*}
u(x, 0) = \sin(x), \quad u_{t}(x, 0) = 0
\end{equation*}
\begin{equation*}
    u(0, t) = u(\pi, t) = u_\mathrm{xx}(0, t) = u_\mathrm{xx}(\pi, t) = 0
\end{equation*}

For the error estimation, the relative percentage error ($\mathcal{R}$) used in \cite{mishra2022estimates} is chosen. Here, $u^*$ is the prediction and $u$ is the analytical solution.

\begin{equation*}
\begin{aligned}
\mathcal{R} = \frac{||u^{*} - u ||_{2}}{||u||_{2}} \times 100
\end{aligned}
\end{equation*}

The same neural network architecture as the previous case is chosen to solve this resulting nondimensional PDE. A low training loss is obtained, indicating that the PINN is trained successfully. The analytical solution for this case is $u(x, t) = \sin(x)\cos(4\pi t)$, which is used to quantify the error in the approximated solution. The nondimensional displacement of the Euler-Bernoulli beam is computed within \mbox{$\mathcal{R} = 5.3e-4$} percent. The nondimensional displacement prediction using PINN is shown in Fig.~\ref{fig3}.(a). Fig.~\ref{fig3}.(b) shows the absolute error between the exact and predicted solutions. 

The contour plot for the approximate solution shows the dynamics of a simply supported beam under a force, where the x-axis represents the time, the y-axis represents the position along the length of the beam, and the colors represent the displacement of the beam. In Fig.~\ref{fig3}.(b) the red regions indicate high displacement, while the blue regions indicate low displacement. There is a strong displacement at the position of the beam when a substantial force is applied, which is consistent with the known physics of this system. The network accurately captures the displacement behavior of the beam, which is evident by the smooth and continuous transition of colors across the plot.

The contour plot for the error in Fig.~\ref{fig3}.(b) shows the difference between the approximate solution obtained from the network and the true solution. The x-axis represents the time, the y-axis represents the position along the length of the beam, and the colors represent the error. The red regions indicate high error, while the blue regions indicate low error. The areas where the training point concentration is low account for more error, and areas where the concentration of training points is more have relatively low error. One approach to reduce the error is to have more training points in the regions of high error. However, the overall error is low, which indicates that the network accurately captures the displacement behavior of the beam.

From Fig.~\ref{fig3}.(b), the PINNs are found to solve the dimensionless Euler-Bernoulli beam equation accurately and hence, for all further experiments,  nondimensional PDEs are simulated using PINNs. Additionally, the nondimensional displacement is henceforth referred to as displacement for conciseness. The presented methodology predicts the dimensionless quantities and hence all the plots of results and their associated error plots are dimensionless. Consequently no units are mentioned in the plots of the presented results. Next, we describe the inverse problem-solving strategy using nondimensional equations.
     
\subsection{PINN Framework for Inverse Problems}

The abstract dimensionless PDE described by~\eqref{eq6} is well-posed, and the forward problem can be solved uniquely. However, in the case of an inverse problem, the problem is ill-posed and either the initial/boundary conditions or the parameters/forces are unknown. Hence the generic abstract PDE can be re-written as
\begin{equation}
\mathcal{K^{'}}(x, t) := \mathcal{D}[u](x, t; \lambda) - f(x, t) \quad \forall (x, t)\in  \Omega \times T \subset  \mathbb{R}^d \times \mathbb{R}
\label{eq10}
\end{equation}

The algorithm for the PINN framework is presented to solve inverse problems.

\begin{algorithm}[H]
\caption{Inverse PINN algorithm}\label{alg:alg2}
\begin{steps}
\item[\bf{Goal:}] To predict the unknown parameter $\bar{\lambda}$ or function $\bar{f}(\bar{x},\bar{t})$.
\item [\bf{Step 1:}] Nondimensionalize the governing PDE to approximate the dimensionless parameter $\lambda$ or function $f(x, t)$.
\item [\bf{Step 2:}] Choose the training set from the space-time domain $\Omega \times {T}$, and augment with ($x_{\mathrm{data}}, t_\mathrm{data}$) at which additional data ($u_\mathrm{data}$) are provided. 
\item [\bf{Step 3:}] Construct a feedforward deep neural network with inputs $(x, t)$ and outputs $u$, $\lambda$ or $f(x, t)$.
\item [\bf{Step 4:}] Minimize the loss function~\eqref{eq11} with a suitable optimization algorithm, and find the optimal parameters.
\item [\bf{Step 5:}] Use the optimal parameters to approximate the parameter $\lambda^{*}$ or the function $f^{*}(x,t)$.
\end{steps}
\end{algorithm}

The aim of the inverse problem is to predict the unknown parameter $\lambda$ or the force function $f(x, t)$, when data are provided for the observable $u$ in some part of the training domain. In this paper, $u_\mathrm{data}$ denotes the available data for the inverse problem at $N_\mathrm{data}$ points. The prediction of the unknown parameter requires additional information in the loss function as shown in Fig~\ref{fig4}. It is essential for the Jacobian matrix utilized in the inverse operation study employing neural networks to exhibit a nonzero determinant, to be invertible, and to possess a reasonable ratio between its largest and smallest eigenvalues to guarantee a unique solution and ensure computational stability. The algorithm for the inverse problem is the same as for the forward problem with a minor modification in the loss function. In addition to the output $u$, the PINNs now predict the unknown parameter, force, initial or boundary conditions of the physical problems by leveraging the known data. The loss function for the inverse problem can be defined as

\begin{figure*} 
\centering
\includegraphics[width=1\columnwidth]{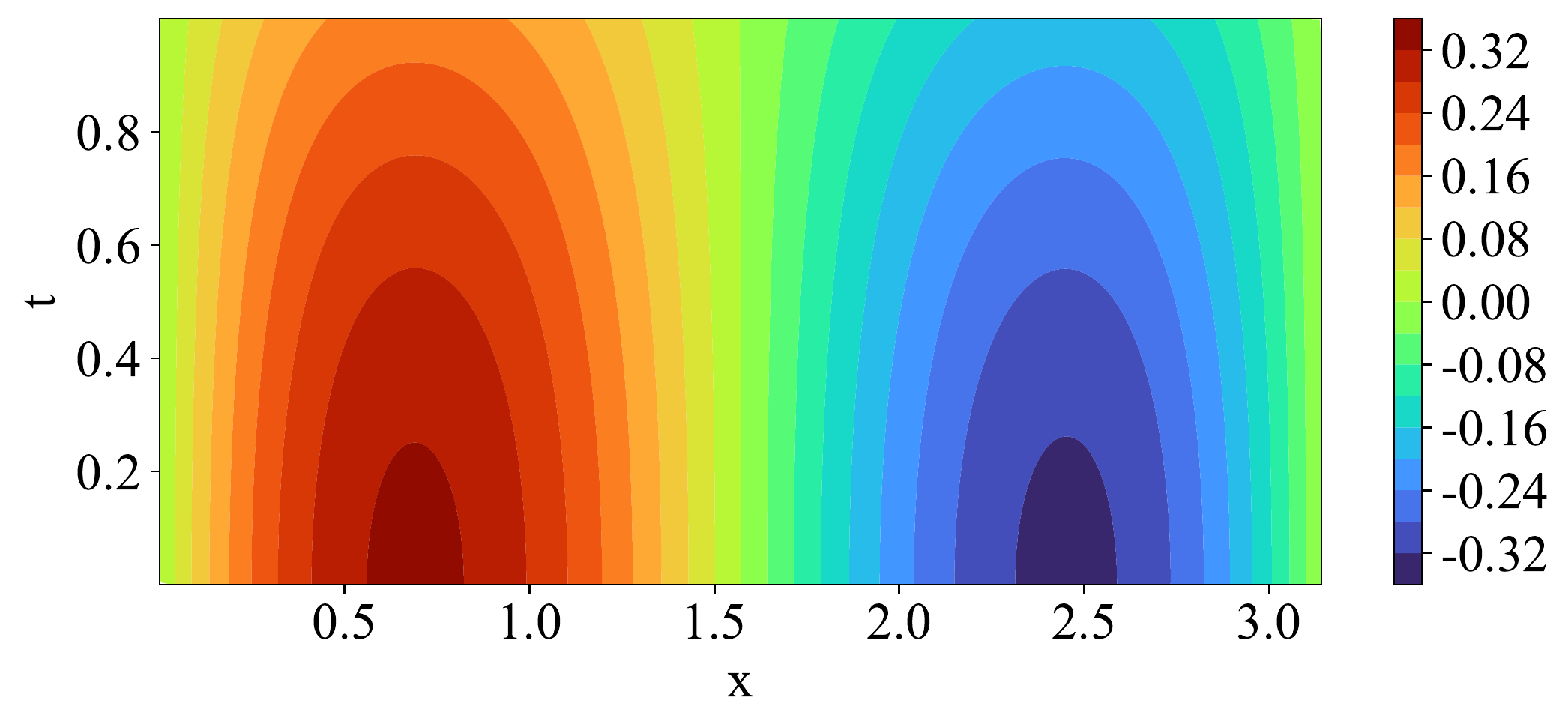}
\includegraphics[width=1\columnwidth]{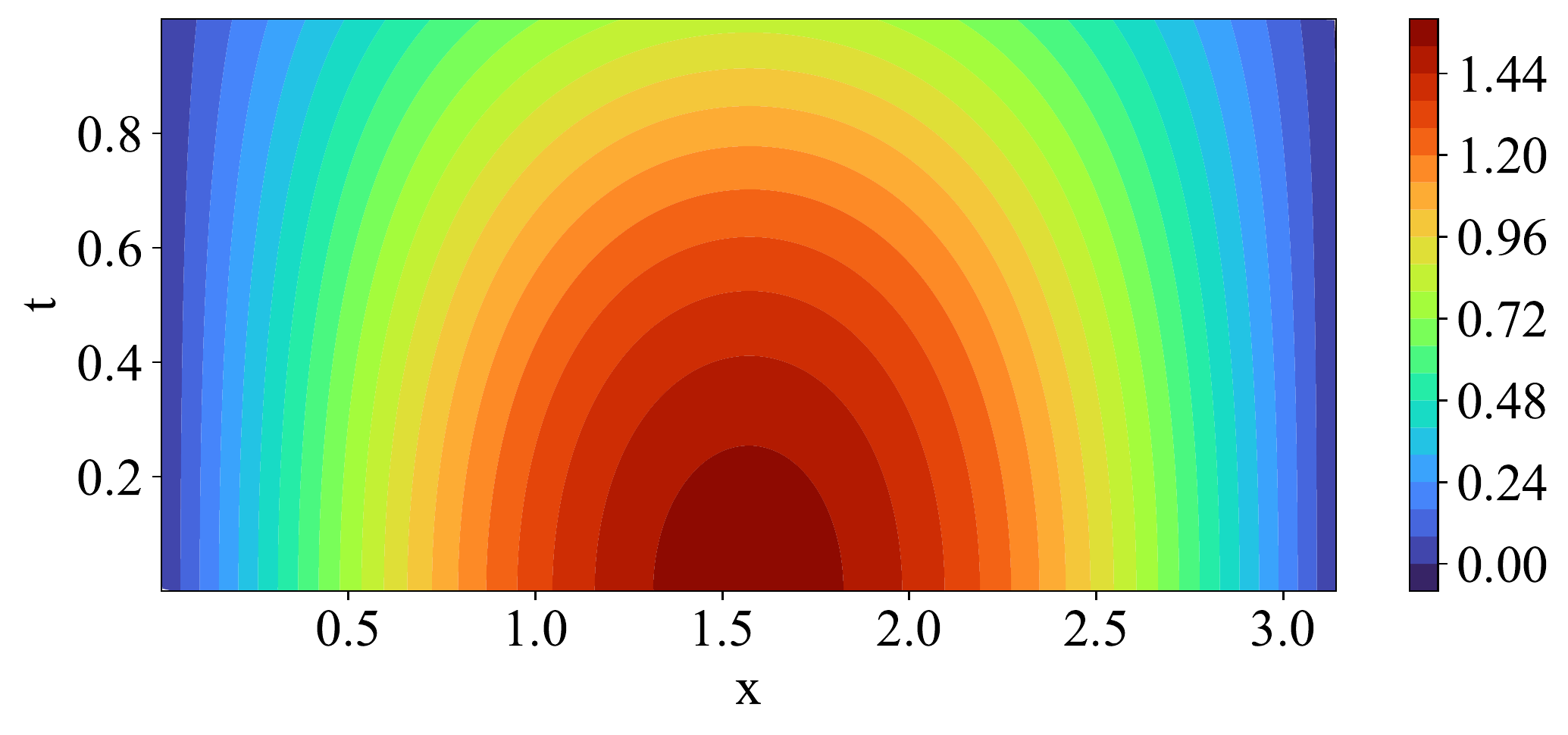}
\caption{Timoshenko single beam;  Color bar represents \textbf{Left:} Cross-sectional rotation ($\theta^\mathrm{*}$); \textbf{Right:} Transverse displacement ($w^\mathrm{*}$). } 
\label{fig5}
\end{figure*}
\begin{multline}
    \mathcal{L^{'}}(\theta) = \mathrm{\underset{\theta }{Min}}(\frac{1}{N_\mathrm{train}}\sum_{n=1}^{ N_\mathrm{train}}||\mathcal{K}(x_\mathrm{n}, t_\mathrm{n})||^2 + \\ 
    \frac{1}{N_\mathrm{data}}\sum_{n=1}^{ N_\mathrm{data}}|| u_\mathrm{data}(x_\mathrm{n}, t_\mathrm{n}) - u_\mathrm{pred}(x_\mathrm{n}, t_\mathrm{n})||^2)
\label{eq11}
\end{multline}

Here, $u_\mathrm{{pred}}$ denotes the prediction of \textit{u} by the neural network section implementing the PINN algorithm for forward and inverse problems of dimensionless beam equations.

\section{Numerical EXPERIMENTS AND DISCUSSION}
In the following subsections, five numerical experiments are presented. The experiments are conducted in a progressive manner, beginning with simple models such as a single beam system and then progressing to more complex ones such as a double beam connected to a Winkler foundation. To verify the proposed method, we first investigate forward and inverse problems for a single beam, which serves as the proof of the concept. Then, we apply the method to more intricate cases of double-beam systems to simulate forward and inverse problems.

\subsection{Timoshenko Beam Forward Problem}
The Euler-Bernoulli theory of beams is widely used in the literature and has been successfully applied in structures such as the Eiffel Tower and Ferris wheels. However, it does not consider the effects of transverse shear deformations, which are often significant in the vertical displacements of short and thick beams\cite{goerguelue2009beam}. Timoshenko beam theory provides a mathematical framework for analyzing thick-beam bending\cite{goerguelue2009beam}. According to Timoshenko theory, upon the action of an external force, the beam undergoes some cross-sectional rotation in addition to transverse displacement. Mathematically, the dynamics are modeled by a coupled system of PDEs with two variables: transverse displacement and cross-sectional rotation. The model is given by
\begin{equation}
\begin{aligned}
    \rho I \bar{\theta}_\mathrm{\bar{t}\bar{t}} - EI\bar{\theta}_\mathrm{\bar{x}\bar{x}} - kAG(\bar{w}_{\bar{x}}-\mathrm{\bar{\theta}}) = 0\\
    \rho A \bar{w}_\mathrm{\bar{t}\bar{t}}-kAG(\bar{w}_\mathrm{\bar{x}\bar{x}}-\bar{\theta}_\mathrm{\bar{x}}) = \bar{g}(\bar{x},\bar{t})
\label{eq12}
\end{aligned}
\end{equation}
where $\rho$, $A$, $E$ and $I$ have the usual meaning as in the case of the Euler-Bernoulli beam; $k$ is called the Timoshenko shear coefficient; $G$ is the shear modulus; and $\bar{g}(\bar{x},\bar{t})$ is the external force acting on the beam. The transverse displacement is $\bar{w}(\bar{x},\bar{t})$ and $\bar{\theta} (\bar{x}, \bar{t})$ is the cross-sectional rotation of the beam at position $\bar{x}$ and time $\bar{t}$. After nondimensionalizing~\eqref{eq12} and taking the resulting parameters \cite{akellavisualization} to be unity, the nondimensional equation can be written as follows:

\begin{equation}
\begin{aligned}
 \theta_\mathrm{tt} - \theta_\mathrm{xx} + (\theta - w_\mathrm{x}) = 0     \\
w_\mathrm{tt} + (\theta - w_\mathrm{x})_\mathrm{x} = g(x, t) 
\label{eq13}
\end{aligned}
\end{equation}

We consider the external force \cite{semper1994semi} to be $g(x, t)$ = $\cos(t)-\frac{\pi}{2}\sin(x)\cos(t)$ and the computational domain to be $x \in [0, \pi]$ and $t \in [0, 1]$. To make~\eqref{eq13} well-posed, the initial and boundary conditions are supplemented as:

\begin{figure*} 
\centering
\includegraphics[width=1\columnwidth]{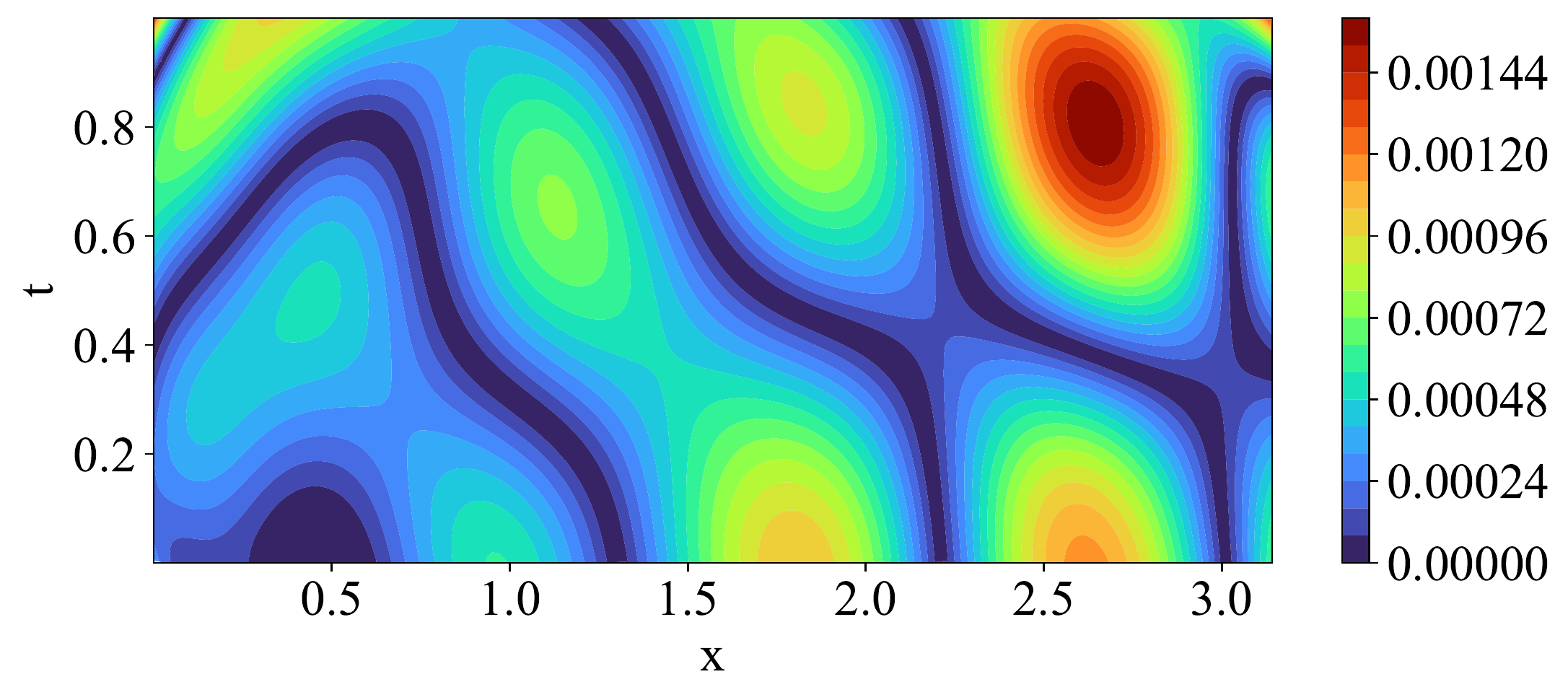}
\includegraphics[width=1\columnwidth]{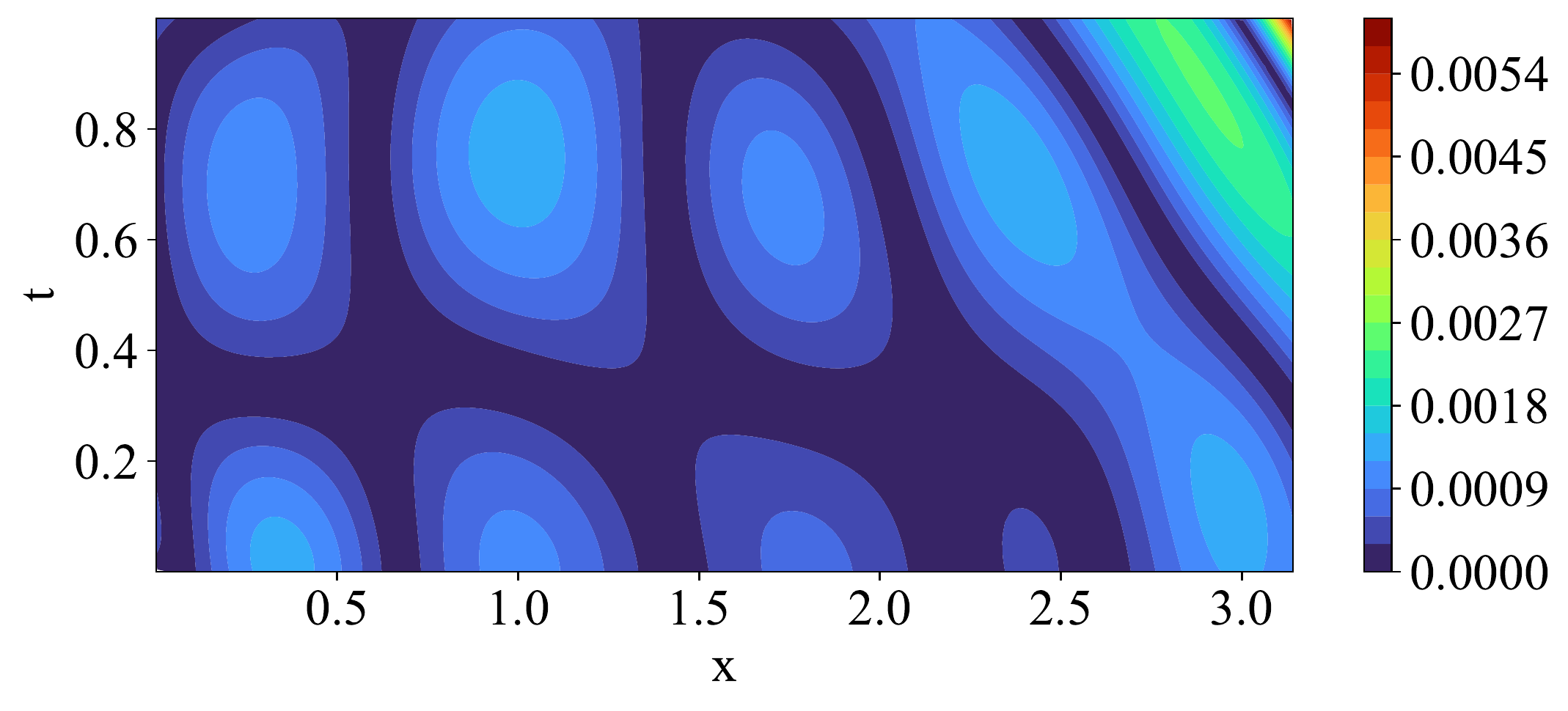}
\caption{Timoshenko single beam absolute error in predictions \textbf{Left:} $|\theta - \theta^\mathrm{*}|$; \textbf{Right:} Absolute error $|w - w^\mathrm{*}|$.} 
\label{fig6}
\end{figure*}

\begin{equation*}
    \theta(x, 0) = \frac{\pi}{2}\cos(x) + \left(x - \frac{\pi}{2}\right), \quad \theta_t(x, 0) = 0
\end{equation*}
\begin{equation*}
    w(x, 0) = \frac{\pi}{2}\sin(x),\quad w_t(x, 0) = 0
\end{equation*}
\begin{equation*}
    \theta(0, t) = \theta(\pi, t) = w(0, t) = w(\pi, t) = 0
\end{equation*}
 
To estimate the error in the approximated solutions, the analytical solution for the considered problem is used, which is
\begin{equation*}
    \theta(x, t)= \left(\frac{\pi}{2}\cos(x) + \left(x - \frac{\pi}{2}\right)\right)\cos(t)
\end{equation*}
\begin{equation*}
    w(x, t) = \frac{\pi}{2}\sin(x)\cos(t)
\end{equation*}

 When analytical solutions are not available, there are various ways to validate the PINN solution. One approach is to compare the solutions with those obtained using numerical methods such as finite difference, finite element, finite volume or spectral methods. This can be done by comparing the predicted solutions from the PINNs with the solutions from the numerical simulation for the same physical equation. Another approach is to compare the solutions obtained through PINNs with experimental data. One can compare the predicted solutions from the PINNs with values experimentally measured over space and time. Finally, one can validate the solutions obtained through PINNs by checking if they satisfy the known physical constraints of the system. In summary, one can use available experimental data, numerical methods or physical constraints to evaluate the accuracy of the solution obtained using PINNs.

The difficulty of solving a system of PDEs is greater than that solving a single PDE, but the neural network structure used for the Euler-Bernoulli equation is successful in approximating solutions for Timoshenko beams. In particular, the transverse displacement of the beam is computed within $\mathcal{R} = 3.3e-4$ percent, and the cross-sectional rotation is approximated within $\mathcal{R} = 2.8e-3$ percent. Approximated solutions and absolute errors in predicting the transverse displacement and cross-sectional rotation are presented in Figs.~\ref{fig5} and ~\ref{fig6}. Fig.~\ref{fig5} demonstrates that when a sinusoidal force is applied to a Timoshenko beam, the beam bends more than it rotates. As indicated by the scale in the figures, the maximum deflection is $1.44$ and the maximum rotation is $0.32$. Additionally, the low error in predictions demonstrates that even with the increase in the PDE complexity, the PINN successfully solves the Timoshenko PDE with comparable results to the Euler-Bernoulli equation.

 We compare the results obtained from our method with three other methods. The first method we consider is the widely used numerical technique called the finite difference method (FDM). The other two methods are neural network-based approaches, namely physics-guided neural networks (PGNN) \cite{7959606, yu2020structural, zhang2021structural, karpatne2017physics, khandelwal2020physics} and gradient-enhanced physics-informed neural networks (gPINN) \cite{yu2022gradient}. First, for FDM we employ a central difference scheme to approximate space derivatives and a leapfrog scheme to approximate time derivatives. This approach allows us to solve problems with second-order accuracy in space and time. The results for the Timoshenko beam show that PINNs can achieve a higher level of accuracy than the FDM even with a smaller number of training points. Specifically, $30,000$ points are used in the FDM scheme while only $16,000$ points were used for training with PINNs and Table ~\ref{tab:1} indicates that PINNs perform better than FDM.

Second, the performance of PINN is compared to a neural network-based approach PGNN, which leverages physical knowledge embedded in the available data, for instance, the relationship between beam acceleration and displacement for the Timoshenko beam problem. Accelerometers can be employed at discrete locations along the beam to obtain acceleration data. Acceleration data at five equidistant points along the beam are used, with $2000$ data points at each location. This dataset is augmented with the boundary and initial conditions of displacement to match the training data size of PINN. PGNN is a deep neural network-based architecture with inputs: position ($x$), time ($t$), and acceleration. Displacement ($w$) is taken as the output of this neural network. Training PGNN with identical hyperparameters to those used in PINN, PGNN predicts the displacement ($w$) with an error of approximately $0.002739$$\%$, as shown in Table ~\ref{tab:1}.

Furthermore, utilizing the displacement values ($w$), the neural network's auto differentiation and~\eqref{eq13}, we derived $\theta_\mathrm{x}$. Subsequently, a second neural network was constructed to predict $\theta$, where $\theta_\mathrm{x}$ is used as the input. Boundary and initial conditions for cross sectional rotation ($\theta$) are also used to guide the PGNN towards the optimal solution. After training the PGNN, cross-sectional rotation is predicted with approximately $3.486727$$\%$ error. It can be inferred from Table ~\ref{tab:1} that both displacement and rotation predictions exhibited higher errors than PINN. This discrepancy can be attributed to the restricted availability of acceleration data at only discrete spatial locations within the interior domain rather than a random distribution across the entire domain. Furthermore, the second neural network, employed for rotation prediction, demonstrated inferior performance potentially due to error propagation.

Third, we perform another comparison with a neural network-based method to simulate PDEs, gradient-enhanced PINN (gPINN) \cite{yu2022gradient}, which differs from PINN in terms of the loss function. The acronym "gPINN" proposed in \cite{yu2022gradient} is used in this work instead of "GPINN" as it is used for another method \cite{miao2023gpinn}. In addition to the loss function of PINN, gPINN leverages gradient information of the PDE residual and embed the gradient into the loss function. For the Timoshenko beam problem, derivatives of the system of PDE~\eqref{eq13} with respect to space ($x$) and time ($t$) are supplemented in the loss function. Table~\ref{tab:1} shows that gPINN exhibits higher relative error percentages in learning displacement and cross-sectional rotation than PINN. The high-order derivatives of the physical equations in the loss function of gPINN make it challenging for autodifferentiation \cite{bettencourt2019taylor} and backpropagation of the loss function, resulting in poor predictions of deflection and rotation for the Timoshenko beam. Table ~\ref{tab:1} demonstrates that PINN outperforms FDM, PGNN, and gPINN in accurately predicting displacement and cross-sectional rotation for the Timoshenko beam, emphasizing its superior performance compared to the three alternative methods.

\begin{table}
\caption{Timoshenko beam: $\mathcal{R}$ at $t=1$}
    \begin{center}
    \begin{tabular}{|c|c|c|c|c|}\hline
     
       $w$, $\theta$  & PINN & FDM &  PGNN & gPINN\\  \hline
 $w$ (\%) & 3.3e-4 & 0.005615 & 0.002739 & 0.249849  \\ 
 \hline
  $\theta$ (\%) & 2.8e-3 & 0.004733 & 3.486727 &  5.498449  \\ 
 \hline 
    \end{tabular}
    \label{tab:1}
    \end{center}
\end{table}

     

     



\subsection{Timoshenko Beam Inverse Problem}
This section addresses the inverse problem for the Timoshenko beam, with the aim to determine the material properties of a beam leveraging the PDE and beam's displacement and rotation data. In structural engineering, the inverse problem of a Timoshenko beam PDE is significant for determining the beam system's structural behavior and for health monitoring. This helps engineers infer the internal material properties and unknown forces from observed responses such as displacement and rotation measurements. The PINN solves this problem by combining the knowledge of physics and deep learning. The PINN uses a neural network to learn the mapping between the unknown parameters of the PDE and observed data while incorporating the constraints of physics in the form of PDEs. This parameter identification aids in providing crucial information for structural diagnosis and repair and helps engineers ensure the safety and stability of structures. The Timoshenko model for parameter estimation is presented as follows.
\begin{figure}[htbp]
\centerline{\includegraphics[width=0.8\columnwidth]{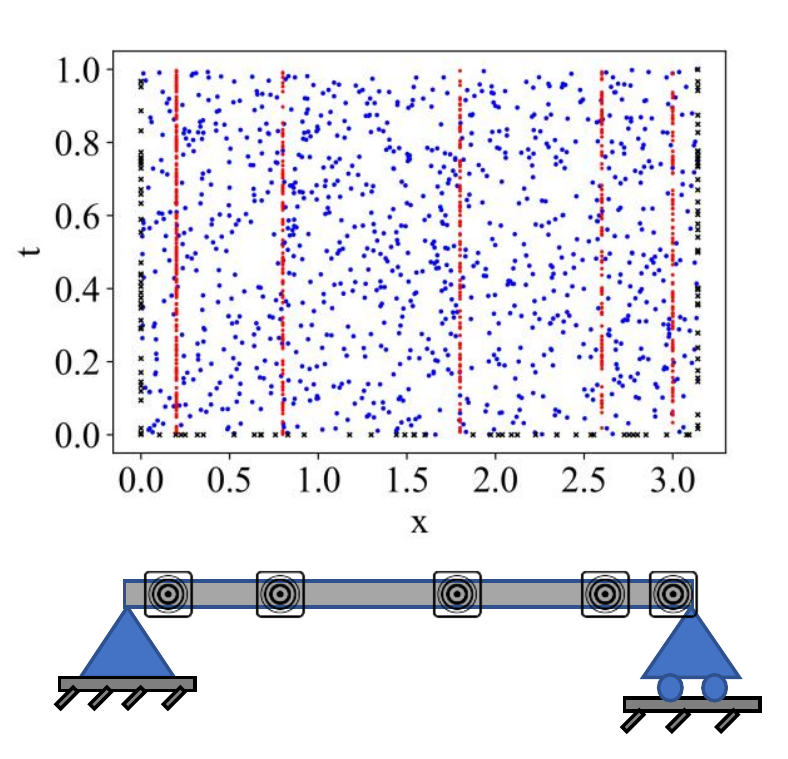}}
\caption{Data to learn the parameters for the Timoshenko single-beam: \textbf{Blue dots} Collocation points. \textbf{Red dots} Additional data points {{of rotations ($\theta$) and displacement (u)}}. \textbf{Black dots} Initial and boundary points.}
\label{fig7}
\end{figure}

\begin{equation}
\begin{aligned}
\alpha \theta_\mathrm{tt} - \theta_\mathrm{xx} + (\theta - w_\mathrm{x}) = 0     \\
w_\mathrm{tt} + (\theta - w_\mathrm{x})_\mathrm{x} = g(x, t) 
\label{eq14}
\end{aligned}
\end{equation}

In the context of the inverse problem of the Timoshenko beam, the PINN is trained on the observed deflections and rotations of the beam, and the material properties are treated as the unknowns to be estimated. In this case, the force g(x, t) applied to the beam is considered to be known, and the only unknown in the model is $\alpha$. This makes the problem ill-posed, requiring additional data at a priori to predict the unknown parameter. For $\alpha=1$, the transverse displacement and cross-sectional rotation data obtained from the forward problem is supplied to approximate the parameter value. This data is not error-free and comes with $10^{-3}$ percent error for transverse displacement and with $10^{-4}$ percent error for cross-sectional rotation. As shown in Fig. ~\ref{fig7} the additional data is supplied on $5000$ points (red dots) at five positions on the beam $(x = 0.2, 0.8, 1.8, 2.6, 3)$. In practice, this data can be
collected using sensors installed at the corresponding locations on the beam as shown in Fig. ~\ref{fig7}.
    
To solve the inverse problem, the neural network consist of $1600$ random training points with the distribution $N_{\mathrm{i}} = 200$, $N_{\mathrm{b}} = 400$, and $N_{\mathrm{int}}=1000$. To regularize the PDE term in the loss function, a regularization parameter of $1$ was chosen \cite{raissi2019physics}. Using the L-BFGS optimizer $5000$ iterations are performed and the other parameters are kept the same as in the forward Timoshenko problem. At $t = 0.5$, the unknown parameter $\alpha = 1.0136$ is learned. 

\begin{figure*} 
\centering
\includegraphics[width=1\columnwidth]{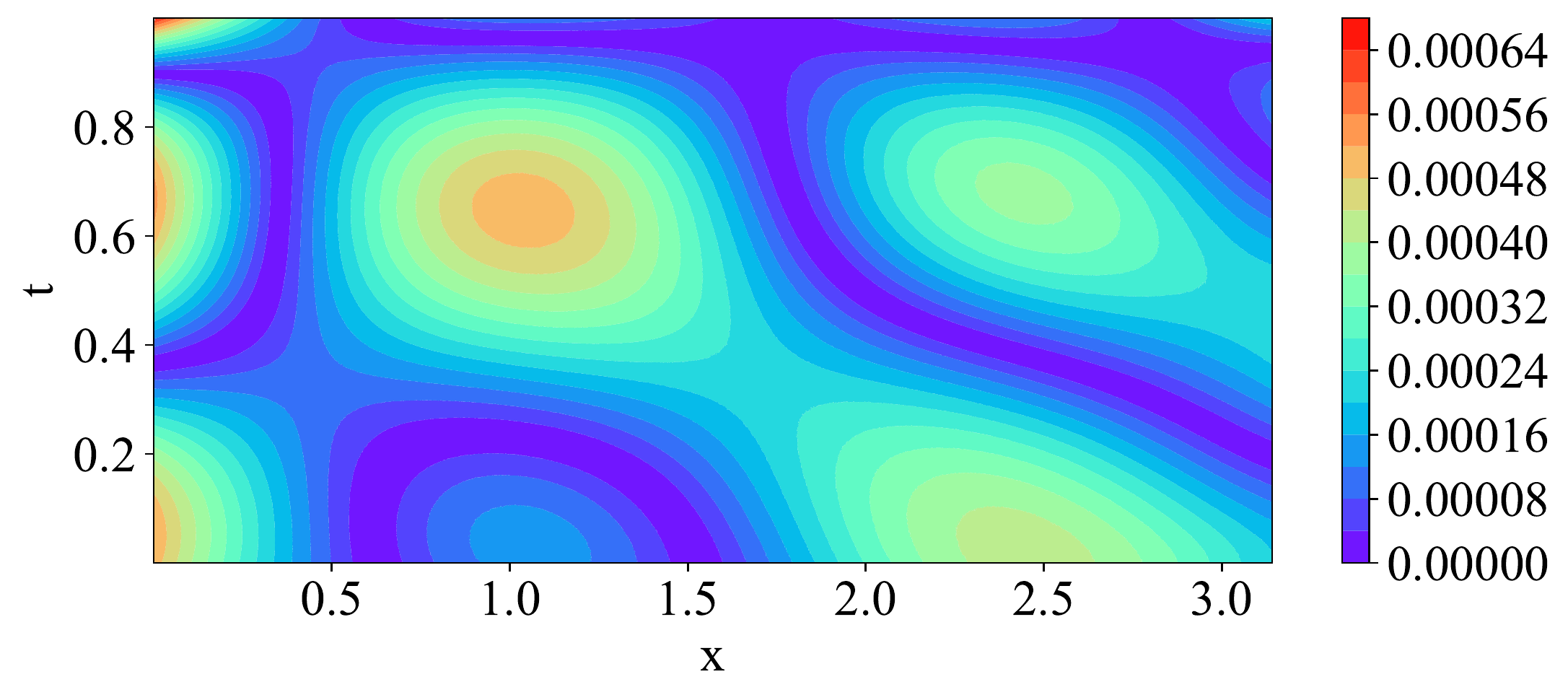}
\includegraphics[width=1\columnwidth]{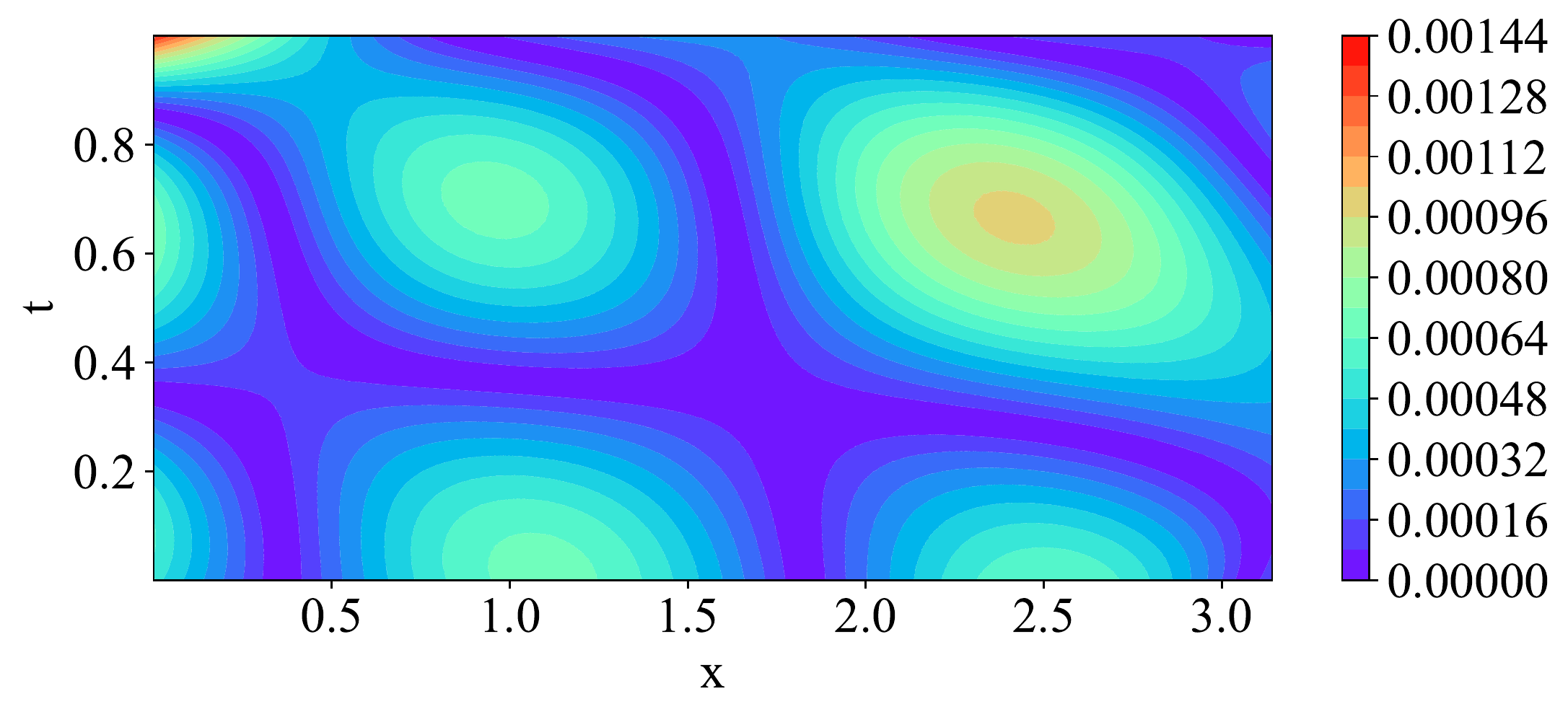}
\caption{Euler-Bernoulli double-beam: color bar represents absolute error in predictions \textbf{Left:} $|w_1 - w_1^\mathrm{*}|$; \textbf{Right:} $|w_2 - w_2^\mathrm{*}|$.}
\label{fig8}
\end{figure*}

\begin{figure*} 
\centering
\includegraphics[width=0.6\columnwidth]{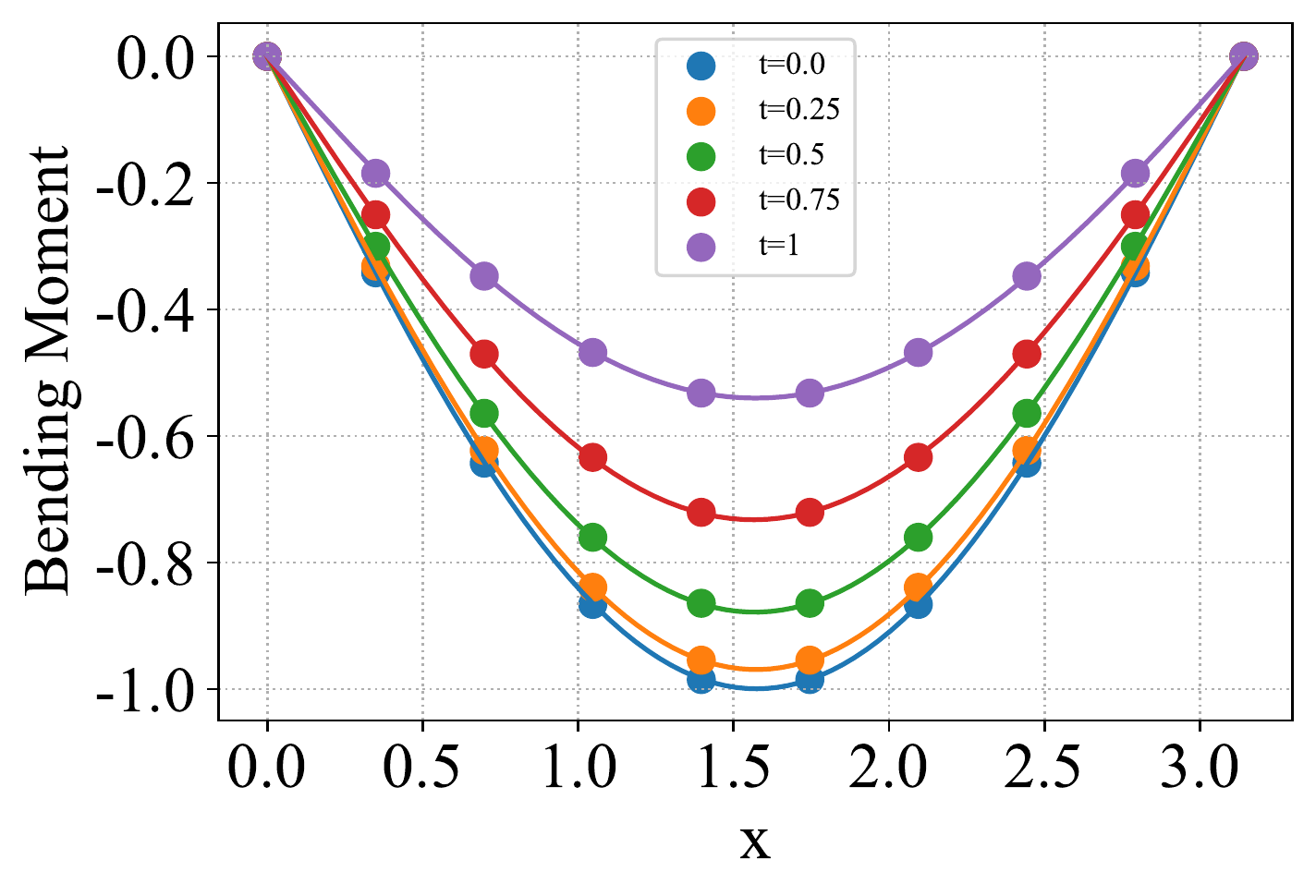}
\includegraphics[width=0.6\columnwidth]{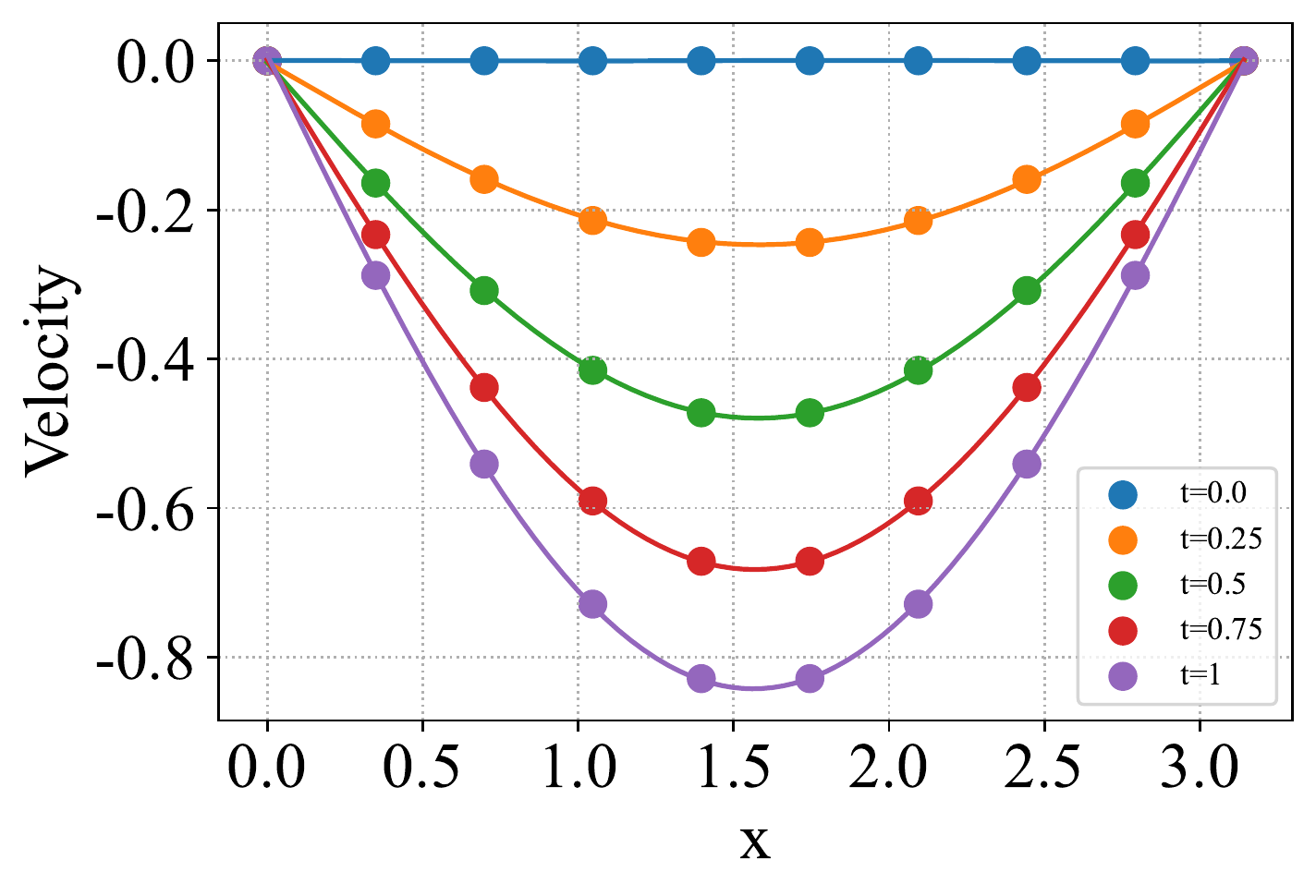}
\includegraphics[width=0.6\columnwidth]{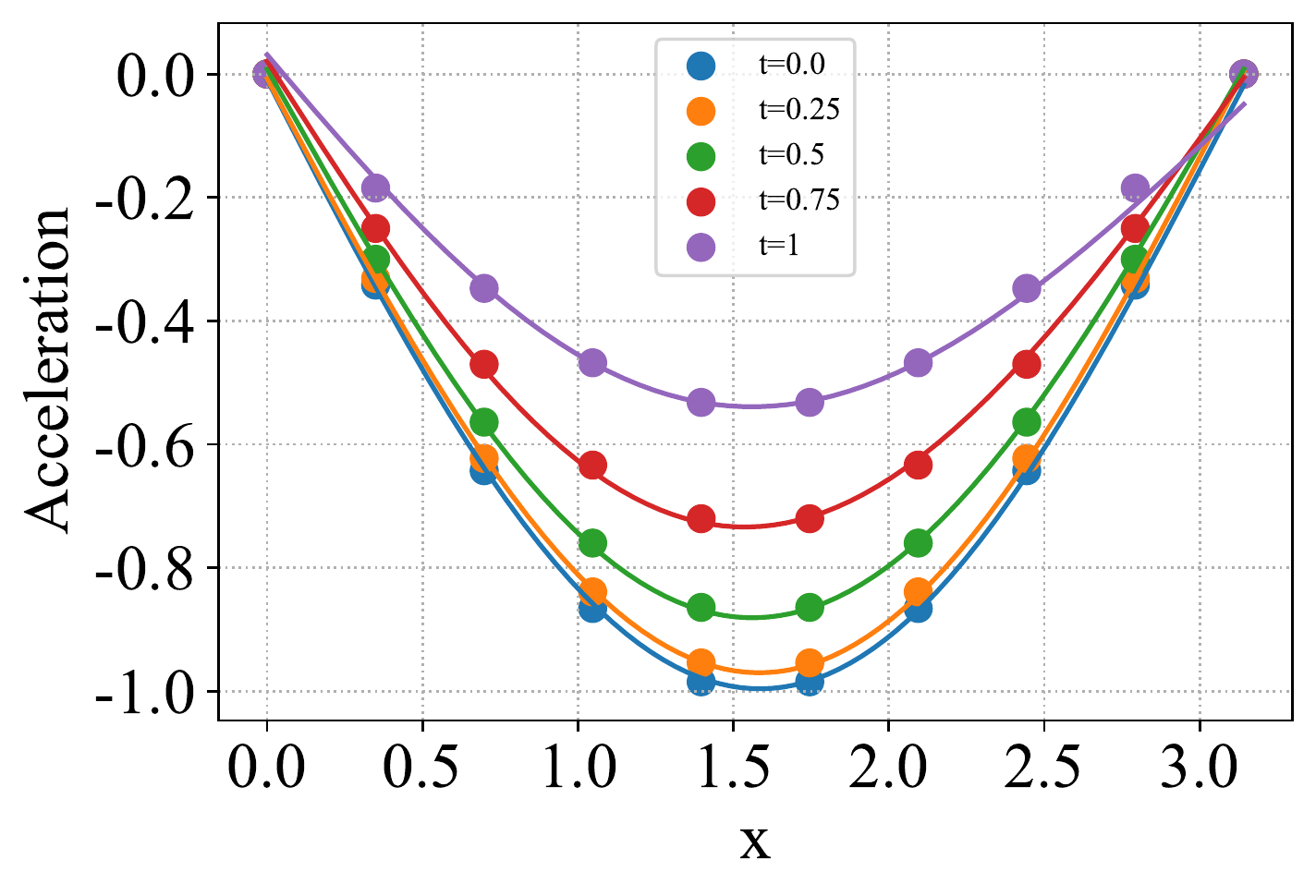}
\includegraphics[width=0.6\columnwidth]{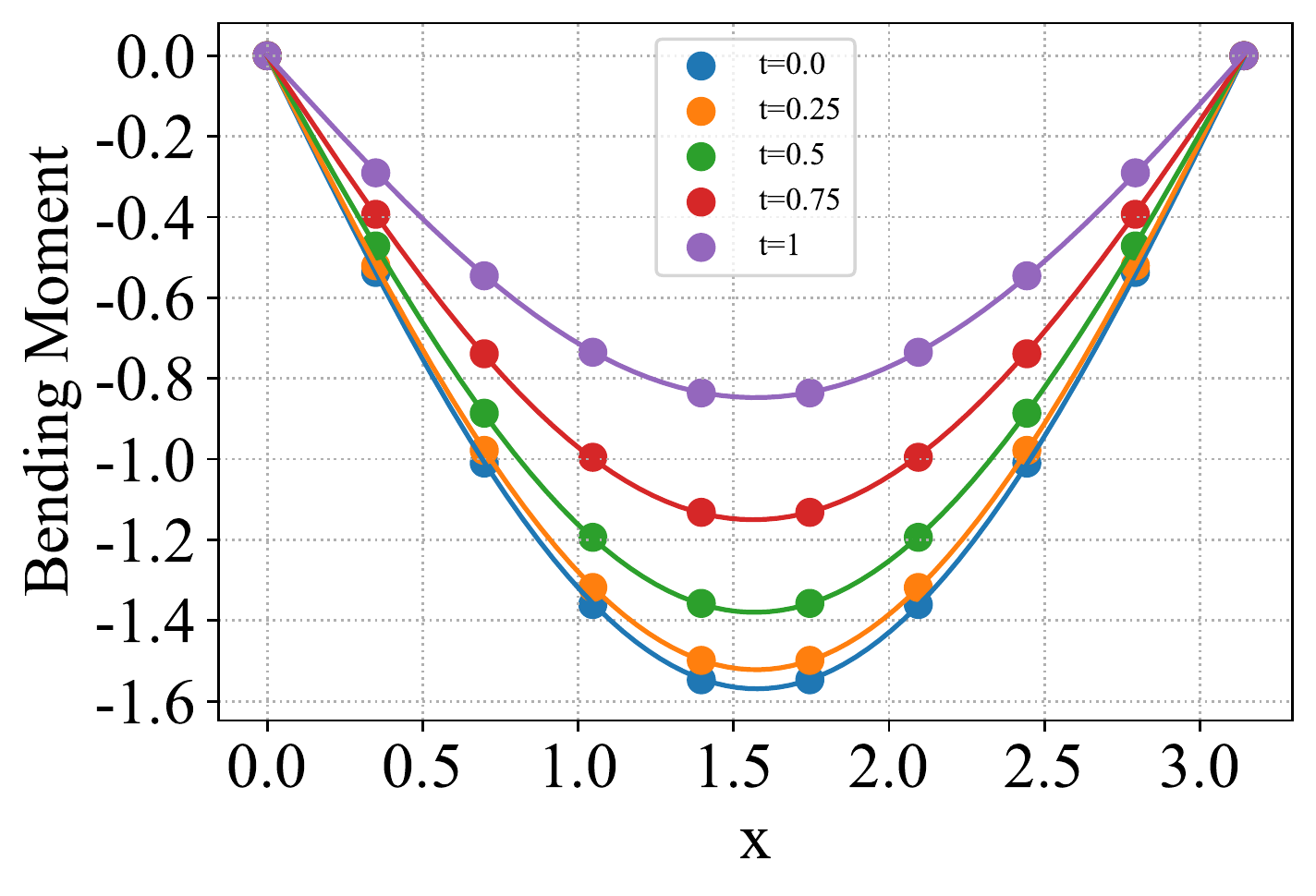}
\includegraphics[width=0.6\columnwidth]{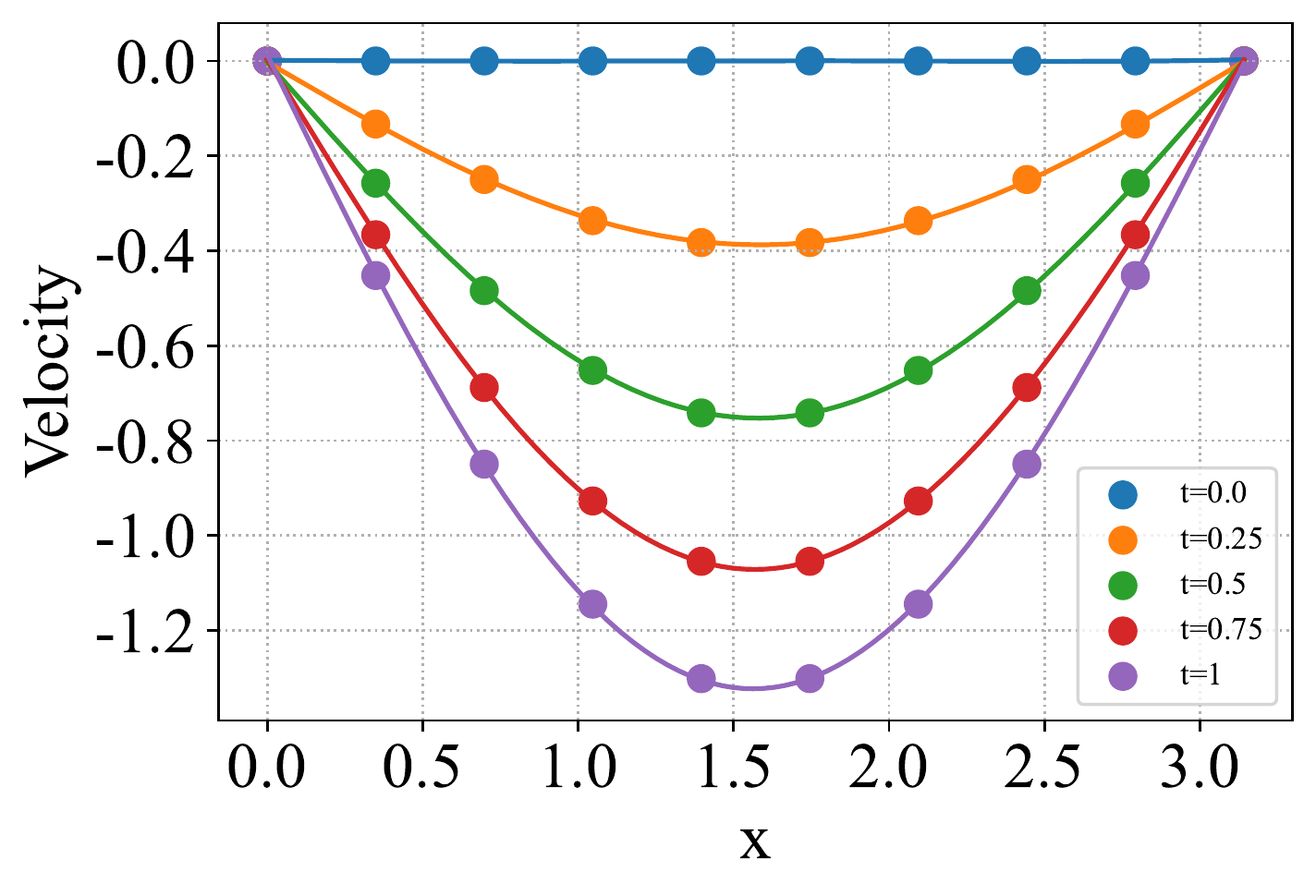}
\includegraphics[width=0.6\columnwidth]{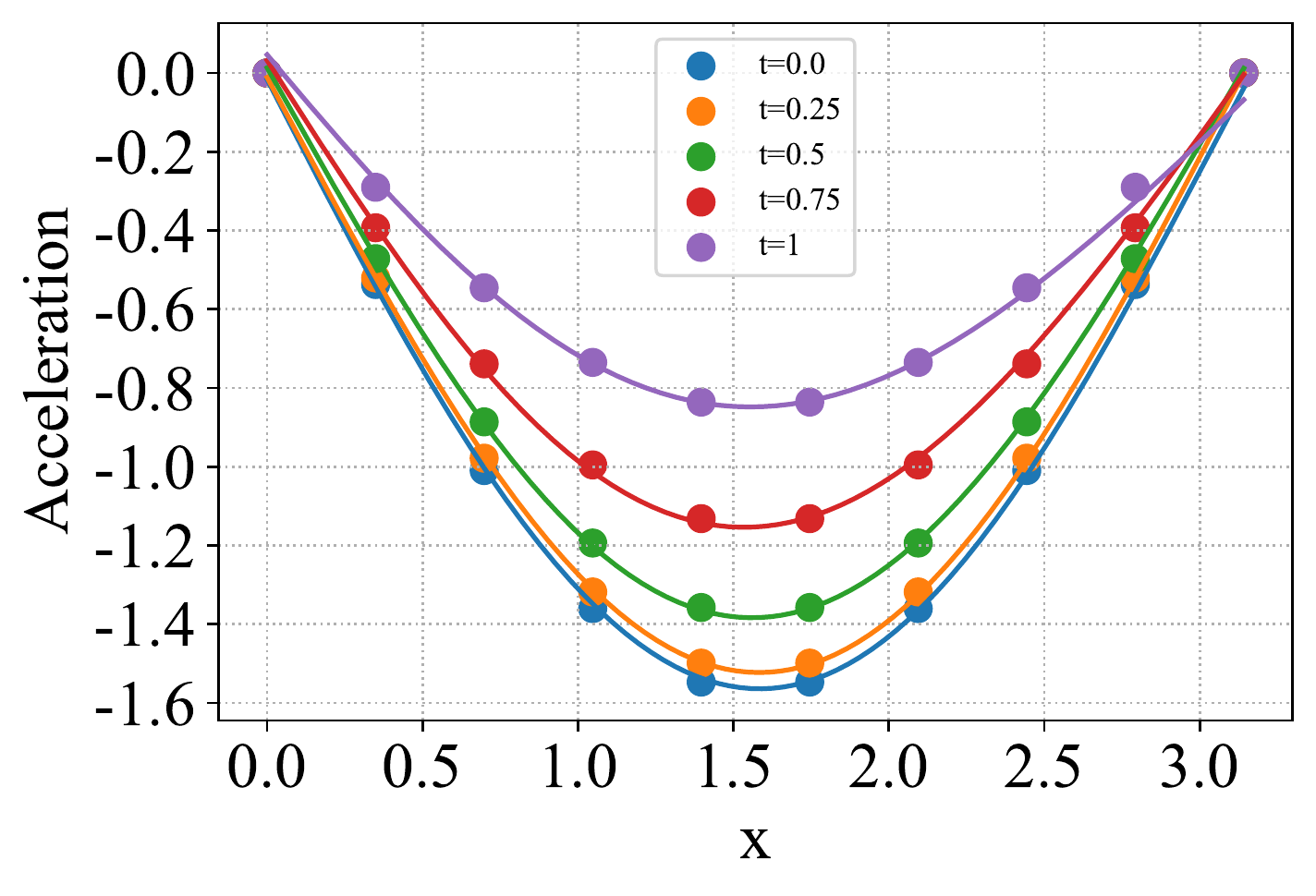}
\caption{Derived quantities for the Euler-Bernoulli double beam. Scattered points represent the exact solution and the continuous line refers to the derived solution. \textbf{Top:} First beam \textbf{Left} Bending moment; \textbf{Mid} Velocity; \textbf{Right} Acceleration. \textbf{Bottom:} Second beam \textbf{Left} Bending moment; \textbf{Mid} Velocity; \textbf{Right} Acceleration.}
\label{fig9}
\end{figure*}

\begin{figure}[htbp]
\centerline{\includegraphics[width=1\columnwidth]{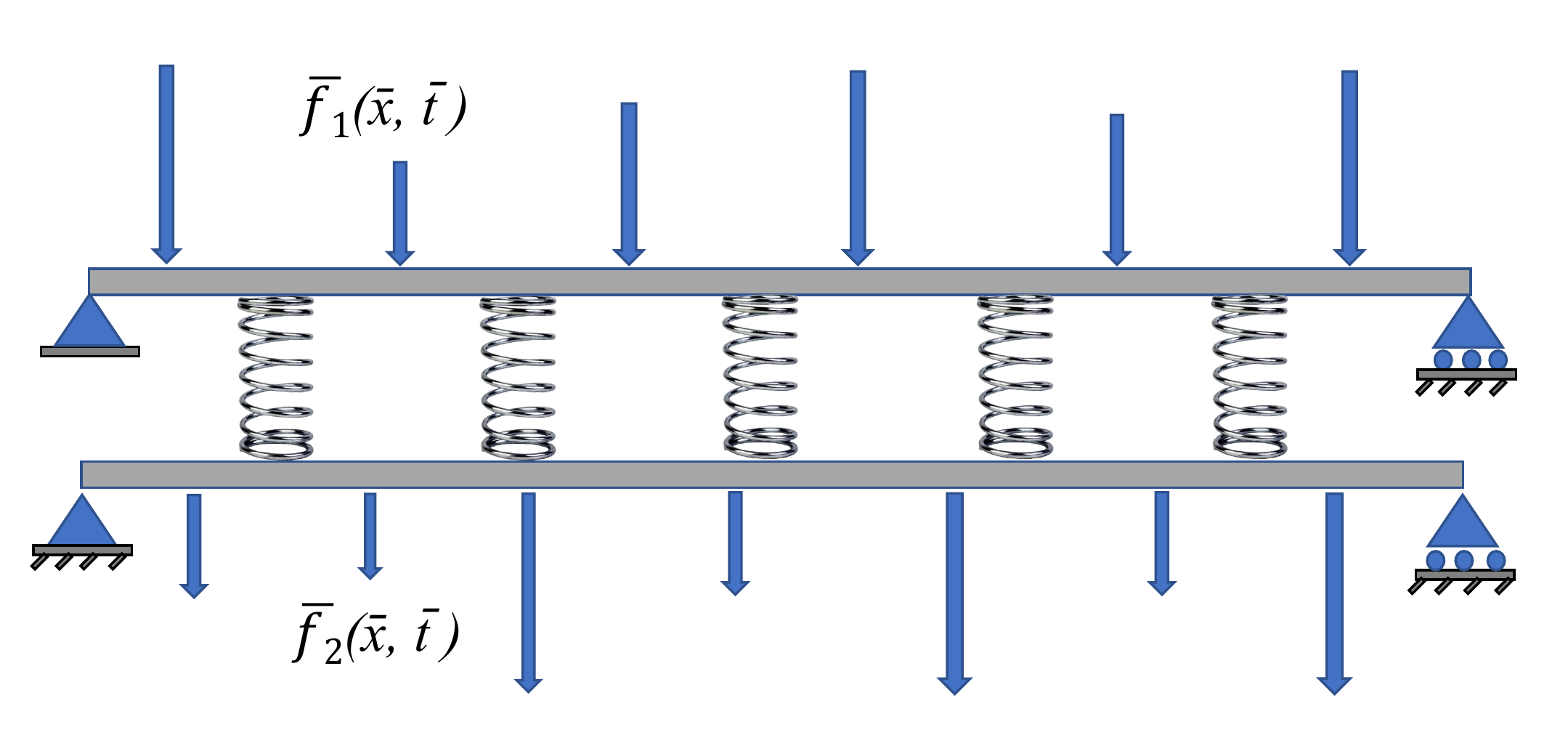}}
\caption{Double beam system connected by a Winkler foundation.}
\label{fig10}
\end{figure}

We perform a comparison between the PINN and DNNs, as using a numerical iterative method for inverse problems is computationally expensive. From PINNs, at $t = 0.5$, the unknown parameter $\alpha = 1.0136$ is learned.  We utilize DNNs to identify the parameters of a Timoshenko single beam. We use the same architecture for DNN as used by the PINN. The predicted value of alpha is $0.6124$ using DNN. PINN is more accurate than DNNs for the inverse problem of beam systems. 

However, there are several issues that one may need to take care of while solving inverse problems through the presented framework. First, to avoid overfitting, the minimum training data points required to solve the problem should be determined empirically by gradually increasing the number of training points until the model’s performance is satisfactory. Second, for some physical problems, noisy data may lead to nonconvergence of the optimization algorithm. Hence, suitable filtering or preprocessing of data may be required before using the PINN framework. Finally, for every run of the neural network, one may learn a different parameter or function value; due to the convergence of the optimizers at different local minima, it may be useful to find the statistics of the inverse problem solution through multiple runs.

 Experimental results for single beam equations illustrate that PINNs can efficiently solve forward and inverse problems for single beams. In this study, we investigate the ability of PINNs to handle more complex systems, specifically double-beam systems connected by a Winkler foundation, as depicted in Fig.~\ref{fig10}.

\subsection{Euler-Bernoulli Double-Beam Forward Problem}

In this section, and for all further experiments, forced transverse vibrations of two parallel beams are studied. Structurally, two parallel beams of equal lengths joined by a Winkler massless foundation are considered. Both beams are considered slender and have homogeneous material properties. The transverse displacement of both beams is governed by the following system of PDEs \cite{onis2}:

\begin{equation}
\begin{aligned}
m_{1}\bar{w}_{1_\mathrm{\bar{t}\bar{t}}} + K_{1}\bar{w}_{1_\mathrm{\bar{x}\bar{x}\bar{x}\bar{x}}} + k(\bar{w}_1 - \bar{w}_2) = \bar{f}_1(\bar{x}, \bar{t}) \\
m_{2}\bar{w}_{2_\mathrm{\bar{t}\bar{t}}} +  K_{2}\bar{w}_{2_\mathrm{\bar{x}\bar{x}\bar{x}\bar{x}}} + k(\bar{w}_2 - \bar{w}_1) = \bar{f}_2(\bar{x}, \bar{t})
\label{eq15}
\end{aligned}
\end{equation}

Here, $\bar{w}_1$ and $\bar{w}_2$ are the beam displacements for the first and the second beams respectively. The distributed continuous forces acting transversely on the beams are $\bar{f}_1$ and $\bar{f}_2$ as shown in Fig.~\ref{fig10}. The product of the density and the cross-sectional area of the beams is given by $m_{1} = \rho_{1}A_{1}$ for the first beam and $m_{2} = \rho_{2}A_{2}$ for the second beam. The parameters $K_1$ and $K_2$ denote the flexural rigidity of the beams and are given by $K_1 = E_1 I_1$ and $K_2 = E_2 I_2$. The stiffness modulus of the Winkler elastic layer connecting both beams is given by $k$. For simplicity, we consider $m_1 = m_2$, and $K_1 = K_2$, and nondimensionalize~\eqref{eq15}. After taking all the resulting parameters to be unity, the nondimensional equation has the same form as~\eqref{eq15} with unit coefficients. The initial conditions are

\begin{figure*} 
\centering
\includegraphics[width=0.8\columnwidth]{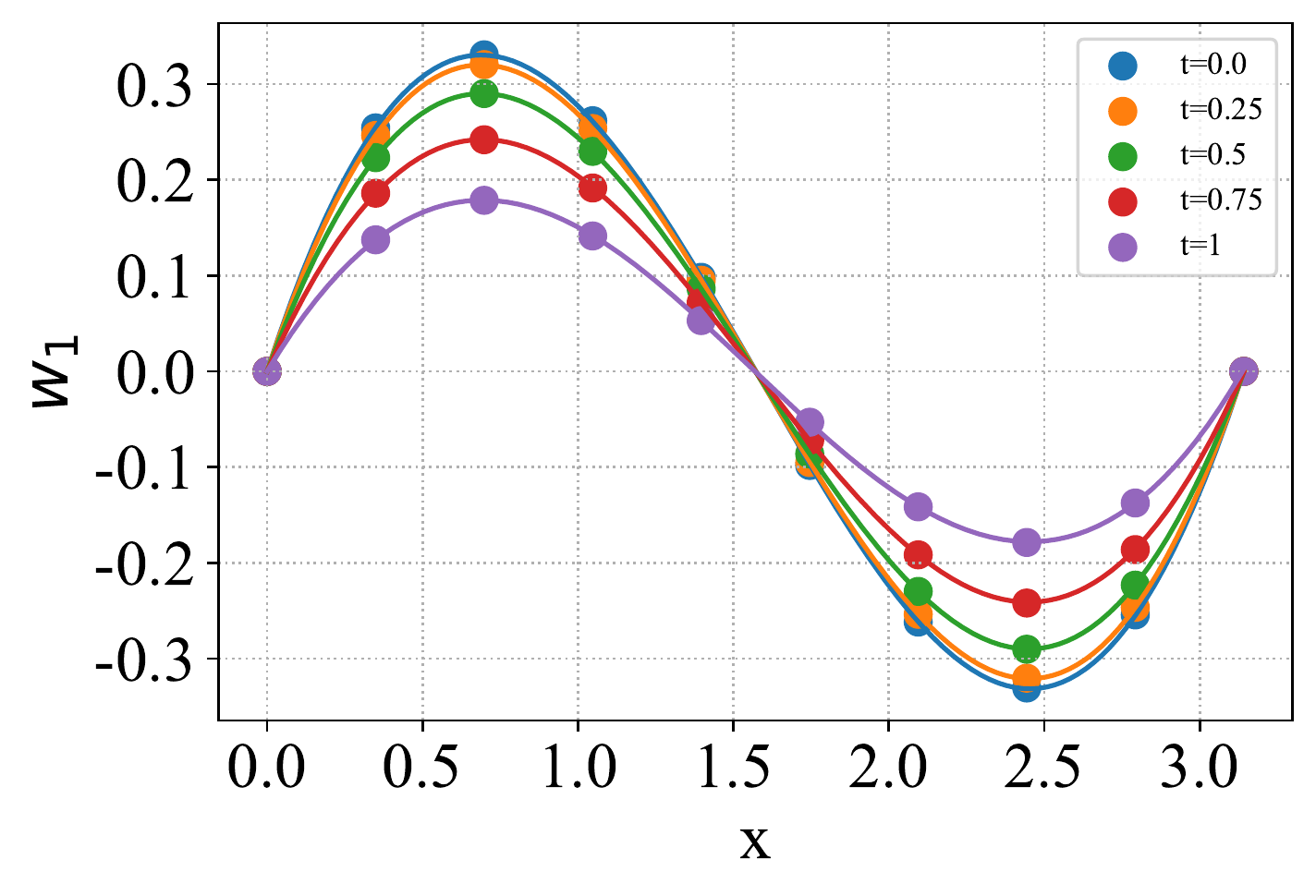}
\includegraphics[width=0.8\columnwidth]{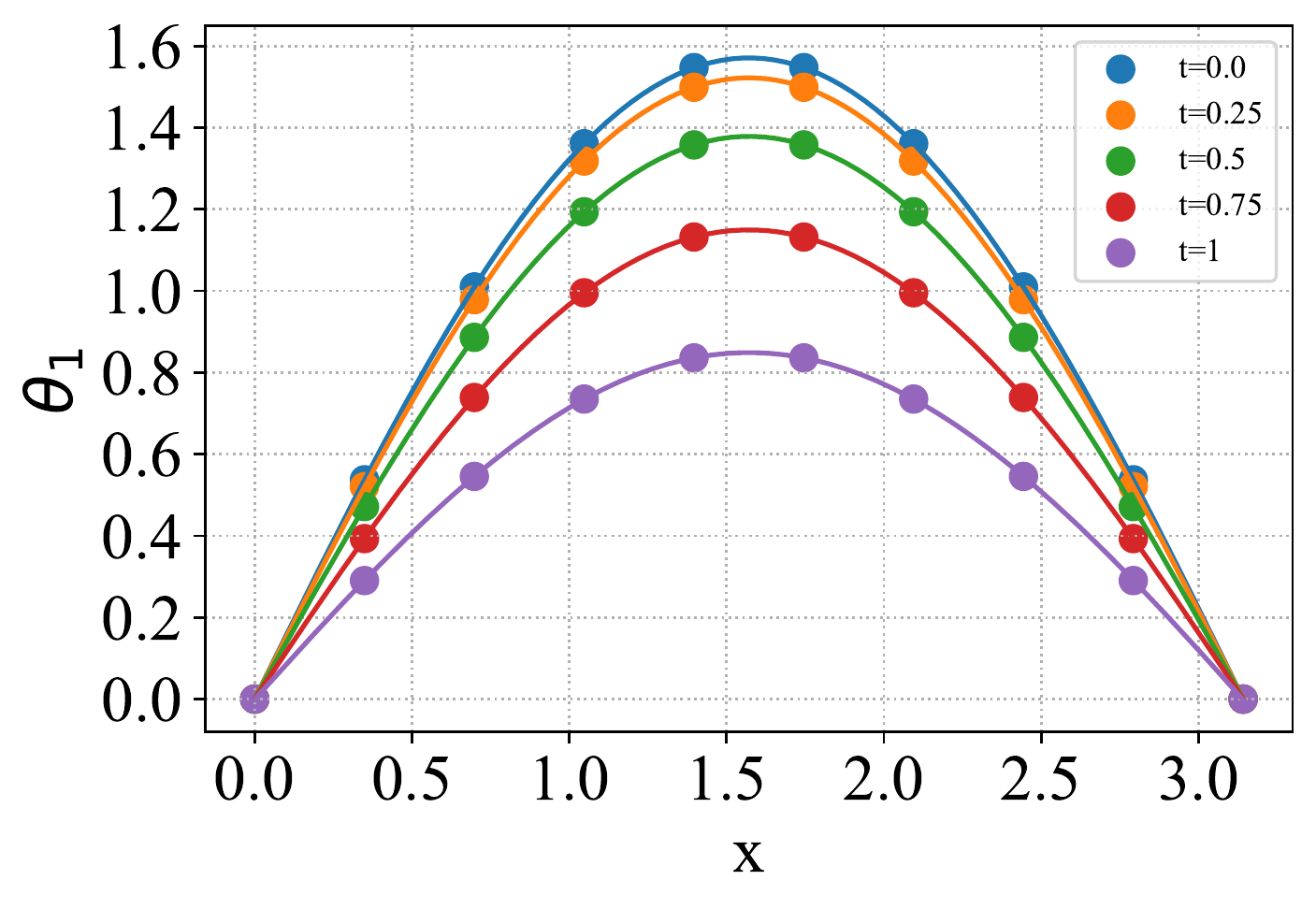}
\includegraphics[width=0.8\columnwidth]{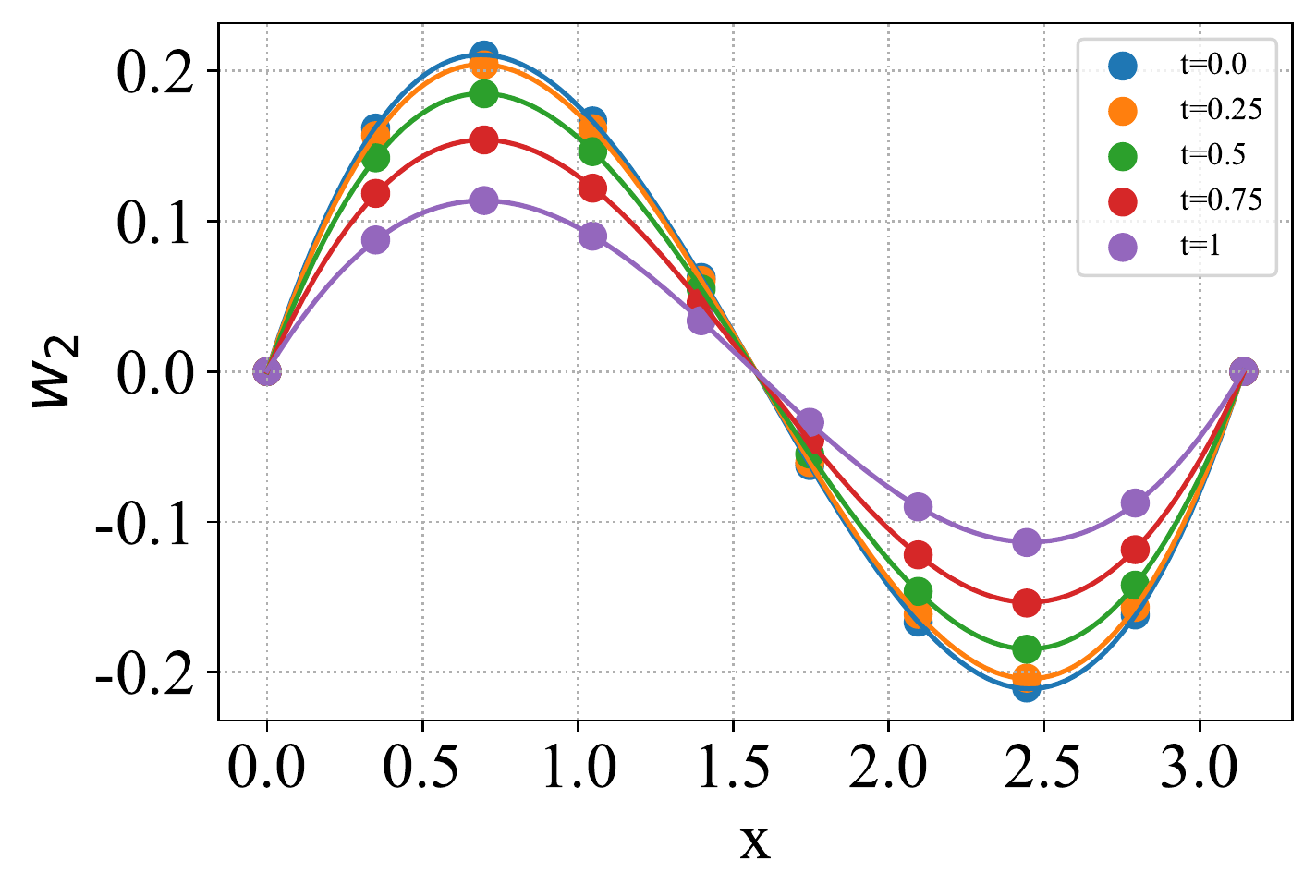}
\includegraphics[width=0.8\columnwidth]{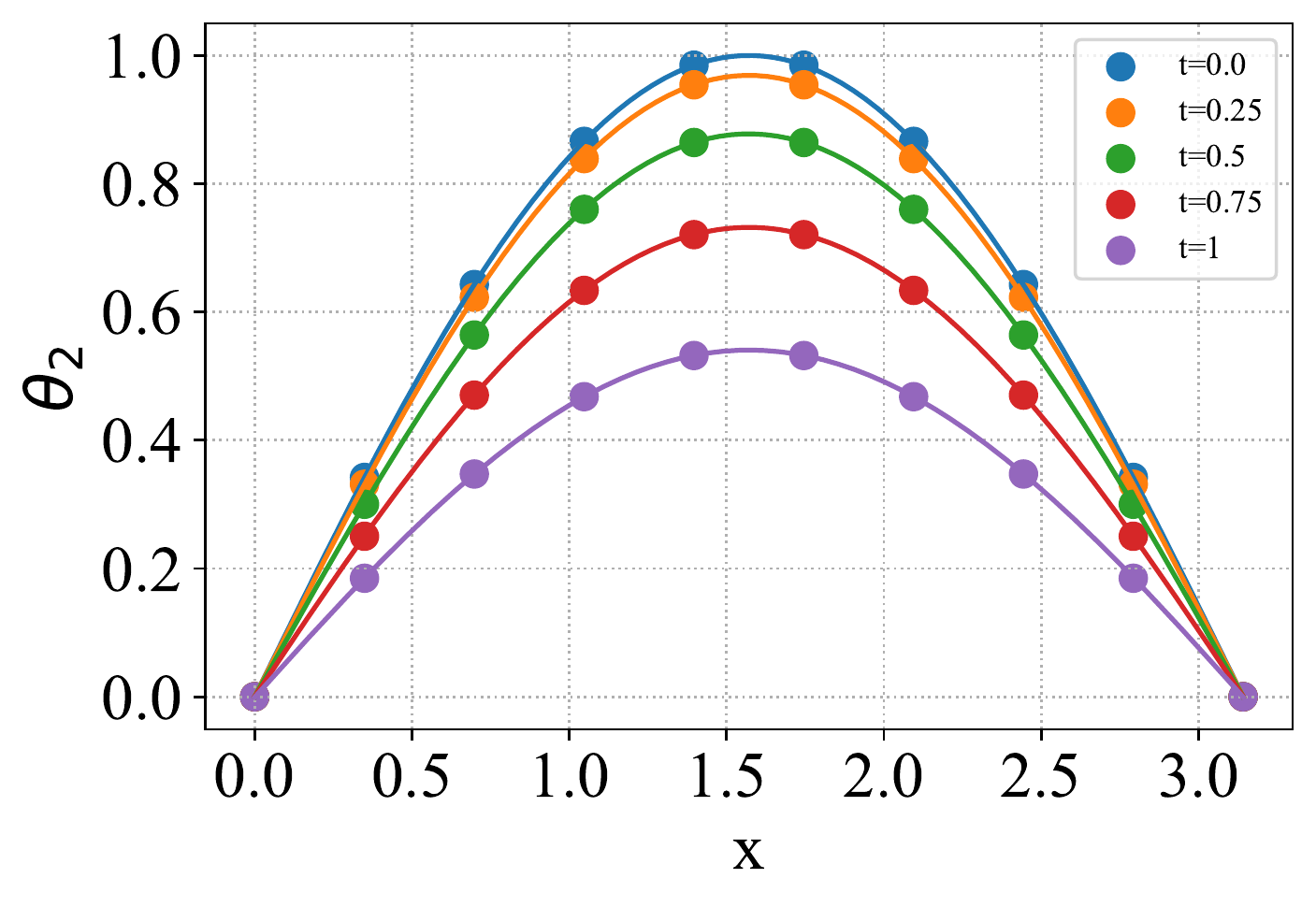}
\caption{Timoshenko double beam. Scattered points represent the exact solution, and the continuous line refers to the predicted solution. \textbf{Top:} First beam \textbf{Left} Displacement $(w_{1})$;  \textbf{Right} Rotation $(\theta_1)$. \textbf{Bottom:} Second beam \textbf{Left} Displacement $(w_2)$; \textbf{Right} Rotation $(\theta_2)$}.
\label{fig11}
\end{figure*}

\begin{equation*}
\begin{aligned}
   w_1(x, 0) = \sin(x), \quad
   w_{1_\mathrm{t}}(x, 0) = 0\\
   w_2(x, 0) = \frac{\pi}{2}\sin(x), \quad
   w_{2_\mathrm{t}}(x, 0) = 0
\end{aligned}
\end{equation*}

All four ends of the beams are assumed to be simply supported, expressed as

\begin{equation*}
    \begin{aligned}
    w_1(0, t) = w_1(\pi, t) = w_{1_\mathrm{xx}}(0, t) = w_{1_\mathrm{xx}}( \pi, t)  = 0 \\ w_2(0, t) = w_2(\pi, t) = w_{2_\mathrm{xx}}(0, t) = w_{2_\mathrm{xx}}(\pi, t) = 0
    \end{aligned}
\end{equation*}

The external acting force is

\begin{equation*}
    \begin{aligned}
    f_1(x, t) = \left(1-\frac{\pi}{2}\right)\sin(x)\cos(t) \\
    f_2(x, t) = \left(\frac{\pi}{2}-1\right)\sin(x)\cos(t)
    \end{aligned}
\end{equation*}

For the considered problem, the analytical solution is given by

\begin{equation*}
    \begin{aligned}
    w_1(x, t)= \sin(x)\cos(t),\quad
    w_2(x, t) = \frac{\pi}{2}\sin(x)\cos(t)
    \end{aligned}
\end{equation*}

In addition to computing the beam displacements, derived quantities such as velocity, acceleration, and bending moment are also computed for this problem. These derived quantities also help in the prognosis and diagnostics of the system. For instance, the bending moment estimates the bending effect when an external force is applied to a structural element. Estimating the bending moment can be used to quantify the bending upon the action of applied forces. The beam is the most common structural member vulnerable to bending moments because it can bend at any point along its length when subjected to an external force. 

\begin{table}[htbp]
\caption{Euler-Bernoulli double-beam: $\mathcal{R}$ at $t=1$}
\begin{center}
\begin{tabular}{|c|c|c|}
\hline
& First beam& Second beam \\
\hline

Displacement (\%)& $1.9348\times 10^{-5}$& $4.3253 \times 10^{-5}$ \\
\hline
{{Bending Moment (\%) }}&  $9.6112 \times 10^{-4}$& $6.5506 \times 10^{-4}$   \\
\hline
Velocity (\%)& $1.9043 \times 10^{-3}$ &  $2.0161 \times 10^{-3}$ \\
\hline
Acceleration (\%)& $1.9011 \times 10^{-2}$ &  $1.4442 \times 10^{-2}$ \\
\hline
\end{tabular}
\label{tab1}
\end{center}
\end{table}

For simulating Euler-Bernoulli double beams, the same neural network architecture as for the single Euler-Bernoulli beam is considered. The only change is in the residual parameter, which is $1$ for this case. The results are illustrated in Figs. (~\ref{fig8}-~\ref{fig9}) and Table ~\ref{tab1}. The absolute difference between the PINN predicted solution and the exact solution for the first beam is approximately $10^{-4}$, and for the second beam, it is approximately $10^{-3}$, as shown in Fig.~\ref{fig8}. The bending moment, velocity and acceleration are computed using the neural network's autodifferentiation and backpropagation features. Table ~\ref{tab1} describes the efficiency in the computation of these quantities at $t=1$ for both beams. The relative percent error in computing the transverse displacement of the beams on the order of $10^{-5}$, and for acceleration, this error is on the order of $10^{-2}$, which is very low and shows the potential of physics-informed learning. Fig.~\ref{fig9} illustrates the computed velocity, bending moment, and acceleration of both beams. 

\subsection{Timoshenko Double-Beam Forward Problem}

The double-beam system modeled by Euler-Bernoulli theory can also be modelled using Timoshenko theory under the same assumptions as described for the single Timoshenko equations \cite{stojanovic2012forced}. In addition to providing the transverse displacement of the beams, Timoshenko theory also provides the cross-sectional rotation of both beams through the system of PDEs \cite{stojanovic2012forced} given by

\begin{equation*}
\begin{aligned}
kA_{1}G(\bar{\theta}_{1_\mathrm{\bar{x}}} - \bar{w}_{1_\mathrm{\bar{x}\bar{x}}}) + \rho A_{1}\bar{w}_{1_\mathrm{\bar{t}\bar{t}}} + K(\bar{w}_{1} - \bar{w}_{2}) = \bar{f}_{1}(\bar{x}, \bar{t}) 
\end{aligned}
\end{equation*}
\begin{equation*}
   \begin{aligned}
   EI_{2}\bar{\theta}_{2_\mathrm{\bar{x}\bar{x}}} + GA_{2}k(\bar{w}_{2_\mathrm{\bar{x}}} - \bar{\theta}_{2}) - \rho I_{2}\bar{\theta}_{2_\mathrm{\bar{t}\bar{t}}} = 0 
   \end{aligned} 
\end{equation*}
\begin{equation*}
    \begin{aligned}
    kA_{2}G(\bar{\theta}_{2_\mathrm{\bar{x}}} - \bar{w}_{2_\mathrm{\bar{x}\bar{x}}}) + \rho A_{2} \bar{w}_{2_\mathrm{\bar{t}\bar{t}}} + K(\bar{w}_{2} - \bar{w}_{1}) = \bar{f}_{2}(\bar{x}, \bar{t}) 
    \end{aligned}
\end{equation*}
\begin{equation}
    \begin{aligned}
    EI_{1}\bar{\theta}_{1_\mathrm{\bar{x}\bar{x}}} + GA_{1}k (\bar{w}_{1_\mathrm{\bar{x}}} - \bar{\theta}_1) - \rho I_{1}\bar{\theta}_{1_\mathrm{\bar{t}\bar{t}}} = 0
\label{eq16}
    \end{aligned}
\end{equation}
where $\bar{w}_\mathrm{i}(\bar{x}, \bar{t})$ and $\bar{\theta}_\mathrm{i}(\bar{x}, \bar{t})$, $i = 1, 2$ denote the transverse displacement and cross-sectional rotation of the beams respectively. $K$ is the stiffness modulus of the Winkler elastic layer. G is the shear modulus and $k$ is the Timoshenko shear coefficient. The rest of the parameters have the usual meanings as described earlier. For simplicity, we consider $A_1 = A_2$, and $I_1 = I_2$ and nondimensionalize~\eqref{eq16}. With some additional assumptions, the non-dimensional equation has the same form as~\eqref{eq16} with unit coefficients. For the numerical experiment the initial state of the double beam system is taken to be

\begin{figure*} 
\centering
\includegraphics[width=0.49\columnwidth]{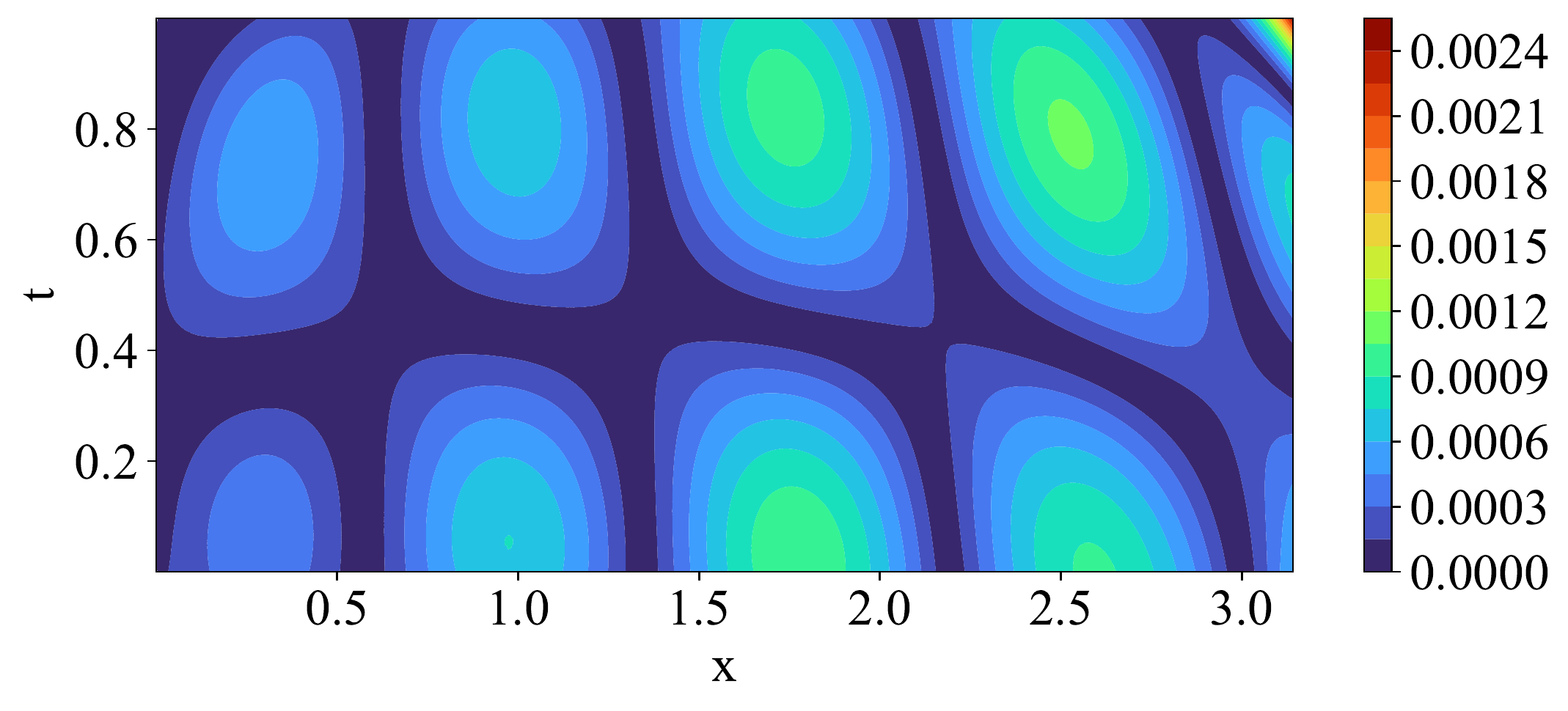}
\includegraphics[width=0.49\columnwidth]{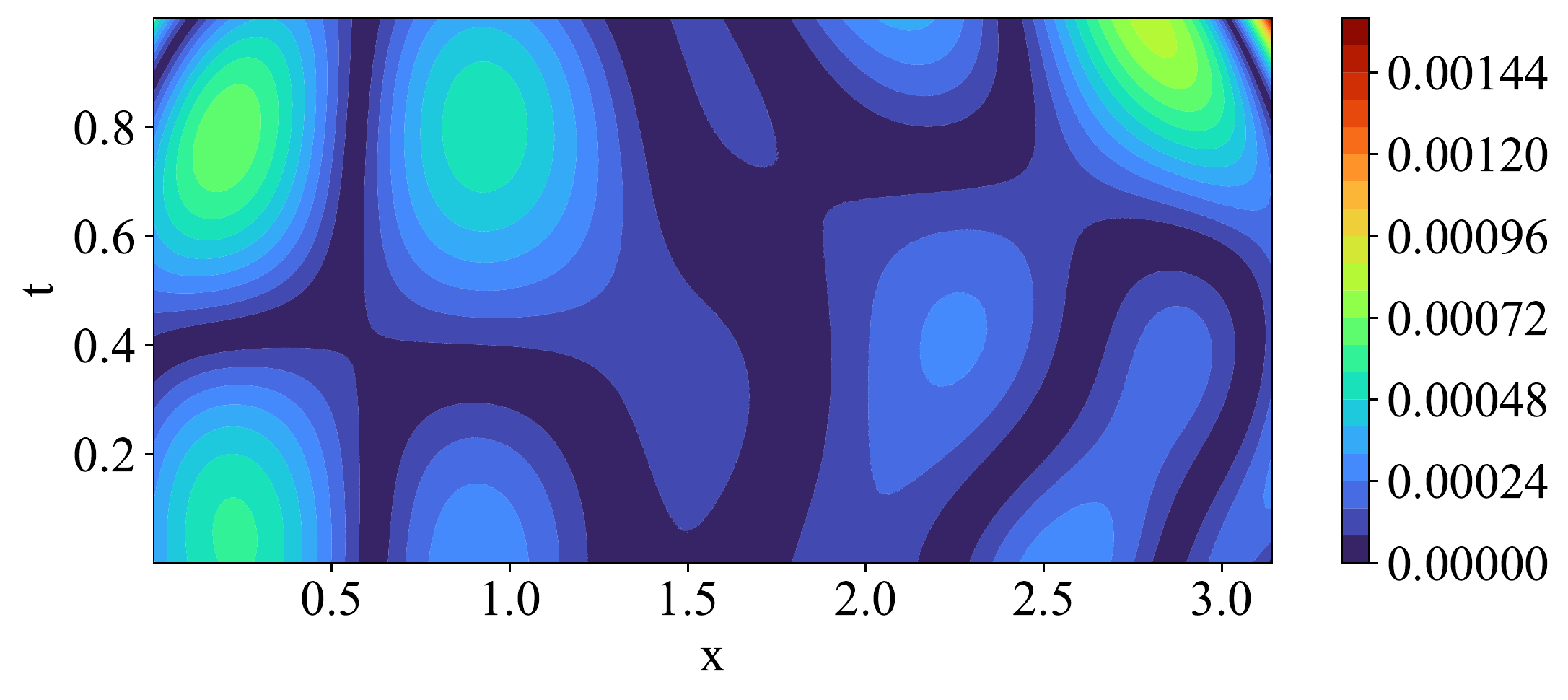}
\includegraphics[width=0.49\columnwidth]{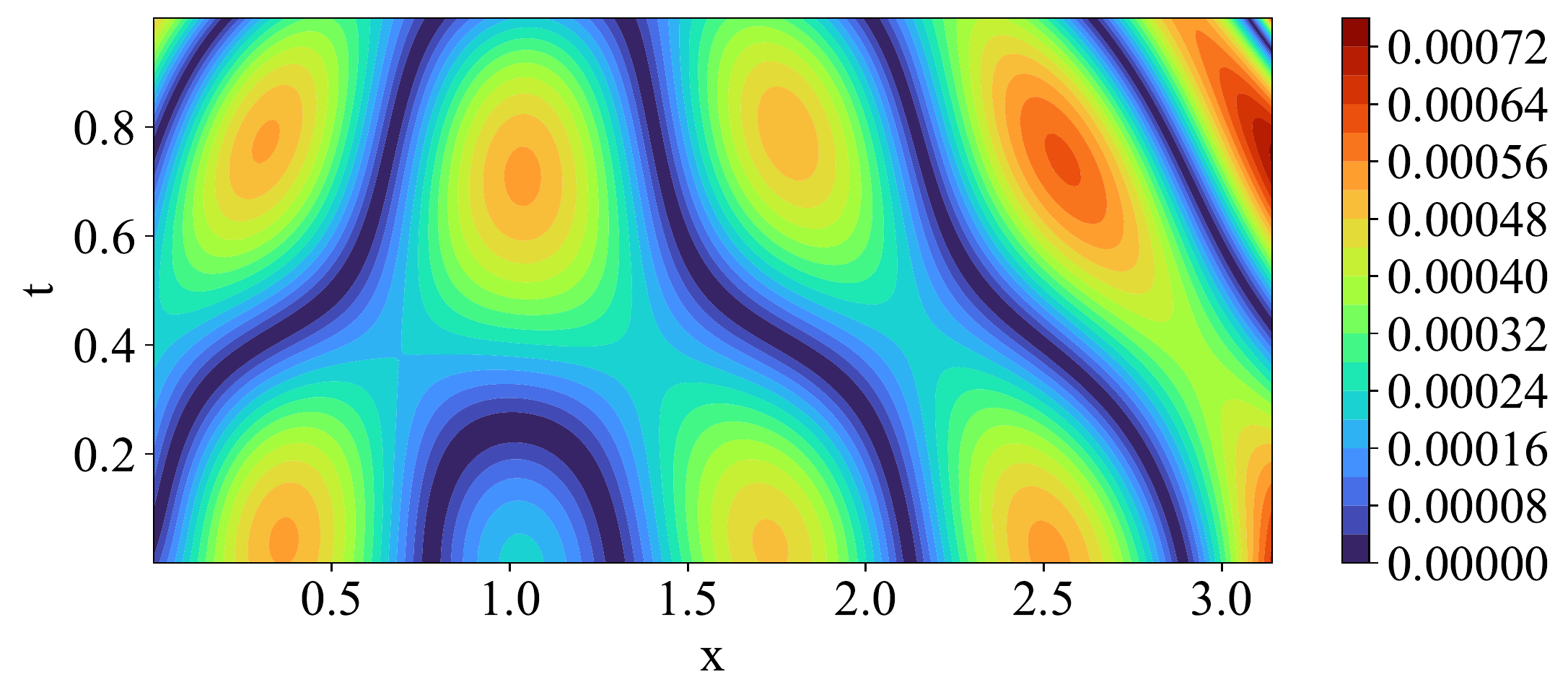}
\includegraphics[width=0.49\columnwidth]{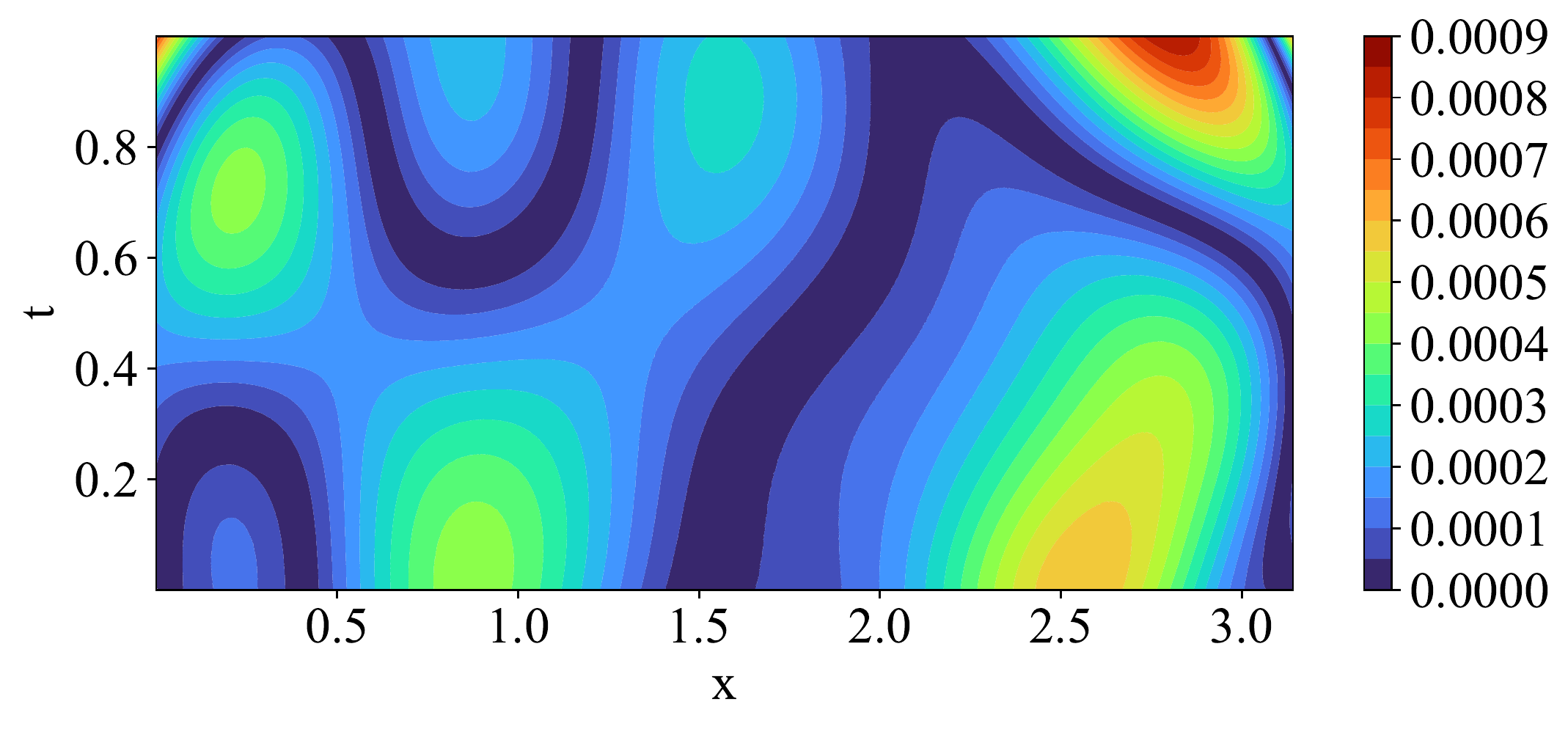}
\includegraphics[width=0.49\columnwidth]{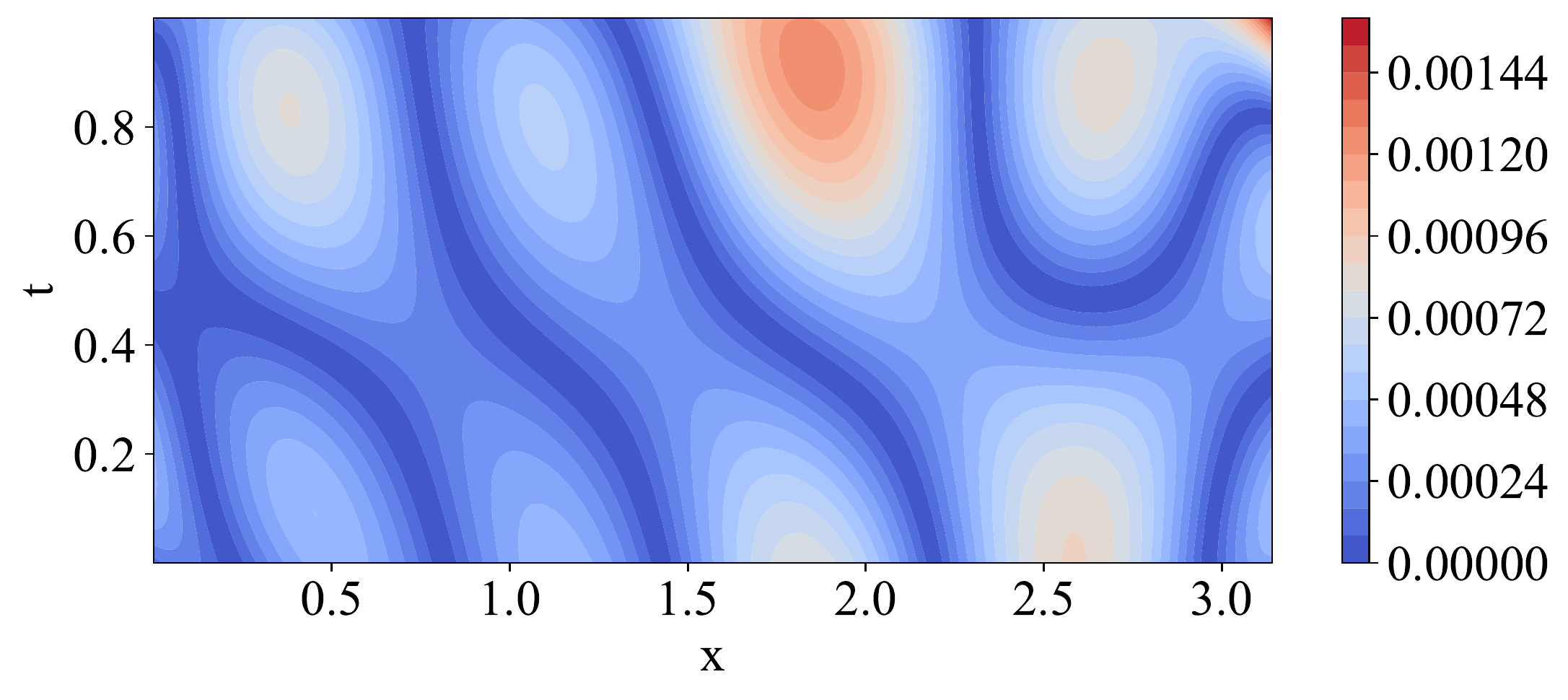}
\includegraphics[width=0.49\columnwidth]{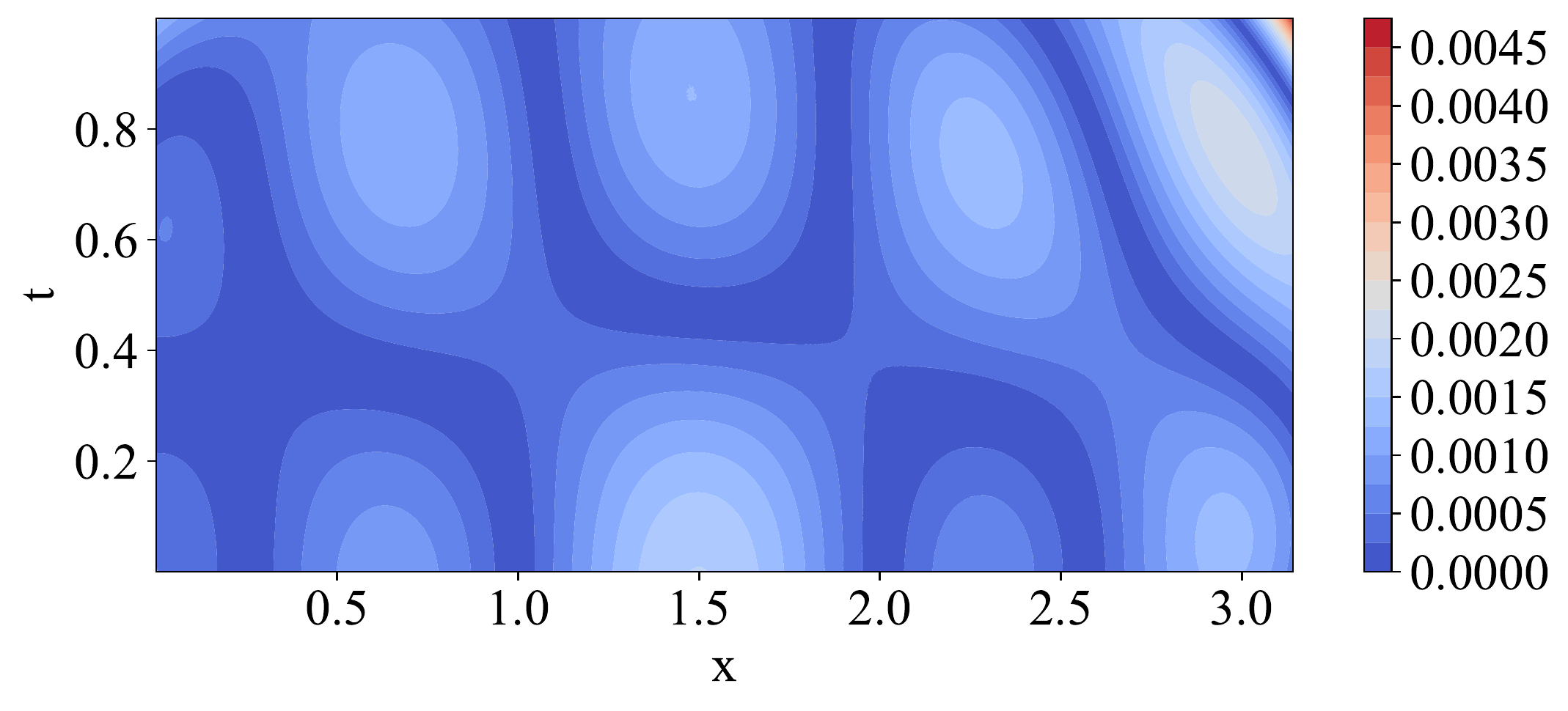}
\includegraphics[width=0.49\columnwidth]{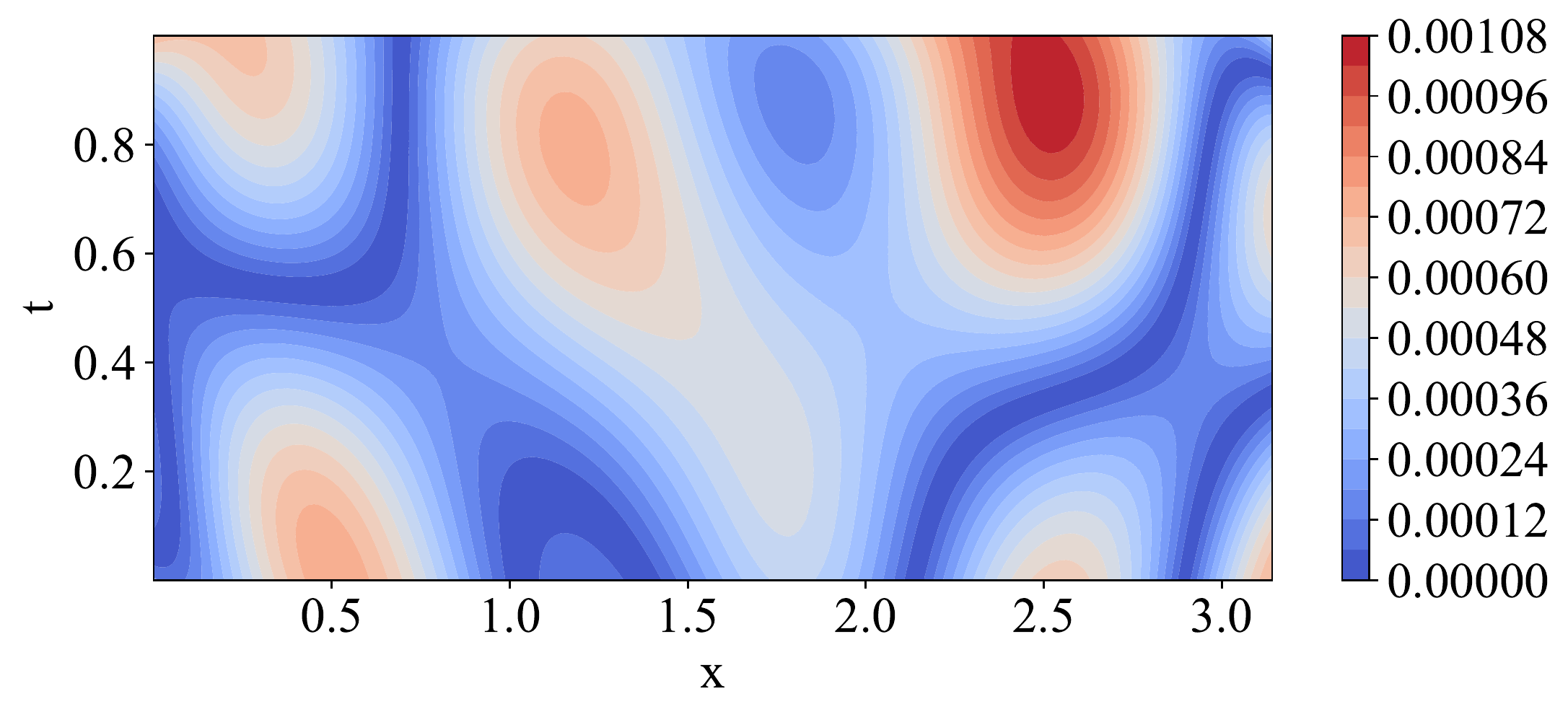}
\includegraphics[width=0.49\columnwidth]{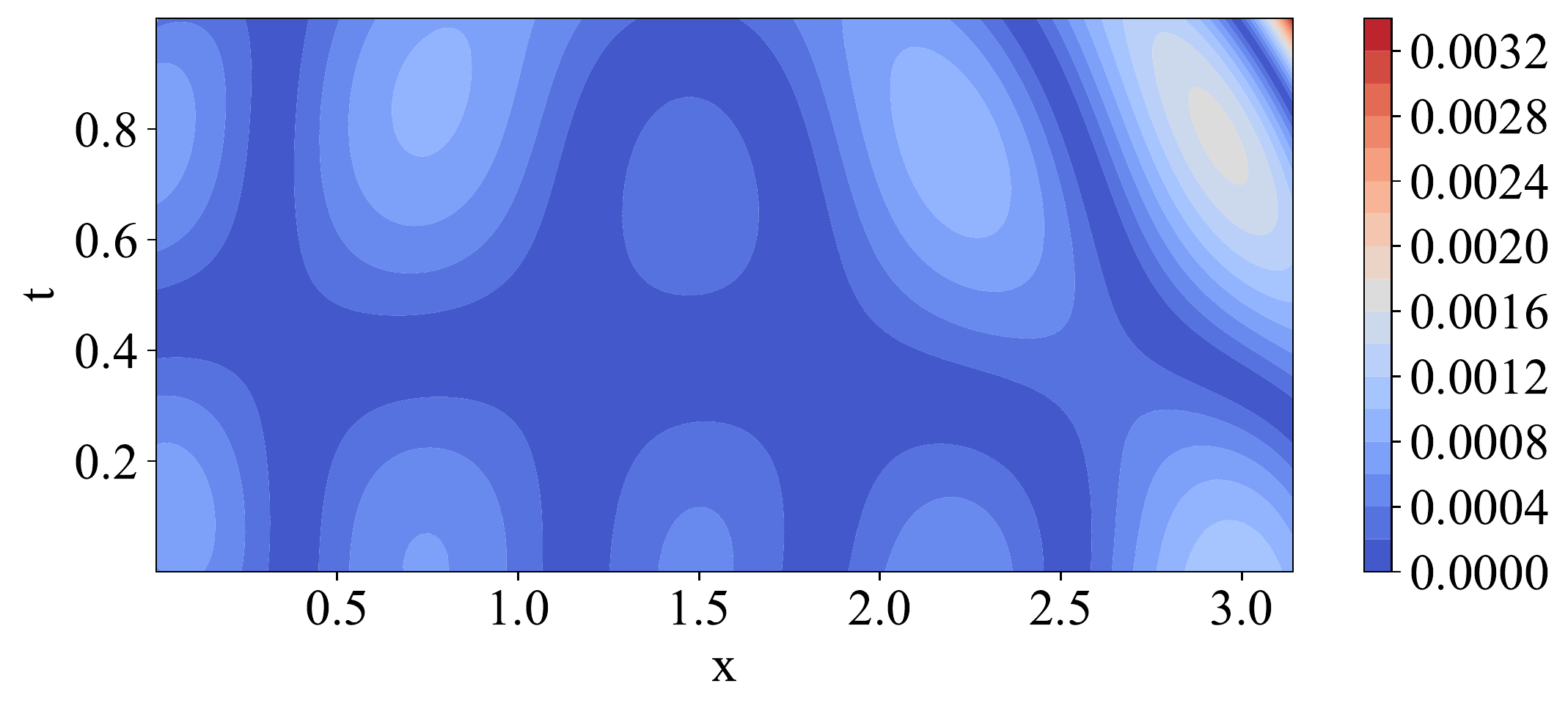}
\caption{Timoshenko double beam absolute errors in prediction;  $\theta_1$ and $w_1$ are the rotation and displacement of the first beam, $\theta_2$ and $w_2$ are the rotation and displacement of the second beam \textbf{Top} $16000$ training points  \textbf{(a)} $|\theta_1 - \theta_1^\mathrm{*}|$; \textbf{(b)}  $|w_1 - w_1^{*}|$; \textbf{(c)} $|\theta_2-\theta_2^\mathrm{*}|$; \textbf{(d)} $|w_2 - w_2^{*}|$. \textbf{Bottom} $1600$ training points \textbf{(a)} $|\theta_1 - \theta_1^\mathrm{*}|$; \textbf{(b)}  $|w_1 - w_1^{*}|$; \textbf{(c)} $|\theta_2-\theta_2^\mathrm{*}|$; \textbf{(d)} $|w_2 - w_2^{*}|$.}
\label{fig12}
\end{figure*}

\begin{equation*}
    \begin{aligned}
    \theta_1(x, 0) &= \left(\frac{\pi}{2}\cos(x) + \left(x - \frac{\pi}{2}\right)\right),\quad
    \theta_{1_{t}}(x, 0) = 0\\
    w_1(x, 0) &= \frac{\pi}{2}\sin(x),\quad
    w_{1_{t}}(x, 0) = 0\\
    \theta_2(x, 0) &= \frac{2}{\pi}\left(\frac{\pi}{2}\cos(x) + \left(x - \frac{\pi}{2}\right)\right),\quad
    \theta_{2_{t}}(x, 0) = 0\\
    w_2(x, 0) &= \sin(x), \quad
    w_{2_\mathrm{t}}(x, 0) = 0
    \end{aligned}
\end{equation*}

Simply supported boundary conditions are provided to make the problem wellposed

\begin{equation*}
    \begin{aligned}
    \theta_1(0, t) &= \theta_1(\pi, t)=w_1(0, t) = w_1(\pi, t) = 0 \\ 
\theta_2(0, t) &= \theta_2(\pi, t)= w_2(0, t) = w_2(\pi, t) = 0
    \end{aligned}
\end{equation*}
Here, $f_1(x, t)$, $f_2(x, t)$ and the analytic solutions are as follows
\begin{equation*}
    \begin{aligned}
    f_1(x, t) &= \cos(t)(1 - \sin(x))\\
f_2(x, t) &= \frac{2}{\pi}\cos(t)-\frac{\pi}{2}\sin(x)\cos(t)
    \end{aligned}
    \end{equation*}

\begin{equation*}
    \begin{aligned}
     \theta_1(x, t)&= \left(\frac{\pi}{2}\cos(x) + \left(x - \frac{\pi}{2}\right)\right)\cos(t)\\
\theta_2(x, t)&= \frac{2}{\pi}\left(\frac{\pi}{2}\cos(x) + \left(x - \frac{\pi}{2}\right)\right)\cos(t)\\
w_1(x, t) &= \frac{\pi}{2}\sin(x)\cos(t), \quad w_2(x, t) = \sin(x)\cos(t) \\
 \end{aligned}
\end{equation*}

\begin{table}[htbp]
\caption{Timoshenko double-beam: hyperparameters}
\begin{center}
\begin{tabular}{|c|c|c|c|c|c|c|c|}
\hline
No. of points& $N_\mathrm{i}$ & $N_\mathrm{b}$ & $N_\mathrm{int}$ & Layers & Neurons & Epochs\\ \hline
16000 &2000&2000&10000&4&20&15K \\ \hline
1600&200&200&1000&4&20&15K\\ \hline

\end{tabular}
\label{tab2}
\end{center}
\end{table}

\begin{table}[htbp]
\caption{Timoshenko double-beam: $\mathcal{R}$ at $t=1$}
\begin{center}
\begin{tabular}{|c|c|c|}
\hline
\textbf{}& $16000$ points& $1600$ points \\
\hline
{{$\theta_1$}} (\%)& $1.6038 \times 10^{-3}$ &  $2.6211 \times 10^{-3}$ \\
\hline
$w_1 (\%)$& $3.9302\times 10^{-5}$& $2.503 \times 10^{-4}$ \\
\hline
{{$\theta_2$ (\%)}}&  $1.0826 \times 10^{-3}$& $4.9405 \times 10^{-3}$   \\
\hline
$w_2 (\%)$& $7.8614 \times 10^{-5}$ &  $3.4904 \times 10^{-4}$ \\
\hline
\end{tabular}
\label{tab3}
\end{center}
\end{table}



Two experiments are performed, varying the number of training points, as shown in Table ~\ref{tab2}. Table ~\ref{tab3} shows the relative percent error in approximating the transverse displacement and cross-sectional rotations for both beams. For cross-sectional rotations $\theta_1$  and $\theta_2$, the magnitude of the percent error remains the same even for fewer training points. 

Using a large number of training points can increase the training time and may not be feasible for problems with many parameters. In these cases, using fewer training points can lead to less accurate solutions, but they can be obtained relatively faster. This approach allows engineers to make informed decisions about the parameters, and once optimal parameters have been identified, forward solutions can be recalculated with higher accuracy by using more training points. This is referred to as training with fewer points for the forward problem.
The absolute difference between the predicted and exact solutions of $\theta_1$, $w_1$, $\theta_2$ and $w_2$, even for $1600$ training points is very small as shown in Fig.~\ref{fig11} Fig.~\ref{fig12}. 
Fig.~\ref{fig11} presents the PINNs prediction for a double Timoshenko beam. The scattered points refer to the exact solution, and the continuous line represents the predicted solution. The force is applied uniformly in both beams; however, the deflection and rotation of the first beam are greater than those of the second beam. The results in Fig.~\ref{fig12} indicate that, for the second beam, a larger number of training points (16000) results in a more accurate prediction of deflection and rotation than a smaller number of training points (1600). Conversely, for the first beam, a smaller number of training points (1600) results in a more accurate prediction of the quantity of interest than a larger number of training points (16000). In any case, the difference in absolute error is relatively small, demonstrating that even with fewer training points, PINNs can still produce accurate predictions.\\

\subsection{Timoshenko Double-Beam Inverse Problem}
The applied force on structural systems is critical for structural design and condition assessment. In design, control, and diagnosis, accurate estimation of dynamic forces acting on a structure is essential. These details can be used to evaluate the structural condition. For example, understanding the impact of heavy vehicles on bridge structures can aid in detecting early damage to them. Indirect force determination is of special interest when the applied forces cannot be measured directly, while the responses can be measured easily.

For the inverse problem, three distinct experiments are performed on~\eqref{eq16}. First, the unknown parameter is learned from the Timoshenko double-beam system. We consider the unknown parameter to be $\rho A_{1}$ from~\eqref{eq16}. For the value of $\rho A_{1} = 1$, the data for transverse displacement and cross-sectional rotation are provided at some points in the computational domain. Second, the unknown applied function on the first beam is learned by providing noise-free simulated displacement and cross-sectional rotation data. For this case, all other parameters, initial and boundary conditions are considered to be known, and only the function $f_{1}(x,t)$ is unknown. Third, the same force function is predicted by providing noisy displacement and cross-sectional rotation data. The data generated for learning the function in the second case are corrupted with noise to be used in the third case. The exact solution for the function to be learned in the second and third cases is $\cos({t})(1-\sin({x}))$. 

\begin{figure*} 
\centering
\includegraphics[width=1\columnwidth]{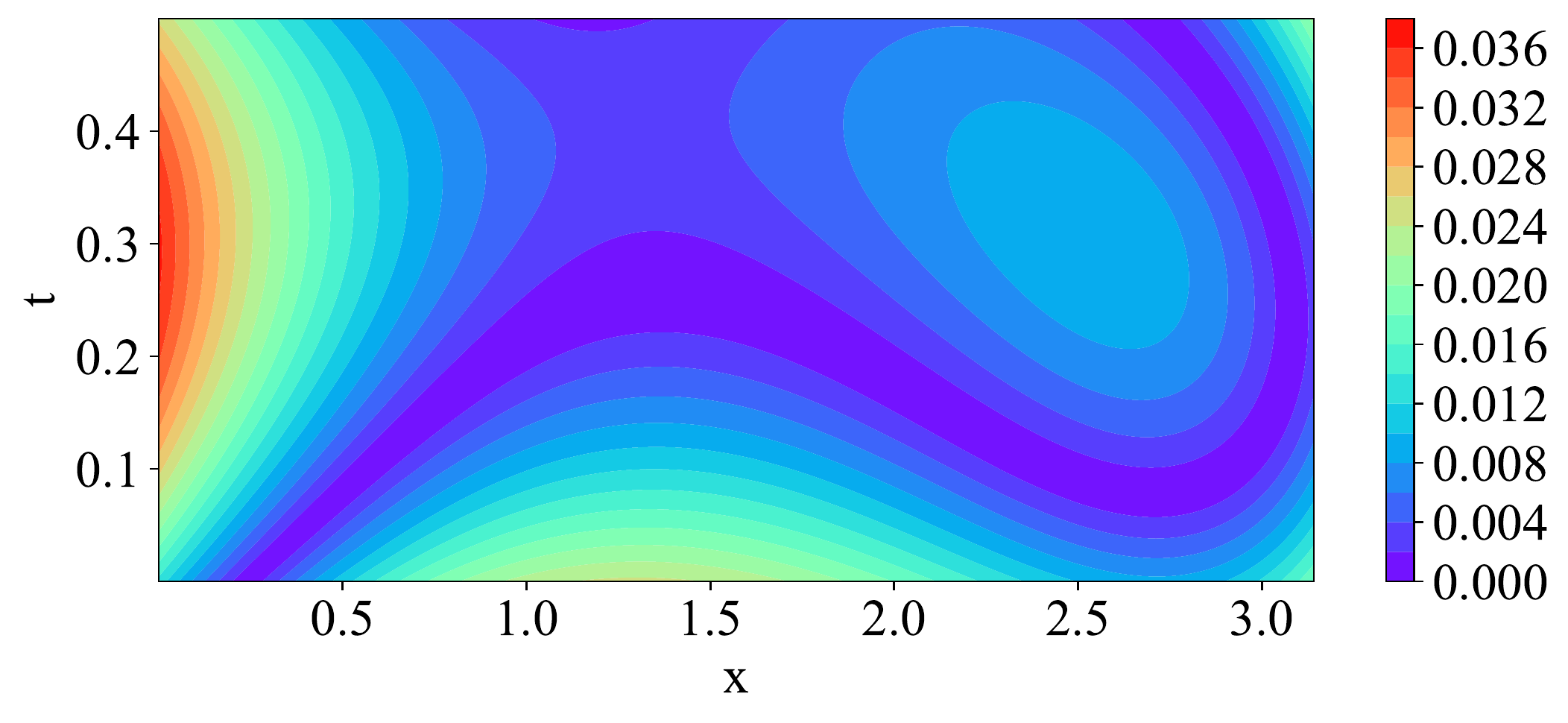}
\includegraphics[width=1\columnwidth]{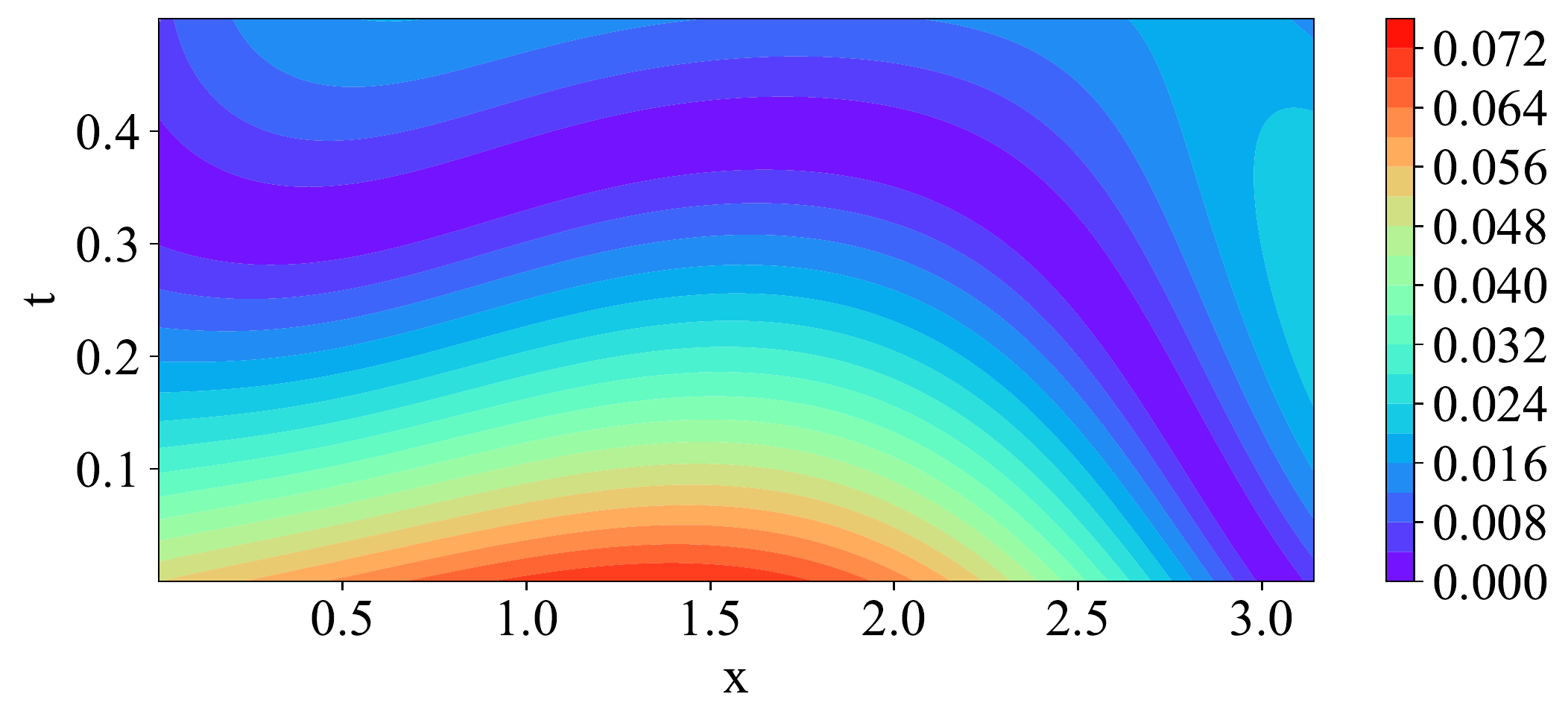}
\caption{Timoshenko double-beam inverse problem: absolute error in the prediction of force when the additional data of rotation and deflections provided at five locations has \textbf{left:} no noise  \textbf{right:} 20 percent Gaussian noise.}
\label{fig13}
\end{figure*}

\begin{figure}[htbp]
\centerline{\includegraphics[width=0.8\columnwidth]{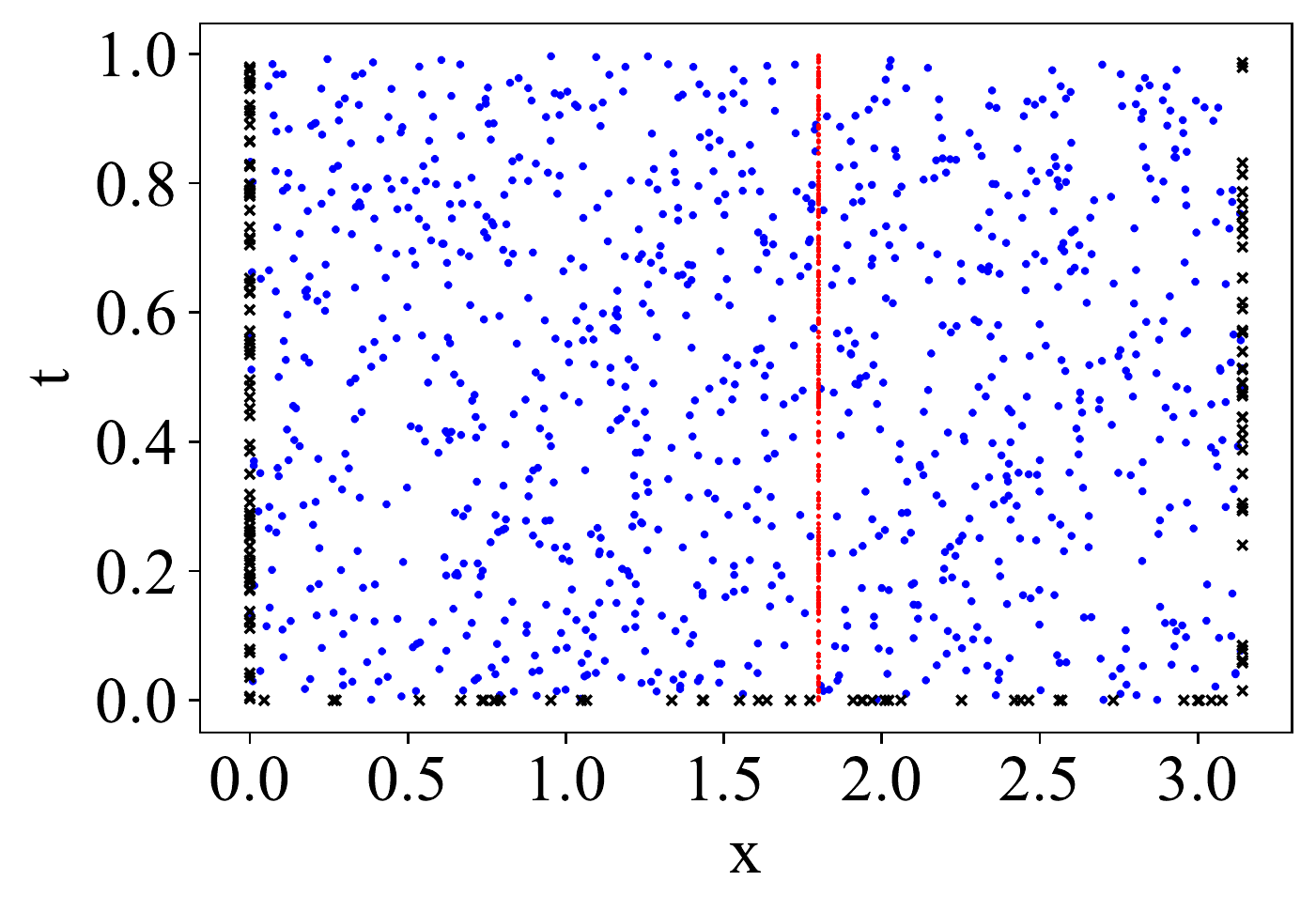}}
\caption{Data to learn material properties for the Timoshenko double beam: \textbf{Blue dots} Collocation points. \textbf{Red dots} Additional data points of displacement and rotation for the double beam at one location. \textbf{Black dots} Initial and boundary points.}
\label{fig14}
\end{figure}

\begin{figure}[htbp]
\centerline{\includegraphics[width=0.8\columnwidth]{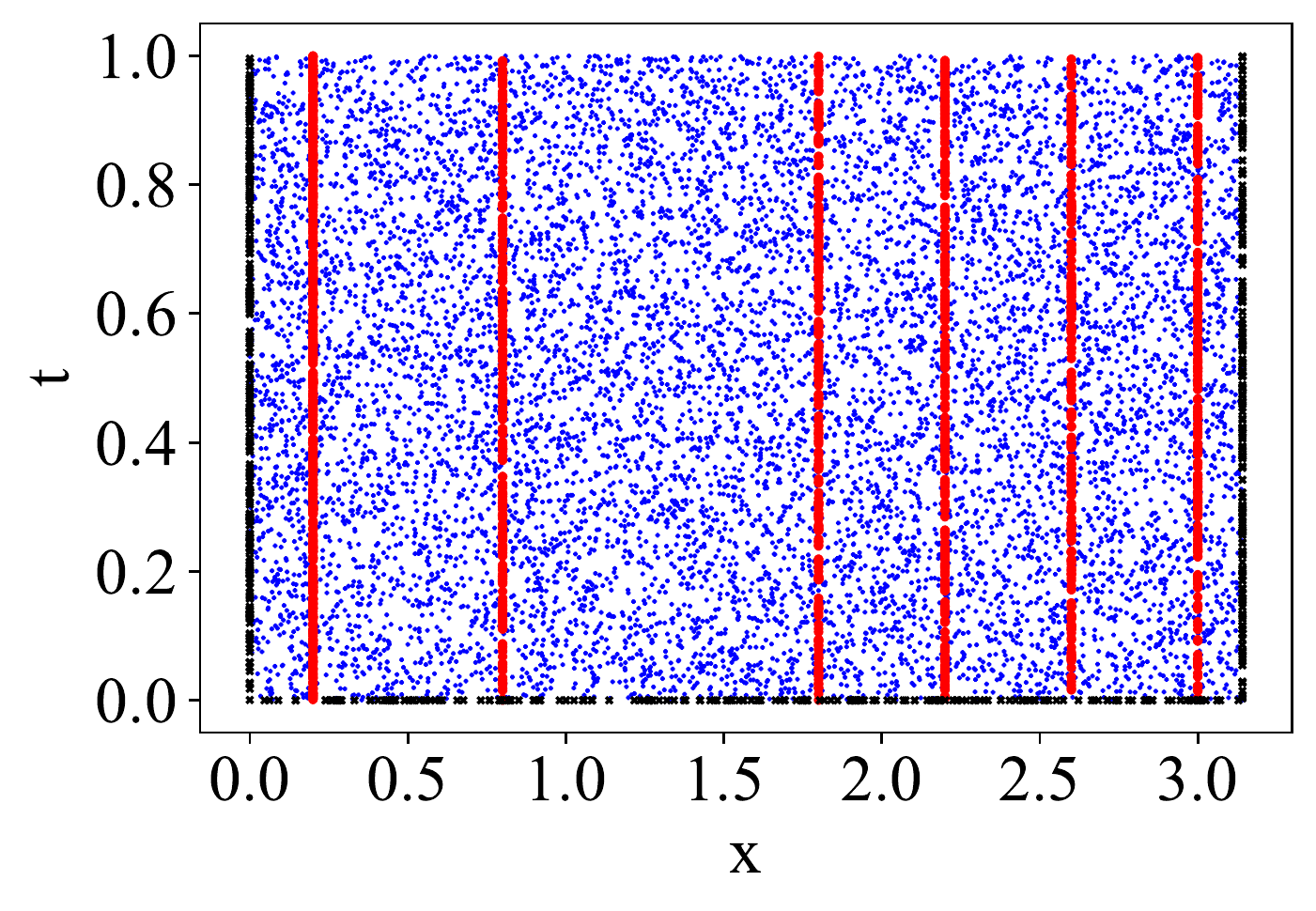}}
\caption{Data to learn force for the Timoshenko double beam: \textbf{Blue dots} Collocation points. \textbf{Red dots} Additional data points of displacement and rotation for the double beam at six different locations. \textbf{Black dots} Initial and boundary points.}
\label{fig15}
\end{figure}
 The inverse problem in engineering refers to the process of estimating unknown parameters or functions from a set of measured data. In PINNs, the inverse problem is usually solved by training a neural network to fit the measured data and the known physical laws. However, the measured data can be affected by various sources of noise, which can make estimation of the quantity of interest more challenging. The noise can make the measured data unreliable, and the neural network may not be able to accurately estimate the unknown parameters or functions. In such a scenario, the optimizer of the neural network does not necessarily converge to local minima.

The same neural network architecture is used as in the forward double-beam Timoshenko problem, with residual parameter $1$ to regularize the physical equation in the loss function. Here, $2500$ epochs are performed using the L-BFGS optimizer to train the neural network. For learning the parameter, $5000$ data points are provided at $x = 1.8$, as shown in Fig.~\ref{fig14}. The exact value of the unknown parameter is $\rho A_1 =1$ in  ~\eqref{eq16}, and the predicted value of the parameter using the PINN framework is 1.0208, which is close to the desired value. Even for a system of four PDEs, by only providing data at one particular beam location, the unknown parameter is learned successfully using PINNs. This shows that PINNs can handle large complex systems of PDEs efficiently.

\begin{table}[htbp]
\caption{Timoshenko double-beam inverse problem: noise vs. $\mathcal{R}$}
\begin{center}
\begin{tabular}{|c|c|}
\hline
Noise percent (\%)& Relative error percent (\%) \\
\hline
$0$ & $4.3271 \times 10^{-2}$\\
\hline
$10$ & $4.8688 \times 10^{-2}$\\
\hline
$20$ & $1.1123 \times 10^{-1}$\\
\hline
\end{tabular}
\label{tab4}
\end{center}
\end{table}

The function $f_1(x, t)$, the applied force on the first beam is predicted in the second experiment. As illustrated in Fig. \ref{fig15}, the data for transverse displacement and cross-sectional rotation are provided at $6$ different locations with $5000$ data points at each location.

For the third experiment, the data provided for learning the unknown function $f_{1}(x,t)$ are provided with $10\%$ and $20\%$ Gaussian noise and the corresponding performance in learning the function is shown in Table ~\ref{tab4}. Even with $10\%$ and $20\%$ noise, the relative error percent between analytic and predicted force is lower, as seen in TABLE ~\ref{tab4}.  Fig.~\ref{fig13} shows the force prediction along the beam when rotation and deflection observations are available at five points. The results demonstrate that the PINN is more precise in its predictions when the data are free from noise compared to when they are noisy. Despite the presence of noise in the data, the absolute error remains within the magnitude of $10^{-2}$, which is comparable to the error observed when data are not noisy. To be more precise, Fig.~\ref{fig13} shows the absolute difference error of the PINN predicted and exact force at $t=0.5$ with $0$ percent and $20$ percent noise. Even with $20$ percent noise, the unknown force is learned with less than $1\%$ error over the entire space-time domain, demonstrating that PINN is a very accurate and robust approach. 

The minimum number of data points required to estimate the model parameters depends on several factors, such as the complexity of the physics, the number of physical parameters in the model, and the quality of the data. More data points and more complex physics require more neural network capacity, resulting in a larger neural network with more hyperparameters. In practice, more data points lead to overfitting. The minimum training data points required for a PINN framework are determined empirically by gradually increasing the number of training points until the model's performance is satisfactory.
 
Finally, a sensitivity analysis is carried out to examine the influence of input variables, specifically the displacement and rotation, on the output variable, which is the force. The analysis involves adding $20\%$ Gaussian noise to the displacement data while no noise is added to the rotation data. The resulting mean accuracy of the force is 0.14313413. In contrast, when $20\%$ noise is introduced to the rotation data with the displacement data remaining unaltered, the mean accuracy of the force is $0.204627$. The results of this analysis show that the force is more sensitive to rotations than the displacement data.

\section{Conclusions}
The design and maintenance of complex structural systems are challenging due to the multiscale interaction of their components. It is desirable to predict the behavior of these complex systems by solving the governing model of interest. Recently, PINNs have emerged as a viable method for simulating PDEs. In this work, we propose using the PINN algorithm with the nondimensionalization step aiding in the learning procedure for complex beam systems. The PINN framework successfully solves the forward and inverse problems for nondimensional single and double-beam systems. Based on the numerical experiments, the following conclusions are drawn.

First, the relative percent error in computing the beam displacement does not increase with increasing model complexity when solving the forward problem. In fact, for both Euler-Bernoulli and Timoshenko theory, the error decreases by an order of magnitude for double-beam systems compared to single-beam systems. In addition, the error in computing the bending rotation is comparable for single and double Timoshenko beam systems. This nonincrease in error as the model complexity increases suggests that the PINN framework is appropriate for simulating large-scale systems with multiple connected components.

Second, it is demonstrated that PINNs precisely discover the unknown force function and model parameters through their inverse problem-solving capability. The proposed algorithm successfully learns the model parameter with less than $3\%$ error for the single Timoshenko beam. In addition, for the double beam Timoshenko system, the unknown function is approximated on the whole space-time domain with less than $0.05 \%$ error, demonstrating the algorithm's effectiveness for solving inverse problems.

Third, physical quantities such as velocity, acceleration, and bending moment characterize the system's behavior. Even though the derived quantities are not directly trained in the neural network, they are approximated with less than $2e-2\%$ error for the Euler-Bernoulli double-beam system.

Fourth, the algorithm's ability to use fewer training points in forward problems and to accomodate noisy data in inverse problems is exploited. The obtained results show that even with $1600$ training points, the double Timoshenko beam displacement is predicted on the entire space-time domain with less than $5e-3\%$ error. In the case of the inverse problem, the force function is discovered with less than $0.2 \%$ error even when the data used in the learning procedure contains $20\%$ Gaussian noise. These findings imply that the algorithm is accurate and robust under the tested noise levels.

To summarize, PINNs enable the simulation of complex structural systems with multiple interacting components efficiently, accurately, and robustly. In the future, this approach could be extended to estimate displacements for various input forces and mechanical vibration modes and incorporate robust methods to account for stochasticities.  Additionally, future works on PINNs could be focused on reducing the computational cost and developing methodologies to augment their generalizability, thereby expanding the applicability of PINNs beyond the training domain.

\section*{Acknowledgment}
The authors would like to express their appreciation to the anonymous reviewers and editors for their valuable comments and feedback, which have significantly improved the quality of this work. The authors extend their appreciation to Prof. Siddhartha Mishra (ETH Zürich) for his insightful suggestions in comparisons with numerical methods.
\bibliographystyle{IEEEtran}
\bibliography{ref}

\begin{thebibliography}{10}
\providecommand{\url}[1]{#1}
\csname url@samestyle\endcsname
\providecommand{\newblock}{\relax}
\providecommand{\bibinfo}[2]{#2}
\providecommand{\BIBentrySTDinterwordspacing}{\spaceskip=0pt\relax}
\providecommand{\BIBentryALTinterwordstretchfactor}{4}
\providecommand{\BIBentryALTinterwordspacing}{\spaceskip=\fontdimen2\font plus
\BIBentryALTinterwordstretchfactor\fontdimen3\font minus
  \fontdimen4\font\relax}
\providecommand{\BIBforeignlanguage}[2]{{%
\expandafter\ifx\csname l@#1\endcsname\relax
\typeout{** WARNING: IEEEtran.bst: No hyphenation pattern has been}%
\typeout{** loaded for the language `#1'. Using the pattern for}%
\typeout{** the default language instead.}%
\else
\language=\csname l@#1\endcsname
\fi
#2}}
\providecommand{\BIBdecl}{\relax}
\BIBdecl

\bibitem{kobbacy2008complex}
K.~A. Kobbacy, D.~P. Murthy \emph{et~al.}, \emph{Complex system maintenance
  handbook}.\hskip 1em plus 0.5em minus 0.4em\relax Springer, 2008.

\bibitem{chatzis2013modeling}
M.~Chatzis and G.~Deodatis, ``Modeling of very large interacting multiple-beam
  systems with application to suspension bridge cables,'' \emph{Journal of
  Structural Engineering}, vol. 139, no.~9, pp. 1541--1554, 2013.

\bibitem{liu2018advances}
Z.~Liu, Y.~Song, Y.~Han, H.~Wang, J.~Zhang, and Z.~Han, ``Advances of research
  on high-speed railway catenary,'' \emph{Journal of modern transportation},
  vol.~26, no.~1, pp. 1--23, 2018.

\bibitem{foias2001navier}
C.~Foias, O.~Manley, R.~Rosa, and R.~Temam, \emph{Navier-Stokes equations and
  turbulence}.\hskip 1em plus 0.5em minus 0.4em\relax Cambridge University
  Press, 2001, vol.~83.

\bibitem{lucor2022simple}
D.~Lucor, A.~Agrawal, and A.~Sergent, ``Simple computational strategies for
  more effective physics-informed neural networks modeling of turbulent natural
  convection,'' \emph{Journal of Computational Physics}, vol. 456, p. 111022,
  2022.

\bibitem{stojanovic2015vibrations}
V.~Stojanovi{\'c} and P.~Kozi{\'c}, \emph{Vibrations and stability of complex
  beam systems}.\hskip 1em plus 0.5em minus 0.4em\relax Springer, 2015.

\bibitem{mercieca2016estimation}
J.~Mercieca and V.~Kadirkamanathan, ``Estimation and identification of
  spatio-temporal models with applications in engineering, healthcare and
  social science,'' \emph{Annual Reviews in Control}, vol.~42, pp. 285--298,
  2016.

\bibitem{li2010modeling}
H.-X. Li and C.~Qi, ``Modeling of distributed parameter systems for
  applications—a synthesized review from time--space separation,''
  \emph{Journal of Process Control}, vol.~20, no.~8, pp. 891--901, 2010.

\bibitem{kapoor2022predicting}
T.~Kapoor, H.~Wang, A.~N{\'u}{\~n}ez, and R.~Dollevoet, ``Predicting traction
  return current in electric railway systems through physics-informed neural
  networks,'' in \emph{2022 IEEE Symposium Series on Computational Intelligence
  (SSCI)}.\hskip 1em plus 0.5em minus 0.4em\relax IEEE, 2022, pp. 1460--1468.

\bibitem{chandra2022physics}
A.~Chandra, M.~Curti, K.~Tiels, E.~A. Lomonova, and D.~M. Tartakovsky,
  ``Physics-informed neural networks for modelling anisotropic and
  bi-anisotropic electromagnetic constitutive laws through indirect data,'' in
  \emph{2022 IEEE Symposium Series on Computational Intelligence (SSCI)}.\hskip
  1em plus 0.5em minus 0.4em\relax IEEE, 2022, pp. 1451--1459.

\bibitem{yuan2022pinn}
L.~Yuan, Y.-Q. Ni, X.-Y. Deng, and S.~Hao, ``A-{PINN}: Auxiliary physics
  informed neural networks for forward and inverse problems of nonlinear
  integro-differential equations,'' \emph{Journal of Computational Physics},
  vol. 462, p. 111260, 2022.

\bibitem{fallah2023physics}
A.~Fallah and M.~M. Aghdam, ``Physics-informed neural network for bending and
  free vibration analysis of three-dimensional functionally graded porous beam
  resting on elastic foundation,'' \emph{Engineering with Computers}, pp.
  1--18, 2023.

\bibitem{kapoor2023physics}
T.~Kapoor, H.~Wang, A.~N{\'u}{\~n}ez, and R.~Dollevoet, ``Physics-informed
  machine learning for moving load problems,'' \emph{arXiv preprint
  arXiv:2304.00369}, 2023.

\bibitem{kapteyn2021probabilistic}
M.~G. Kapteyn, J.~V. Pretorius, and K.~E. Willcox, ``A probabilistic graphical
  model foundation for enabling predictive digital twins at scale,''
  \emph{Nature Computational Science}, vol.~1, no.~5, pp. 337--347, 2021.

\bibitem{wang2019bayesian}
H.~Wang, A.~N{\'u}{\~n}ez, Z.~Liu, D.~Zhang, and R.~Dollevoet, ``A {Bayesian}
  network approach for condition monitoring of high-speed railway catenaries,''
  \emph{IEEE Transactions on Intelligent Transportation Systems}, vol.~21,
  no.~10, pp. 4037--4051, 2019.

\bibitem{yuan2020machine}
F.-G. Yuan, S.~A. Zargar, Q.~Chen, and S.~Wang, ``Machine learning for
  structural health monitoring: challenges and opportunities,'' in
  \emph{Sensors and smart structures technologies for civil, mechanical, and
  aerospace systems 2020}, vol. 11379.\hskip 1em plus 0.5em minus 0.4em\relax
  International Society for Optics and Photonics, 2020, p. 1137903.

\bibitem{borggaard1997pde}
J.~Borggaard and J.~Burns, ``A {PDE} sensitivity equation method for optimal
  aerodynamic design,'' \emph{Journal of Computational Physics}, vol. 136,
  no.~2, pp. 366--384, 1997.

\bibitem{lai2021structural}
Z.~Lai, C.~Mylonas, S.~Nagarajaiah, and E.~Chatzi, ``Structural identification
  with physics-informed neural ordinary differential equations,'' \emph{Journal
  of Sound and Vibration}, vol. 508, p. 116196, 2021.

\bibitem{raissi2018hidden}
M.~Raissi and G.~E. Karniadakis, ``Hidden physics models: Machine learning of
  nonlinear partial differential equations,'' \emph{Journal of Computational
  Physics}, vol. 357, pp. 125--141, 2018.

\bibitem{abhyankar1993chaotic}
N.~Abhyankar, E.~Hall, and S.~Hanagud, ``Chaotic vibrations of beams: numerical
  solution of partial differential equations,'' 1993.

\bibitem{karniadakis2021physics}
G.~E. Karniadakis, I.~G. Kevrekidis, L.~Lu, P.~Perdikaris, S.~Wang, and
  L.~Yang, ``Physics-informed machine learning,'' \emph{Nature Reviews
  Physics}, vol.~3, no.~6, pp. 422--440, 2021.

\bibitem{fang2021high}
{Fang, Zhiwei}, ``{A high-efficient hybrid physics-informed neural networks
  based on convolutional neural network},'' \emph{{IEEE Transactions on Neural
  Networks and Learning Systems}}, {2021}.

\bibitem{cuomo2022scientific}
S.~Cuomo, V.~S. Di~Cola, F.~Giampaolo, G.~Rozza, M.~Raissi, and F.~Piccialli,
  ``Scientific machine learning through physics--informed neural networks:
  where we are and what’s next,'' \emph{Journal of Scientific Computing},
  vol.~92, no.~3, p.~88, 2022.

\bibitem{blechschmidt2021three}
J.~Blechschmidt and O.~G. Ernst, ``Three ways to solve partial differential
  equations with neural networks—a review,'' \emph{GAMM-Mitteilungen},
  vol.~44, no.~2, p. e202100006, 2021.

\bibitem{raissi2019physics}
M.~Raissi, P.~Perdikaris, and G.~E. Karniadakis, ``Physics-informed neural
  networks: A deep learning framework for solving forward and inverse problems
  involving nonlinear partial differential equations,'' \emph{Journal of
  Computational Physics}, vol. 378, pp. 686--707, 2019.

\bibitem{meng2021physics}
{Meng, Yuxin and Rigall, Eric and Chen, Xueen and Gao, Feng and Dong, Junyu and
  Chen, Sheng}, ``{Physics-Guided Generative Adversarial Networks for Sea
  Subsurface Temperature Prediction},'' \emph{{IEEE Transactions on Neural
  Networks and Learning Systems}}, {2021}.

\bibitem{xu2022physics}
{Xu, Wengang and Zhou, Zheng and Li, Tianfu and Sun, Chuang and Chen, Xuefeng
  and Yan, Ruqiang}, ``{Physics-Constraint Variational Neural Network for Wear
  State Assessment of External Gear Pump},'' \emph{{IEEE Transactions on Neural
  Networks and Learning Systems}}, {2022}.

\bibitem{7959606}
A.~Karpatne, G.~Atluri, J.~H. Faghmous, M.~Steinbach, A.~Banerjee, A.~Ganguly,
  S.~Shekhar, N.~Samatova, and V.~Kumar, ``Theory-guided data science: A new
  paradigm for scientific discovery from data,'' \emph{IEEE Transactions on
  Knowledge and Data Engineering}, vol.~29, no.~10, pp. 2318--2331, 2017.

\bibitem{hassoun1995fundamentals}
M.~H. Hassoun \emph{et~al.}, \emph{Fundamentals of artificial neural
  networks}.\hskip 1em plus 0.5em minus 0.4em\relax MIT press, 1995.

\bibitem{oszkinat2022uncertainty}
{Oszkinat, Clemens and Luczak, Susan E and Rosen, IG}, ``{Uncertainty
  quantification in estimating blood alcohol concentration from transdermal
  alcohol level with physics-informed neural networks},'' \emph{{IEEE
  Transactions on Neural Networks and Learning Systems}}, {2022}.

\bibitem{hua2023physics}
{Hua, Jiaqi and Li, Yingguang and Liu, Changqing and Wan, Peng and Liu, Xu},
  ``{Physics-Informed Neural Networks With Weighted Losses by Uncertainty
  Evaluation for Accurate and Stable Prediction of Manufacturing Systems},''
  \emph{{IEEE Transactions on Neural Networks and Learning Systems}}, {2023}.

\bibitem{wang2022and}
S.~Wang, X.~Yu, and P.~Perdikaris, ``When and why {PINNs} fail to train: A
  neural tangent kernel perspective,'' \emph{Journal of Computational Physics},
  vol. 449, p. 110768, 2022.

\bibitem{krishnapriyan2021characterizing}
A.~Krishnapriyan, A.~Gholami, S.~Zhe, R.~Kirby, and M.~W. Mahoney,
  ``Characterizing possible failure modes in physics-informed neural
  networks,'' \emph{Advances in Neural Information Processing Systems},
  vol.~34, 2021.

\bibitem{kissas2020machine}
G.~Kissas, Y.~Yang, E.~Hwuang, W.~R. Witschey, J.~A. Detre, and P.~Perdikaris,
  ``Machine learning in cardiovascular flows modeling: Predicting arterial
  blood pressure from non-invasive {4D} flow {MRI} data using physics-informed
  neural networks,'' \emph{Computer Methods in Applied Mechanics and
  Engineering}, vol. 358, p. 112623, 2020.

\bibitem{9597485}
{Feng, Jiali and Liu, Zhijie and He, Xiuyu and Li, Qing and He, Wei},
  ``{Vibration Suppression of a High-Rise Building With Adaptive Iterative
  Learning Control},'' \emph{{IEEE Transactions on Neural Networks and Learning
  Systems}}, pp. {1--12}, {2021}.

\bibitem{joghataie2013simulating}
{Joghataie, Abdolreza and Torghabehi, Omid Oliyan}, ``{Simulating dynamic
  plastic continuous neural networks by finite elements},'' \emph{{IEEE
  {Transactions} on neural networks and learning systems}}, vol.~{25}, no.~{8},
  pp. {1583--1587}, {2013}.

\bibitem{9741838}
{Liu, Yu and Wu, Xiaoqi and Yao, Xiangqian and Zhao, Jingyi}, ``{Backstepping
  Technology-Based Adaptive Boundary {ILC} for an Input-Output-Constrained
  Flexible Beam},'' \emph{{IEEE Transactions on Neural Networks and Learning
  Systems}}, pp. {1--9}, {2022}.

\bibitem{7879337}
{He, Wei and Meng, Tingting and Huang, Deqing and Li, Xuefang}, ``{Adaptive
  Boundary Iterative Learning Control for an {Euler–Bernoulli} Beam System
  With Input Constraint},'' \emph{{IEEE Transactions on Neural Networks and
  Learning Systems}}, vol.~{29}, no.~{5}, pp. {1539--1549}, {2018}.

\bibitem{onis1}
Z.~Oniszczuk, ``Free transverse vibrations of elastically connected simply
  supported double-beam complex system,'' \emph{Journal of sound and
  vibration}, vol. 232, no.~2, pp. 387--403, 2000.

\bibitem{stojanovic2012forced}
V.~Stojanovi{\'c} and P.~Kozi{\'c}, ``Forced transverse vibration of {Rayleigh}
  and {Timoshenko} double-beam system with effect of compressive axial load,''
  \emph{International Journal of Mechanical Sciences}, vol.~60, no.~1, pp.
  59--71, 2012.

\bibitem{onis2}
Z.~Oniszczuk, ``Forced transverse vibrations of an elastically connected
  complex simply supported double-beam system,'' \emph{Journal of sound and
  vibration}, vol. 264, no.~2, pp. 273--286, 2003.

\bibitem{abu2006dynamic}
M.~Abu-Hilal, ``Dynamic response of a double {Euler--Bernoulli} beam due to a
  moving constant load,'' \emph{Journal of sound and vibration}, vol. 297, no.
  3-5, pp. 477--491, 2006.

\bibitem{zhao2020forced}
X.~Zhao, B.~Chen, Y.~Li, W.~Zhu, F.~Nkiegaing, and Y.~Shao, ``Forced vibration
  analysis of {Timoshenko} double-beam system under compressive axial load by
  means of green's functions,'' \emph{Journal of Sound and Vibration}, vol.
  464, p. 115001, 2020.

\bibitem{li2007spectral}
J.~Li and H.~Hua, ``Spectral finite element analysis of elastically connected
  double-beam systems,'' \emph{Finite Elements in Analysis and Design},
  vol.~43, no.~15, pp. 1155--1168, 2007.

\bibitem{palmeri2011galerkin}
A.~Palmeri and S.~Adhikari, ``A {Galerkin}-type state-space approach for
  transverse vibrations of slender double-beam systems with viscoelastic inner
  layer,'' \emph{Journal of Sound and Vibration}, vol. 330, no.~26, pp.
  6372--6386, 2011.

\bibitem{ying2017response}
Z.~Ying and Y.~Ni, ``A response-adjustable sandwich beam with harmonic
  distribution parameters under stochastic excitations,'' \emph{International
  Journal of Structural Stability and Dynamics}, vol.~17, no.~07, p. 1750075,
  2017.

\bibitem{ying2018vibration}
Z.~Ying, Y.~Ni, and R.~Huan, ``Vibration response characteristics of
  quasi-periodic sandwich beam with magnetorheological visco-elastomer core
  under random support excitations,'' \emph{Journal of Vibration and
  Acoustics}, vol. 140, no.~5, 2018.

\bibitem{murmu2010nonlocal}
T.~Murmu and S.~Adhikari, ``Nonlocal transverse vibration of
  double-nanobeam-systems,'' \emph{Journal of Applied Physics}, vol. 108,
  no.~8, p. 083514, 2010.

\bibitem{chen2021closed}
B.~Chen, B.~Lin, X.~Zhao, W.~Zhu, Y.~Yang, and Y.~Li, ``Closed-form solutions
  for forced vibrations of a cracked double-beam system interconnected by a
  viscoelastic layer resting on {Winkler}--{Pasternak} elastic foundation,''
  \emph{Thin-Walled Structures}, vol. 163, p. 107688, 2021.

\bibitem{liu2019closed}
S.~Liu and B.~Yang, ``A closed-form analytical solution method for vibration
  analysis of elastically connected double-beam systems,'' \emph{Composite
  Structures}, vol. 212, pp. 598--608, 2019.

\bibitem{zhao2021free}
X.~Zhao and P.~Chang, ``Free and forced vibration of double beam with arbitrary
  end conditions connected with a viscoelastic layer and discrete points,''
  \emph{International Journal of Mechanical Sciences}, vol. 209, p. 106707,
  2021.

\bibitem{li2021state}
Y.~Li, F.~Xiong, L.~Xie, and L.~Sun, ``State-space approach for transverse
  vibration of double-beam systems,'' \emph{International Journal of Mechanical
  Sciences}, vol. 189, p. 105974, 2021.

\bibitem{ong2022coupled}
O.~Z.~S. Ong, M.~H. Ghayesh, D.~Losic, and M.~Amabili, ``Coupled dynamics of
  double beams reinforced with bidirectional functionally graded carbon
  nanotubes,'' \emph{Engineering Analysis with Boundary Elements}, vol. 143,
  pp. 263--282, 2022.

\bibitem{9298920}
{Radev, Stefan T. and Mertens, Ulf K. and Voss, Andreas and Ardizzone, Lynton
  and Köthe, Ullrich}, ``{BayesFlow: Learning Complex Stochastic Models With
  Invertible Neural Networks},'' \emph{{IEEE Transactions on Neural Networks
  and Learning Systems}}, vol.~{33}, no.~{4}, pp. {1452--1466}, {2022}.

\bibitem{7302572}
{Ning, Hanwen and Qing, Guangyan and Jing, Xingjian}, ``{Identification of
  Nonlinear Spatiotemporal Dynamical Systems With Nonuniform Observations Using
  Reproducing-Kernel-Based Integral Least Square Regulation},'' \emph{{IEEE
  Transactions on Neural Networks and Learning Systems}}, vol.~{27}, no.~{11},
  pp. {2399--2412}, {2016}.

\bibitem{9716789}
{Jin, Pengzhan and Zhang, Zhen and Kevrekidis, Ioannis G. and Karniadakis,
  George Em}, ``{Learning {Poisson} Systems and Trajectories of Autonomous
  Systems via {Poisson} Neural Networks},'' \emph{{IEEE Transactions on Neural
  Networks and Learning Systems}}, pp. 1--13, {2022}.

\bibitem{mishra2022estimates}
S.~Mishra and R.~Molinaro, ``Estimates on the generalization error of
  physics-informed neural networks for approximating {PDEs},'' \emph{IMA
  Journal of Numerical Analysis}, 2022.

\bibitem{bazmara2023physics}
M.~Bazmara, M.~Silani, M.~Mianroodi \emph{et~al.}, ``Physics-informed neural
  networks for nonlinear bending of {3D} functionally graded beam,'' in
  \emph{Structures}, vol.~49.\hskip 1em plus 0.5em minus 0.4em\relax Elsevier,
  2023, pp. 152--162.

\bibitem{goerguelue2009beam}
U.~Goerguelue, ``Beam theories the difference between {Euler-Bernoulli} and
  {Timoschenko},'' \emph{Lecture Handouts}, 2009.

\bibitem{akellavisualization}
P.~Akella, E.~G. Hemingway, and O.~M. O’Reilly, ``A visualization tool for
  the vibration of {Euler-Bernoulli} and {Timoshenko} beams.''

\bibitem{semper1994semi}
B.~Semper, ``Semi-discrete and fully discrete {Galerkin} methods for the
  vibrating {Timoshenko} beam,'' \emph{Computer methods in applied mechanics
  and engineering}, vol. 117, no. 3-4, pp. 353--360, 1994.

\bibitem{yu2020structural}
Y.~Yu, H.~Yao, and Y.~Liu, ``Structural dynamics simulation using a novel
  physics-guided machine learning method,'' \emph{Engineering Applications of
  Artificial Intelligence}, vol.~96, p. 103947, 2020.

\bibitem{zhang2021structural}
Z.~Zhang and C.~Sun, ``Structural damage identification via physics-guided
  machine learning: a methodology integrating pattern recognition with finite
  element model updating,'' \emph{Structural Health Monitoring}, vol.~20,
  no.~4, pp. 1675--1688, 2021.

\bibitem{karpatne2017physics}
A.~Karpatne, W.~Watkins, J.~Read, and V.~Kumar, ``Physics-guided neural
  networks ({PGNN}): An application in lake temperature modeling,'' \emph{arXiv
  preprint arXiv:1710.11431}, vol.~2, 2017.

\bibitem{khandelwal2020physics}
A.~Khandelwal, S.~Xu, X.~Li, X.~Jia, M.~Stienbach, C.~Duffy, J.~Nieber, and
  V.~Kumar, ``Physics guided machine learning methods for hydrology,''
  \emph{arXiv preprint arXiv:2012.02854}, 2020.

\bibitem{yu2022gradient}
J.~Yu, L.~Lu, X.~Meng, and G.~E. Karniadakis, ``Gradient-enhanced
  physics-informed neural networks for forward and inverse {PDE} problems,''
  \emph{Computer Methods in Applied Mechanics and Engineering}, vol. 393, p.
  114823, 2022.

\bibitem{miao2023gpinn}
Y.~Miao and H.~Li, ``{GPINN}: Physics-informed neural network with graph
  embedding,'' \emph{arXiv preprint arXiv:2306.09792}, 2023.

\bibitem{bettencourt2019taylor}
J.~Bettencourt, M.~J. Johnson, and D.~Duvenaud, ``Taylor-mode automatic
  differentiation for higher-order derivatives in {JAX},'' in \emph{Program
  Transformations for ML Workshop at NeurIPS 2019}, 2019.

\end{thebibliography}
\vspace{-33pt}
\begin{IEEEbiography}[{\includegraphics[width=1in,height=1.25in,clip,keepaspectratio]{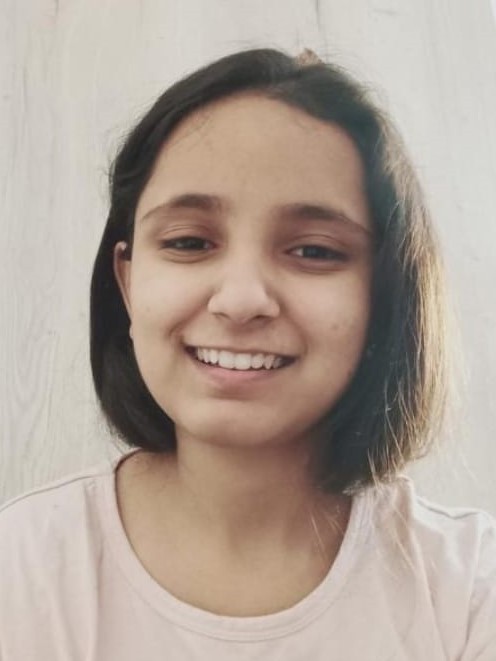}}]{Taniya Kapoor}is currently pursuing Ph.D. degree in the department of Engineering Structures at Delft University of Technology, The Netherlands. Before this, she was an intern in the seminar of applied mathematics at ETH Zürich, Switzerland. She received her M.Sc. degree in applied mathematics from South Asian University, India and in scientific computing from Université de Lille, France. She currently works on scientific machine learning with applications in engineering structures.
\end{IEEEbiography}
\vspace{-33pt}
\begin{IEEEbiography}[{\includegraphics[width=1in,height=1.25in,clip,keepaspectratio]{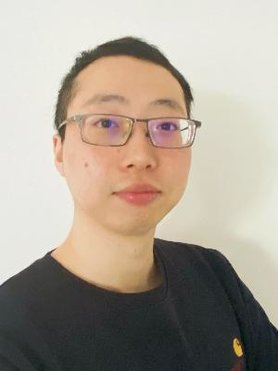}}]{Hongrui Wang }(Member, IEEE) received the Ph.D. degree from the Section of Railway Engineering, Delft University of Technology, Delft, The Netherlands, in 2019. He was a Post-Doctoral Researcher with the Delft University of Technology, until November 2020, where he is currently an Assistant Professor with the Department of Engineering Structures. His research interests include signal processing, artificial intelligence, and their applications in the structural health monitoring and digital modeling and the design of railway infrastructures.

Dr. Wang is an Associate Editor of the IEEE TRANSACTIONS ON INSTRUMENTATION AND MEASUREMENT.
\end{IEEEbiography}
\vspace{-33pt}
\begin{IEEEbiography}[{\includegraphics[width=1in,height=1.25in,clip,keepaspectratio]{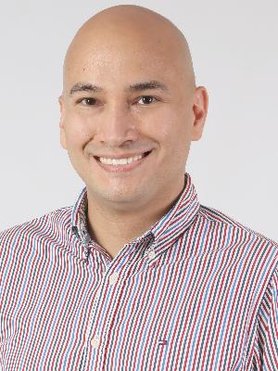}}]{Alfredo Núñez}(Senior Member, IEEE) received the Ph.D. degree in electrical engineering from the University of Chile, Santiago, Chile, in 2010. He is currently an Associate Professor in the field of data-based maintenance for railway infrastructure with the Section of Railway Engineering, Department of Engineering Structures, Delft University of Technology, Delft, The Netherlands. He was a Postdoctoral Researcher with the Delft Center for Systems and Control, Delft University of Technology. He has authored or coauthored more than a hundred international journal and international conference papers. His current research interests include railway infrastructures, intelligent conditioning monitoring and maintenance of engineering structures, computational intelligence, big data, risk analysis, and optimization.

Dr. Núñez is on the Editorial Board of IEEE TRANSACTIONS ON INTELLIGENT TRANSPORTATION SYSTEMS and Applied Soft Computing.
\end{IEEEbiography}
\vspace{-33pt}
\begin{IEEEbiography}[{\includegraphics[width=1in,height=1.25in,clip,keepaspectratio]{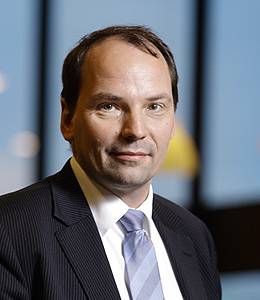}}]{Rolf Dollevoet} received the M.Sc. degree in mechanical engineering from the Eindhoven University of Technology, Eindhoven, The Netherlands, in 2003, and the Ph.D. degree in rail research on rolling contact fatigue from the University of Twente, Enschede, The Netherlands, in 2010. Since 2003, he has been with the Railway Sector, ProRail, Utrecht, The Netherlands. Since 2012, he has been appointed as a part-time Professor with the Section of Railway Engineering, Delft University of Technology, Delft, The Netherlands. He was also a Railway System Expert with ProRail, where he was responsible for all the scientific research and innovation with the Civil Engineering Division, ProRail Asset Management.
Dr. Dollevoet was a recipient of the Jan van Stappen Spoorprijs 2010 Award (a yearly prize for contributions to the travel quality and service for passengers in The Netherlands) from the railway sector for his Ph.D. research and its huge potential to reduce track maintenance costs.
\end{IEEEbiography}
\vspace{-33pt}
\end{document}